\documentclass[12pt]{article}

\usepackage[breakable]{tcolorbox}
\usepackage{placeins}
\usepackage{float}
\usepackage[utf8]{inputenc}

\usepackage{newtxtext,newtxmath}
\usepackage{graphicx}

\usepackage[letterpaper,margin=1in]{geometry}

\renewenvironment{abstract}
	{\quotation}
	{\endquotation}

\date{}

\makeatletter
\renewcommand{\fnum@figure}{\textbf{Figure \thefigure}}
\renewcommand{\fnum@table}{\textbf{Table \thetable}}
\makeatother

\usepackage{scicite}

\usepackage{url}

\newcommand {\bsec}[2]{\subsection*{#1}
                       \label{sec:#2} }

\newcommand {\bsubsec}[2]{\mymarginpar{sec:#2}
                       \subsubsection*{#1}
                       \label{sec:#2} }

\newcommand {\rappendix}[1]{(Supplementary Text)}

\newcommand {\mymarginpar}[1]{\marginpar{#1}}
\renewcommand {\marginpar}[1]{}

\newcommand {\rfig}[1]{Fig.~\ref{fig:#1}}
\newcommand {\rappendfig}[1]{Fig.~\ref{fig:#1}}

\newcommand {\beq}[1]{
                      \begin{equation}
                      \label{eq:#1} }
\newcommand {\eeq}{\end{equation}}
\newcommand {\beqno}[1]{\begin{eqnarray}
                      \nonumber}

\newcommand {\eeqno}{ && \end{eqnarray}}

\newcommand {\bear}[1]{
                       \begin{eqnarray}
                       \label{eq:#1} }

\newcommand {\bearno}[1]{
                       \begin{eqnarray}
                       \nonumber}

\newcommand {\eear}{\end{eqnarray}}
\newcommand {\eearno}{\end{eqnarray}}

\newcommand {\rtab}[1]{Table \ref{tab:#1}}

\newcommand {\shil}[1]{{\color{red}{#1}}}

\newcommand{\mkclean}{
   \renewcommand{\shil}[1]{}
}

\mkclean

\def\scititle{LLM-based MOFs Synthesis Condition Extraction\\using Few-Shot Demonstrations}
\title{\bfseries \boldmath \scititle}

\author{
	Lei Shi$^{1\dagger}$,
	Zhimeng Liu$^{2\dagger}$,
    Yi Yang$^{1}$,
	Weize Wu$^{1}$,
    Yuyang Zhang$^{1}$,
    Hongbo Zhang$^{3}$\and
    Jing Lin$^{2}$,
    Siyu Wu$^{1}$,
    Zihan Chen$^{1}$,
    Ruiming Li$^{1}$,
    Nan Wang$^{1}$,
    Yuankai Luo$^{1}$\and
    Rui Wang$^{3}$,
    Zipeng Liu$^{1}$,
    Huobin Tan$^{1}$,
    Hongyi Gao$^{2\ast}$,
    Yue Zhang$^{3\ast}$,
    Ge Wang$^{2\ast}$\and
 	\small$^{1}$School of Computer Science, Beihang University, Beijing \& 100191, China.\and
	\small$^{2}$ School of Materials Science \& Engineering, University of Science and Technology Beijing, 100083, China.\and
    \small$^{3}$Westlake University, Hangzhou \& 310030, China.\and
	\small$^\ast$Corresponding author. Email: hygao@ustb.edu.cn, yue.zhang@wias.org.cn, gewang@ustb.edu.cn\and
	\small$^\dagger$These authors contributed equally to this work.
}

\begin{document}

% Insert the title and author list
\maketitle

% Abstract, in bold
% There are strict length limits, and not all formats have abstracts.
% Consult the journal instructions to authors for details.
% Do not cite any references in the abstract.
\begin{abstract} \bfseries \boldmath
The extraction of Metal-Organic Frameworks (MOFs) synthesis route from literature has been crucial for the logical MOFs design with desirable functionality. The recent advent of large language models (LLMs) provides disruptively new solution to this long-standing problem. While the latest researches mostly stick to primitive zero-shot LLMs lacking specialized material knowledge, we introduce in this work the few-shot LLM in-context learning paradigm. First, a human-AI interactive data curation approach is proposed to secure high-quality demonstrations. Second, an information retrieval algorithm is applied to pick and quantify few-shot demonstrations for each extraction. Over three datasets randomly sampled from nearly 90,000 well-defined MOFs, we conduct triple evaluations to validate our method. The synthesis extraction, structure inference, and material design performance of the proposed few-shot LLMs all significantly outplay zero-shot LLM and baseline methods. The lab-synthesized material guided by LLM surpasses 91.1\% high-quality MOFs of the same class reported in the literature, on the key physical property of specific surface area.
\end{abstract}

\bsec{Introduction}{Intro}

The extraction of material synthesis route from scientific texts using machine learning techniques has long been a popular task in AI for science \cite{vaucher2020automated,trewartha2022quantifying,dagdelen2024structured,choi2024accelerating} as well as Cheminformatics \cite{he2020similarity,park2022mining,glasby2023digimof,zheng2023chatgpt,sasidhar2023enhancing,zhang2024fine}. This work explores the topic based on an emerging family of 100k+ crystalline porous materials called Metal-Organic Frameworks (MOFs) \cite{MOFsDef}. MOFs have been widely used in catalysis \cite{sun2024accelerated}, gas adsorption \cite{wang2024comprehensive}, energy storage \cite{wu2017metal}, and many other fields \cite{xue2023customized,alawadhi2024harvesting} due to their quantitatively tunable structure and functionality driven by flexible synthesis. These characteristics make MOFs a typical material class where precise and comprehensive knowledge of their synthesis route becomes extremely critical for further material design and application \cite{yaghi2003reticular, chen2020machine}.

% Related work & background on ML-based literature analysis for synthesis extraction and more tasks, limitations (What limitations?)
% Related work on LLM on the same topic & their advantages over classical ML methods

Currently, there have been 100k+ MOFs successfully synthesized in the laboratory. Their detailed synthesis conditions are basically recorded by academic literature in various textual or tabular formats. Machine learning methods have been applied to the literature text to automatically extract synthesis conditions, ranging from basic pattern recognition methods in the natural language processing (NLP) field \cite{swain2016chemdataextractor,glasby2023digimof} to advanced deep learning models \cite{vaucher2020automated,trewartha2022quantifying,kim2017machine,he2020similarity,park2022mining,gupta2022matscibert}, and Large Language Models (LLMs) \cite{zhang2024fine,dagdelen2024structured,polak2024extracting,zheng2023chatgpt}. Before the LLM era, the best-performing model using exact-match evaluations (\rtab{MLvsLLM}(A)), i.e. He et al. \cite{he2020similarity}, achieves an F1 score of 0.9, though on a much easier task of precursor name extraction (our model has macro-F1$\geq$0.98 on the same task). In addition, classical supervised learning approaches suffer from complex model building and parameter tuning overhead, and inherent subjective errors due to human-annotated training data \cite{huang2020database}.

The introduction of LLMs to materials chemistry \cite{boiko2023autonomous, zheng2023gpt,zheng2023shaping,kang2024chatmof,polak2024extracting} brings a disruptively new solution to the problem of material synthesis route extraction owing to LLMs' strong capability in handling disparate forms of scientific texts with the same off-the-shelf model interface. Among state-of-the-art results (\rtab{MLvsLLM}(B)), the best model for our task is the zero-shot GPT-3.5 by Zheng et al. \cite{zheng2023chatgpt}, which achieves a macro-F1 of 0.92 on the SIMM dataset. In this work, by conducting comprehensive evaluations, we find that the performance of zero-shot LLM falls from 0.92 by subjective evaluation in \cite{zheng2023chatgpt} (or called manual score in \cite{dagdelen2024structured}) to 0.74 by objective evaluation (or called exact match in \cite{dagdelen2024structured}), due to its well-known deficiency of lacking specialized knowledge in sparse scenario such as MOFs \cite{brown2020language}. It is further shown by our computational and material experiments that, the low performance and non-exact-match of zero-shot LLMs let down the material structure inference metrics by 23.9\% and key MOFs physical property by 75.1\%, when applied to the important tasks of material inference and design.

%It should be noted that the baseline zero-shot LLMs are notorious for their low performance on sparse scenarios like material synthesis, which are infrequently covered by the general-purpose LLM training data \cite{brown2020language}.

This paper introduces the few-shot in-context learning paradigm as the standard approach to augment general-purpose LLMs on the material synthesis extraction problem. First, to secure high-quality demonstrations for few-shot LLMs and overcome the accuracy bottleneck of human annotations, we leverage the complementary nature of human expertise and AI intelligence, and propose a human-AI interactive data curation process. Our method enjoys the best of both worlds and offers the highest data quality in ground-truth demonstrations produced. Second, to quantify the optimal number of few-shot demonstrations, we propose to apply the popular retrieval-augmented generation (RAG) algorithms to adaptively select a small number of 4$\sim$6 best few-shots combination for each synthesis route extraction. The approach efficiently supports high-throughput synthesis extraction and flexible application deployment by greatly cutting down the input context size for commercial LLMs, in comparison to expensive technologies such as fine-tuning \cite{dagdelen2024structured,zhang2024fine} requiring thousands of demonstrations. Finally, the proposed few-shot LLM method is further enhanced with external material knowledge via prompt engineering, and integrated with offline machine learning models and post-processing to form an end-to-end synthesis extraction pipeline for efficiently processing large amount of literature text.

We conduct triple evaluations to validate the proposed method. On three MOFs datasets, the few-shot LLM consistently delivers synthesis extraction accuracy higher than existing methods (0.94 in average macro-F1 vs. 0.77 by zero-shot LLM). To our knowledge, this is the first method that achieves a full-set material synthesis route extraction accuracy above 0.9 under exact-match evaluations. On material structure inference and design evaluations, the proposed few-shot LLM again outperforms state-of-the-art significantly. The actually-synthesized material sample surpasses 91.1\% high-quality MOFs reported in the literature, on the BET-measured surface area value.

\begin{table}[!t]
    \centering
    \caption{\textbf{Material info extraction performance reported in the literature where the most relevant task from each paper is listed: (A) classical ML models; (B) LLMs.} Here exact match means a true positive is counted only when the extracted info is exactly the same with the annotated ground-truth. Manual/similarity scores mean a true positive can be counted if only the extracted info is considered correct by a human judge \cite{dagdelen2024structured} or is similar enough, in comparison to the ground-truth.}
    \vspace{0.1 in}
    \includegraphics[width=\textwidth]{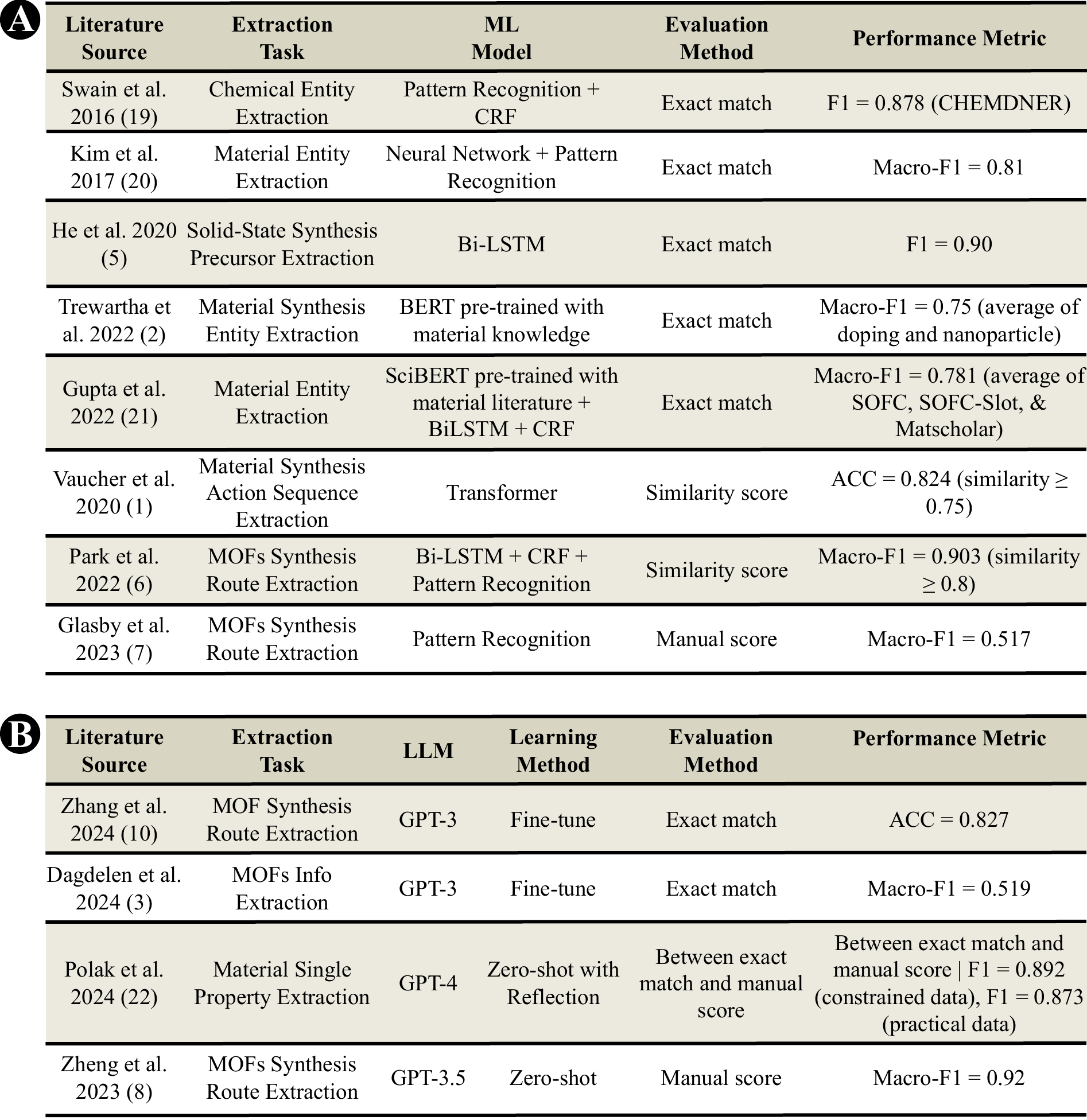}
    \label{tab:MLvsLLM}
\end{table}

\begin{figure}[!t]
\centering
\includegraphics[width=0.85\textwidth]{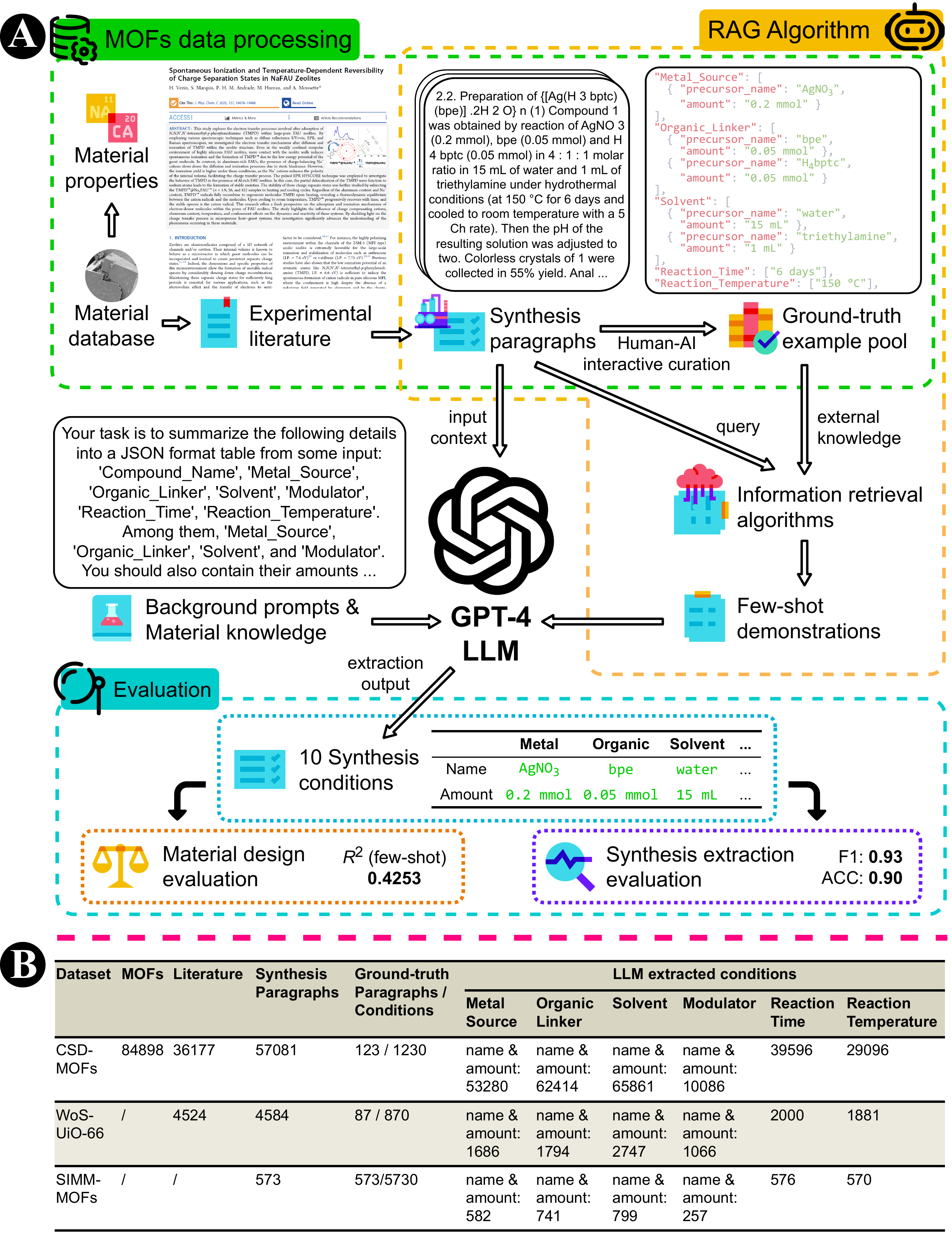}
% \vspace{-0.1 in}
\vspace{-0.6em}
\caption{\textbf{Overview of our MOFs synthesis route extraction proposal using few-shot LLM in-context learning method.} (A) The technical workflow is composed of multiple core components including data processing on material database, the RAG algorithm to select few-shot demonstrations, the LLM engine, and the multi-faceted material evaluation; (B) Statistics of MOFs datasets used in this work.}
%\vspace{-0.15 in}
% \vspace {-1.0em}
\label{fig:OverallPipeline}
\end{figure}

\bsec{Results}{Result}

% Our overall workflow

% Main Workflow Figure: literature collection & annotation (fixed and adaptive annotation), building sample paragraph database, few-shot retriever (RAG method), LLM, MOFs Synthesis Database (application IV), MOFs micro-structural property prediction (application II), MOFs in-field synthesis

% Summary of the best result and method

\begin{figure}[!htb]
\centerline{\includegraphics[width=0.95\linewidth]{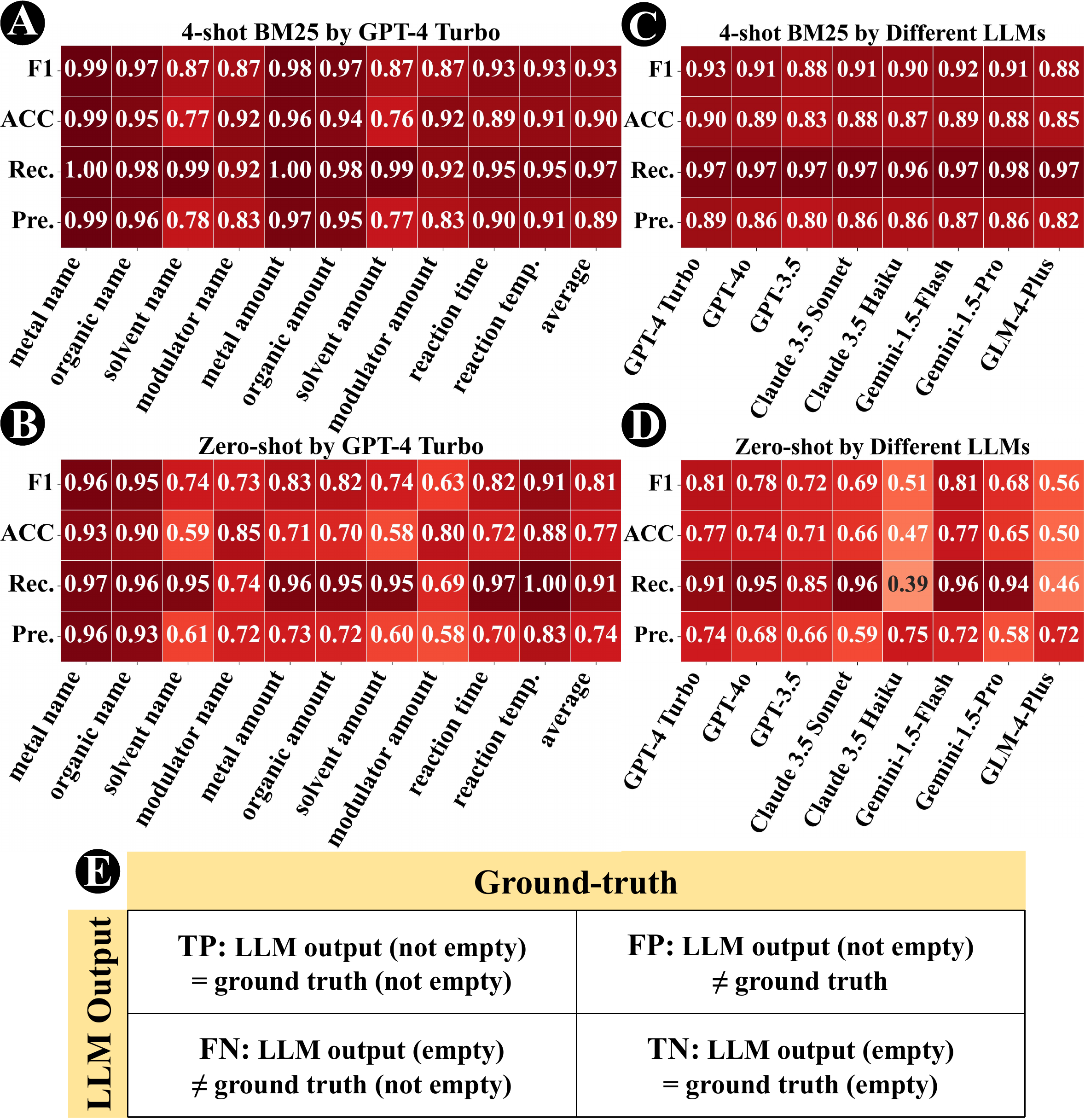}}
% \vspace{-0.1 in}
\vspace {-.9em}
\caption{\textbf{Synthesis route extraction performance on the CSD-MOFs dataset.} Key indicators (F1, ACC, Recall, Precision) are listed. Only literature with annotated ground-truth are included. (A) the proposed few-shot LLM (GPT-4 Turbo) with RAG algorithm; (B) zero-shot LLM as the baseline; (C) few-shot approach on 8 different LLMs; (D) zero-shot approach on 8 different LLMs; (E) confusion matrix definition for performance evaluation.}
%\vspace{-0.15 in}
% \vspace{-1.0em}
\label{fig:BestVSZero}
\end{figure}

Our technical workflow is given in \rfig{OverallPipeline}(A). The MOFs experimental literature is collected from three data sources (\rfig{OverallPipeline}(B), \rappendix{MOFs-Dataset}). The first is the CSD material database v5.43 \cite{moghadam2017development} where we download 36,177 papers covering 84,898 unique MOFs (CSD-MOFs), the second is Web of Science platform \cite{WoS} where we retrieve all the 4,524 papers related to the class of UiO-66 MOF with Zr as metal (WoS-UiO-66), and the last is the SIMM data from Zhang et al. \cite{zhang2024fine} containing 573 MOFs' synthesis route manually annotated over Zheng et al. \cite{zheng2023chatgpt}'s raw data. The full-text of each paper is pre-processed to locate paragraphs relevant to MOFs synthesis (Materials and Methods). The GPT-4 LLM \cite{GPT-4} is employed to extract 10 essential conditions from each paragraph (\rtab{MaterialKnowledge}). The synthesis extraction result is first evaluated on their literal accuracy with respect to ground-truth data, and then tested on the real-world scenarios of MOFs structure inference and synthesis. Take the CSD-MOFs dataset for example, the proposed few-shot LLM method achieves much higher extraction accuracy in 7 out of 10 synthesis conditions (\rfig{BestVSZero}(A), macro-F1=0.93), than the baseline zero-shot LLM (\rfig{BestVSZero}(B), macro-F1=0.81). This advantage is consistent on all 8 state-of-the-art LLMs tested (\rfig{BestVSZero}(C) vs. (D)). The performance at the other two datasets also show similar comparison results (\rappendix{UiO66FewShot}, \rappendfig{UiO66BestVSZero}, \rappendfig{SIMMBestVSZero}). Our fully-tuned high-throughput synthesis extraction workflow now processes over 500 million of scientific texts from all available MOFs literature within 7 hours.

\bsubsec{Human-AI Interactive Data Curation}{Curation}
% Lei & YangYi
%%3
%% Perf-Fig.3

\begin{figure}[!htb]
\centerline{\includegraphics[width=0.95\linewidth]{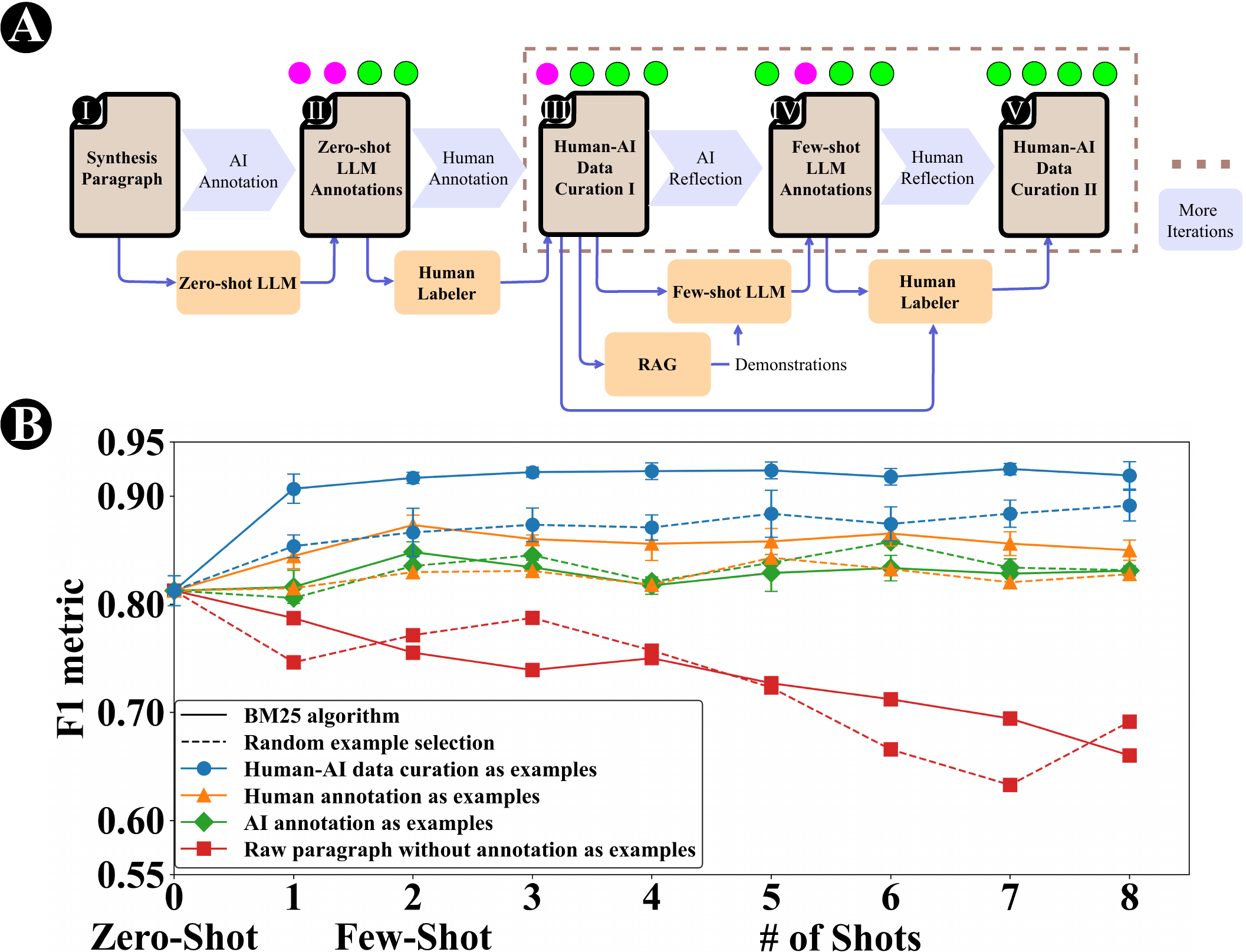}}
% \vspace{-0.1 in}
\vspace {-.6em}
\caption{\textbf{Human-AI interactive data curation method.} (A) the overall workflow to improve the data quality of few-shot examples; (B) the synthesis extraction performance by different data curation methods and varying number of shots. The 95\% CI error bar by 5 repeated tests is only displayed on the human-AI curation plot to reduce visual clutter.}
%\vspace{-0.15 in}
% \vspace {-1.0em}
\label{fig:HumanAIReflection}
\end{figure}

% impact of quality of k-shots

% Human annotation & result & gap from the best result

A prerequisite to our few-shot LLM method is to obtain a high-quality example pool, which should contain the ground-truth synthesis conditions annotated over a set of sample synthesis paragraphs (see an example in the RAG block of \rfig{OverallPipeline}(A)). Otherwise, the few-shot examples without ground-truth annotation will lead to worse performance even than the zero-shot LLM without examples (red+square lines in \rfig{HumanAIReflection}(B)). We first apply a standard human annotation protocol \rappendix{annotation} and software (\rappendfig{AnnotationPlatform}). On the CSD-MOFs dataset, the few-shot LLM using human-annotated example pool achieves a best synthesis extraction macro-F1 score of 0.87 (orange+triangle lines in \rfig{HumanAIReflection}(B)), lagging behind our F1 target of 0.9.

% Human curation over (BM-25 model over human annotation examples) % Best result
% Potential for iterative human-model data curation

We then consider the latest approach of automated AI annotation, in which LLM is initially applied in a zero-shot mode to extract all synthesis conditions. The zero-shot LLM output is then used as ground-truth annotations in a 2nd-round few-shot LLM in-context learning, which generates the final AI annotations. It is shown through experiments that the best few-shot LLM using AI annotated example pool also achieves a macro-F1 of 0.86 (green+diamond lines in \rfig{HumanAIReflection}(B)). Neither human annotations nor purely AI-generated examples achieve the best data quality. In fact, human expertise and AI's capacity in labeling ground-truth synthesis conditions are complementary to each other (more in (Discussion)).

% add the advantage of each step (reduce errors, remove fatigue, etc.)

We propose a new approach of human-AI interactive data curation to retrieve the best-quality ground-truth examples. As shown in \rfig{HumanAIReflection}(A).I, raw synthesis paragraphs are first processed by LLM in a zero-shot mode to obtain an initial AI annotation (\rfig{HumanAIReflection}(A).II). Human labelers then work on the AI annotation and achieve a best-effort human annotation (\rfig{HumanAIReflection}(A).III), which is the first round of human-AI interactive data curation. The complementary advantage here lies in that AI generates annotation in a highly efficient way and greatly alleviates the fatigue issue of human labelers, who can focus on rare cases requiring specialized knowledge. In the second round, the few-shot LLM is applied with the human annotation output in the first round as ground-truth examples. The output (\rfig{HumanAIReflection}(A).IV) represents a reflection by AI that resolves potential random errors made by human labelers in the last round. Finally, human labelers combine the latest human and AI annotations, and achieve the second round of human-AI interactive data curation (\rfig{HumanAIReflection}(A).IV). Here only the cases where human and AI disagree with each other are re-examined. In fact, this human-AI interactive curation process, as in the dashed frame of \rfig{HumanAIReflection}(A), can iterate more than one round. In our work after the second reflection, an excellent performance has already been achieved (macro-F1=0.93, blue+circle lines in \rfig{HumanAIReflection}(B)). The final ground-truth example pool on CSD-MOFs dataset includes 123 rows of 1230 synthesis conditions in total.% 23 rows of human-annotated conditions were removed as they contain multiple suites of synthesis conditions for either one or more MOFs (e.g. chiral MOFs). Though our technical framework can deal with the case of having multiple MOFs in a single synthesis paragraph, we only select the annotated paragraphs describing the synthesis of only one MOF as demonstrations for performance consideration.

\bsubsec{Few-Shot Large Language Model with Material Knowledge}{Few-shot}
%%1

At the core of our technical workflow (\rfig{OverallPipeline}(A)), we apply the GPT-4 Turbo LLM, which exhibits the highest extraction performance among state-of-the-art LLMs (\rfig{BestVSZero}(C)). The latest few-shot in-context learning approach (FS-ICL \cite{brown2020language}) is introduced, which refers to a typical learning paradigm to adapt the task-agnostic language models to various downstream tasks while achieving optimized performance on each task. In more detail, FS-ICL takes a few prompted examples as input (known as shots), each composed of a context and a labeled completion, in addition to background prompts such as task description. For the task of MOFs synthesis route extraction, a context refers to a paragraph containing all the synthesis conditions of a MOF and the labeled completion refers to the ground-truth synthesis conditions annotated and curated by human and AI in our work. The final LLM extraction is made by prompting a new context and asking the language model to complete it.

% Advantage of few-shot ICL vs. fine tuning and others
% allow the same LLM to immediately perform a wide variety of tasks

% \cite{liu2022few, mosbach2023few}

The FS-ICL approach enjoys high flexibility to work on many tasks (e.g., synthesis extraction of various materials) without the need to re-train the model in contrast to fine-tuning \cite{liu2022few, mosbach2023few}. Yet, FS-ICL still faces multiple challenges. First, the cost induced by few-shots to LLM needs to be quantified. Second, the prompt format in FS-ICT (e.g., the wording and ordering of examples) can bring uncertainty to the extraction performance. How to alleviate this uncertainty remains unknown.

%It is also previously believed that FT can achieve better performance than FS-ICT, but the latest study reveals that under the same size of shots, both paradigms obtain similar performance and exhibit large variance depending on the task specification \cite{mosbach2023few}. In our scenario, FS-ICT reaches an excellent performance of F1$>$0.9, which is enough for real-life deployment.

\emph{Using RAG to optimize few-shot data quality and quantity}

% Lei after Hongbo
%%4
% Few-shot algorithm comparison
% Example order

%****************TO CHECK*******************

\begin{figure}[!t]
\centerline{\includegraphics[width=0.85\linewidth]{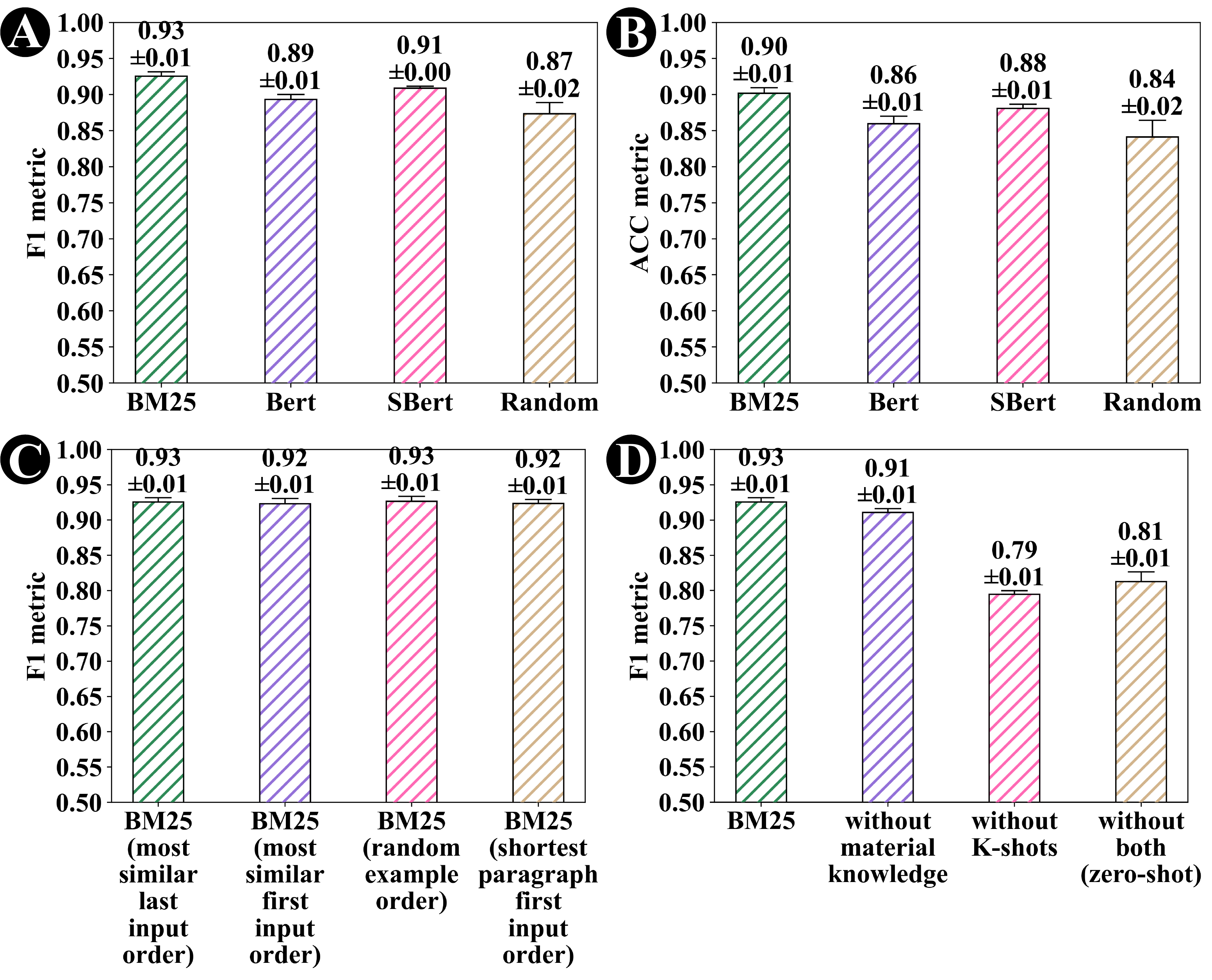}}
% \vspace{-0.1 in}
\vspace {-.6em}
\caption{\textbf{The synthesis route extraction performance by RAG configurations.} (A)(B) comparison of RAG algorithms, on F1 and ACC respectively; (C) comparison of different few-shot orders; (D) the impact of LLM prompt compositions. Error bars indicate 95\% CI within 5 repeated tests.}
%\vspace{-0.15 in}
% \vspace {-1.0em}
\label{fig:BasicComparison}
\end{figure}

We introduce the RAG algorithm \cite{Zhu2023LargeLM, ram2023context} which retrieves the best $K$-shot examples for each input context to augment the LLM and then generates the predicted completion. Three mainstream RAG algorithms are applied here: BM25 \cite{robertson2009probabilistic}, BERT \cite{devlin2018bert}, and Sentence-BERT (SBERT) \cite{reimers2019sentence}. They differ in how to compute the similarity between the input context and candidate examples (Materials and Methods). We then conduct an experiment to compare these RAG algorithms, using the CSD-MOFs dataset. By default, a typical setting of $K=4$ is used.

Experimental results in \rfig{BasicComparison}(A)(B) indicate that the BM25 algorithm (first column) achieves the best synthesis extraction performance among all the compared algorithms, with a macro-F1 of 0.93 and overall accuracy (ACC) of 0.90. The result is quite stable for each algorithm within 5 repeated runs, as shown by the 95\% CI error bar (0.006 for BM25 on F1). Notably, any of the tested algorithms is significantly better than a random selection of examples on F1 (the last column), e.g., $t(8) = 4.68, p=.0016$ by a two-tailed t-test comparing Bert and random algorithms. This showcases the effectiveness of RAG mechanism. On the best BM25 algorithm, we further test the impact of few shots' input order within the LLM prompt. As shown in \rfig{BasicComparison}(C), the differences are not significant among all the tested ordering strategies ($p > 0.05$ by two-tailed t-tests). We decide to fix the few-shot input order to the most similar last setting with the best performance. Next, we evaluate the impact of composition of LLM prompts. The ablation study results in \rfig{BasicComparison}(D) demonstrate that the few-shot examples are the primary drive of performance improvement, followed by the material knowledge provided as background prompts. We then study the optimal few-shot quantity required by RAG algorithms (i.e., $K$). \rfig{HumanAIReflection}(B) and \rappendfig{ShotsComparisonACC} illustrate that, with the best BM25 RAG algorithm and the proposed human-AI ground-truth annotation (solid blue line with circle symbol), both F1 and ACC increase the most from zero-shot to one-shot, and continue to grow until the peak of $K=4$ ($F1=0.93$, $ACC=0.90$). Meanwhile, the few-shot method with random example selection (dashed blue lines) shows consistently lower performance than the BM25 algorithm. This result again demonstrates the effectiveness of the proposed RAG algorithm.

\emph{Sizing the few-shot example pool}

\begin{figure}[t]
\centerline{\includegraphics[width=0.95\linewidth]{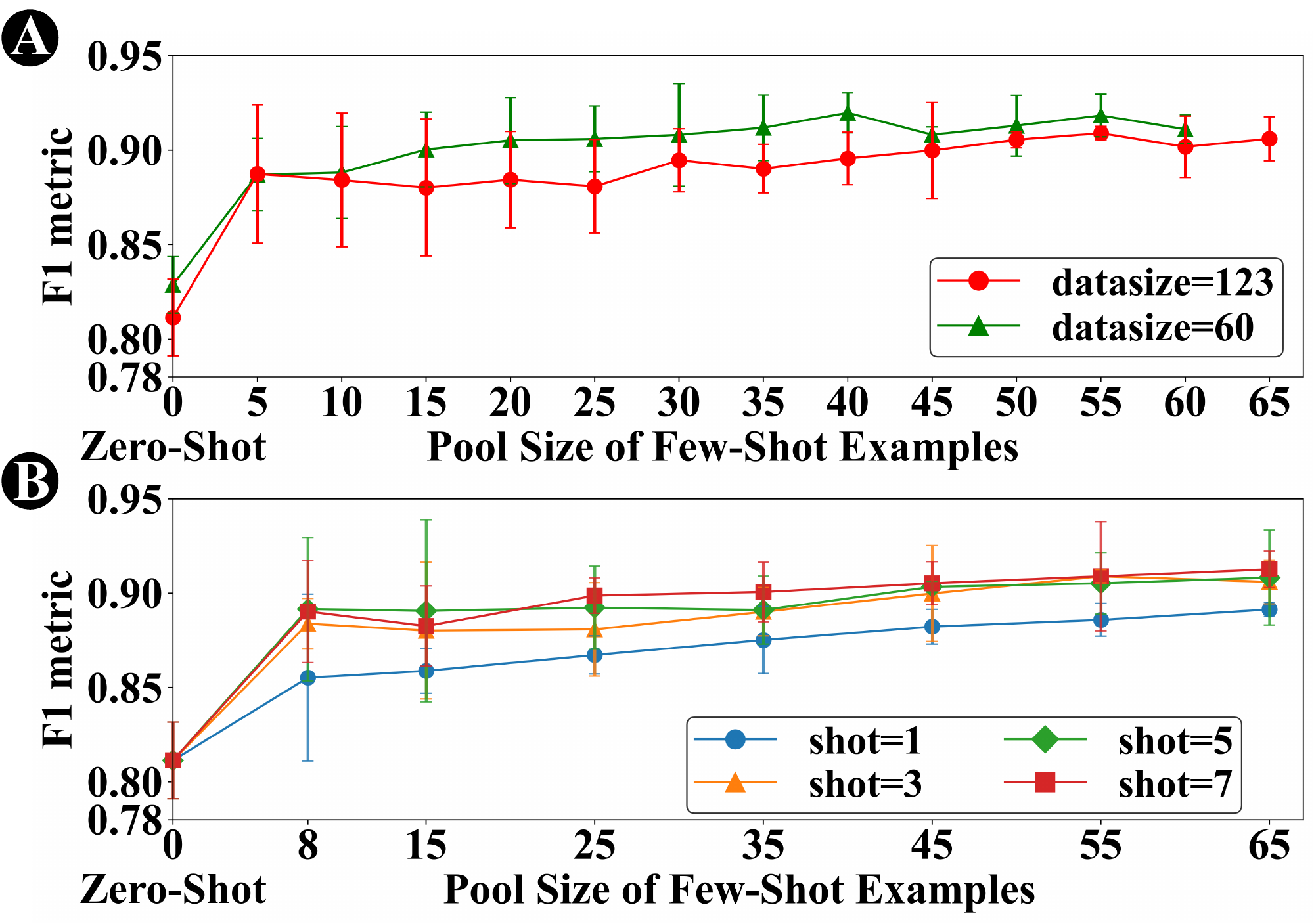}}
% \vspace{-0.1 in}
\vspace {-1.2em}
\caption{\textbf{MOFs synthesis route extraction performance using example pool of varying sizes.} (A) average F1 and its 95\% CI on the 123-paragraph and 60-paragraph CSD-MOFs datasets ($K$=3 for both); (B) on the 123-paragraph dataset using different $K$-shots.}
%\vspace{-0.15 in}
% \vspace {-1.0em}
\label{fig:ExamplePoolSize}
\end{figure}

% Lei after Zihan
% Active learning algorithm for adaptive annotation pool selection
% =>
% Annotation pool size selection

After optimizing both the quantity and quality of few-shot examples, we then study how large an example pool with ground-truth annotations is required for high-throughput MOFs synthesis extraction. Another experiment is set up that assumes the full dataset to be the 123 synthesis paragraphs of CSD-MOFs with known ground-truth conditions. The example pool is randomly chosen from the dataset. In another setting, a subset of 60 ground-truth synthesis paragraphs  of CSD-MOFs is used as the full dataset.

% 1: the first few annotations has the most benefit (same with 1-shot over zero-shot
% 2: smaller dataset requires smaller number of annotations to achieve the same level of performance
% 3: more annotations will bring more gains and less uncertainty until very late (e.g., 2/3 of the data size, if random annotation pool selection is applied. Introducing active learning algorithms may be useful.

\rfig{ExamplePoolSize}(A) illustrates the impact of example pool size on the MOFs synthesis extraction performance. The initial increase of example pool size (from 0 to 5 in the figure) contributes the most performance gain, regardless of the size of the full dataset. This is coherent with the effect observed on zero-shot vs. one-shot learning. More annotations and larger example pools will almost always bring performance gains and less uncertainty, up to 66.7\% and 52.8\% of the full dataset, in the two settings respectively. A smaller dataset (green+triangle line) requires a pool with fewer examples than a larger dataset (red+circle line), in achieving the same level of performance.

\rfig{ExamplePoolSize}(B) further demonstrates the impact of both example pool size and $K$-shots. With an example pool size no smaller than 45, the extraction performance becomes indistinguishable with shot settings of $K \geq 3$. Meanwhile, the one-shot learning has constant a performance gap, suggesting a setting of $K \geq 3$ at least.

\bsubsec{MOFs Structure Inference}{cor}

% Zhimeng
% Micro-structure Inference

\begin{table}[!t]
    \centering
    \caption{\textbf{Performance comparison of the MOFs framework density inference task using different machine learning models.} Left: few-shot vs. zero-shot LLMs on the CSD-MOFs dataset; Right: few-shot LLMs vs. classical machine learning algorithm on the KAIST dataset \cite{park2022mining}.}
    \vspace{0.1 in}
    \includegraphics[width=\textwidth]{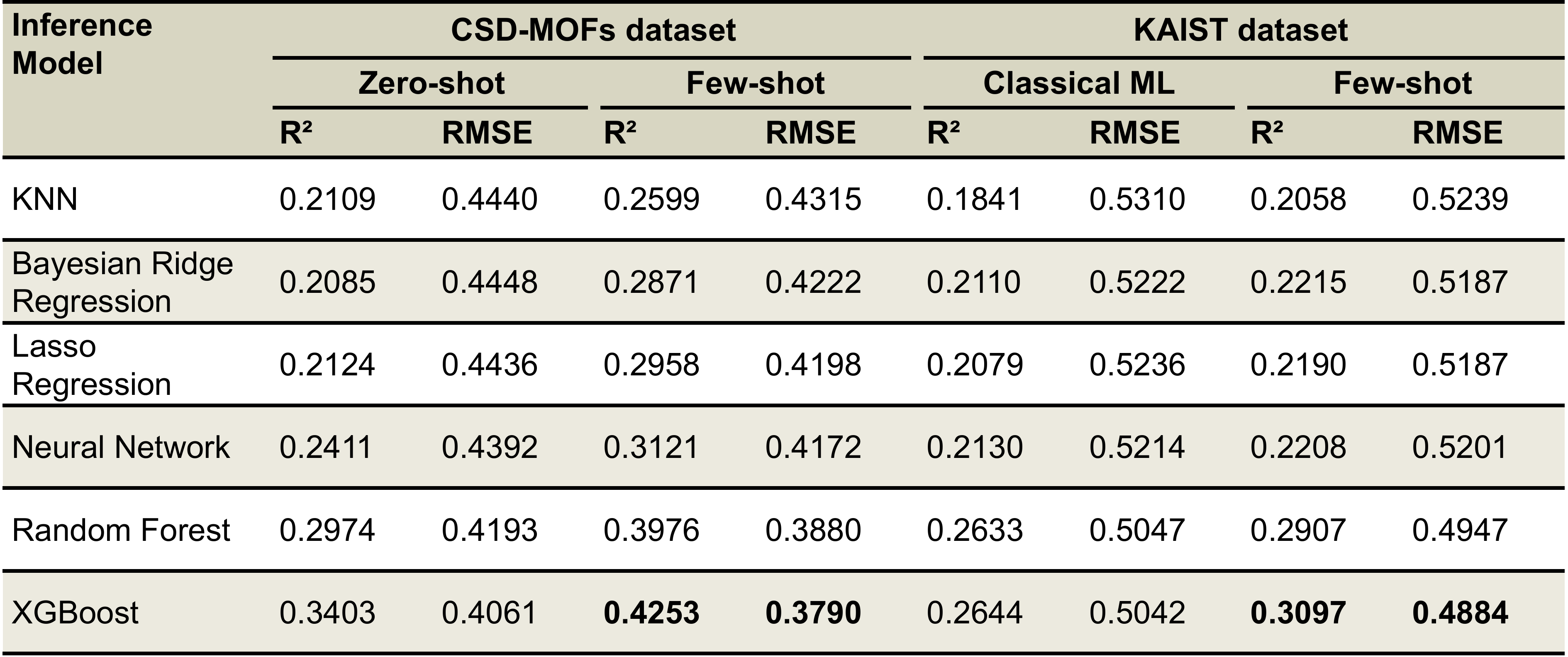}
    \vspace{-0.2 in}
    \label{tab:DensityInferenceTable}
\end{table}

We evaluate the few-shot synthesis extraction method in a real-world MOFs synthesis-structure inference task. The task is to predict the microstructure properties of MOFs, namely global cavity diameter, pore limiting diameter, largest cavity diameter, and framework density, using the extracted synthesis conditions including metals, organic linkers, solvents, modulators, and reaction duration/temperature. The few-shot/zero-shot LLMs and another classical machine learning (ML) algorithm proposed by Park et al. \cite{park2022mining} are compared. Each method is used to extract 10 synthesis conditions from each unique paragraph of the 5269 MOFs in the CSD-MOFs dataset. The raw textual conditions extracted are post-processed by methods in (Materials and Methods) into densely distributed vector representations.

Six ML models are applied for the structure inference: KNN, Lasso Regression, Bayesian Ridge Regression, Neural Networks, Random Forest, and eXtreme Gradient Boosting (XGBoost). The performance is evaluated by the test-set coefficient of determination ($R^2$) in a 10-fold cross-validation. On four microstructure properties inferred, the first three properties irrelevant to MOFs synthesis conditions lead to negative or close to zero $R^2$ in all models, regardless of the extraction method. The last property of MOFs framework density is mostly predictable with synthesis conditions, with the best $R^2$ value larger than 0.4 (\rtab{DensityInferenceTable}). It can be observed in \rtab{DensityInferenceTable} that the synthesis conditions obtained by the few-shot LLM enjoy much larger predictive power than those obtained by the zero-shot LLM, on all six ML models. The $R^2$ values are 31.4\% higher on average and the difference is statistically significant under a paired two-tailed t-test ($t(5)=11.15, p=.0001$). Between the few-shot LLM and the classical ML algorithm, as shown in \rtab{DensityInferenceTable}, the few-shot LLM achieves 8.9\% higher $R^2$ than the classical ML in average with all the 6 predictive models. The difference is statistically significant under a paired t-test ($t(5)=3.55, p=.016$). As the MOFs data size in this trial is smaller than the few-shot vs. zero-shot experiment \rappendix{MOFs-Dataset}, the absolute $R^2$ values are mildly smaller. The comparison on the Root Mean Squared Error (RMSE) metric (\rtab{DensityInferenceTable}) validates the same result that the few-shot LLM enjoys lower predictive error than the zero-shot method and classical ML.

\begin{figure}[!t]
	\centering \includegraphics[width=.95\linewidth]{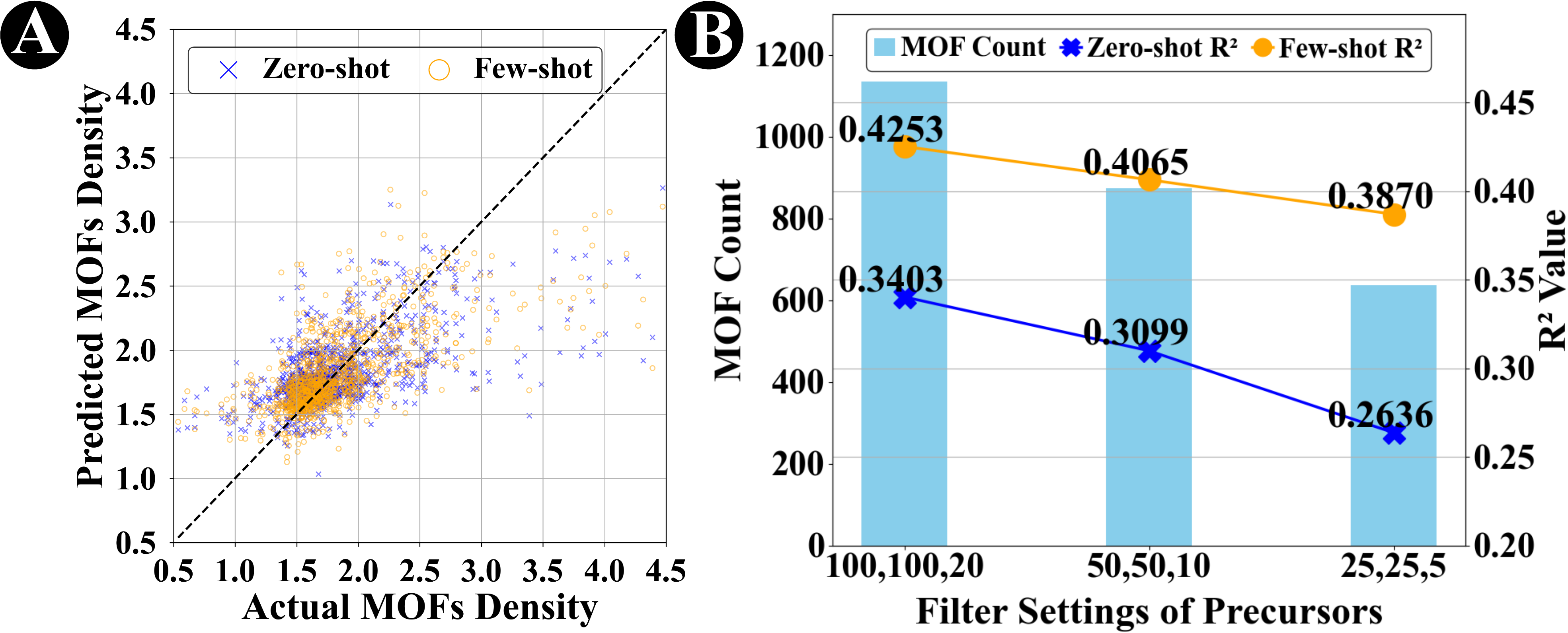}
    \vspace{-0.15 in}
	\caption{\textbf{Detailed MOFs framework density inference performance.} (A) predictive power of the best XGBoost model, few-shot LLM vs. zero-shot LLM; (B) applying different precursor filters.}
	\label{fig:MaterialScatterplot}
\end{figure}

Drilling down to more details, we illustrate the best XGBoost inference result on the scatterplot of \rfig{MaterialScatterplot}(A). It reveals that the actual vs. predicted MOFs density distribution of the few-shot LLM (orange circles) exhibits higher proximity to the optimal prediction line (black dashed line), than the predictions by zero-shot LLM (blue crosses). We also gradually reduce the evaluation dataset by enforcing stricter precursor filters and selecting only higher-ranked synthesis condition values. As shown in \rfig{MaterialScatterplot}(B), the $R^2$ of predictive models by few-shot LLM drops mildly, much slower than the decrease of zero-shot LLM. The result demonstrates that the proposed few-shot LLM not only extracts more accurate synthesis conditions, but also significantly improves the downstream material inference task.

\bsubsec{Real-World Synthesis of New MOFs}{real-syn}

% Problem
% UiO-66 (Zr)
% XRD/SEM/BET experiment output

% Surface Area Inference on UiO-66 MOFs

% PICTURE
% (a)(b) UiO-66 training data and optimization point (surface area distribution), zero-shot vs. few-shot
% (c)UiO-66 few-shot synthesis - XRD (for proof of UiO-66 material) (standard vs. 3 batch of few-shot)
% (d) UiO-66 few-shot synthesis - SEM (for indication of the quality of UiO-66 material) few-shot (vs. zero-shot if comparable);

% Description of the data and actual synthesis task, why select UiO-66 MOFs (large number, easy to synthesis), including paragraph/synthesis extraction/performance (refer to appendix), and post-processing, final synthesis condition data count

% Description of the inference task and surface area extraction, and inference model and performance (may not report)
% Optimization method to infer optimal synthesis conditions, input data (only some synthesis conditions variable due to the single type of UiO-66)

\begin{figure}[!t]
	\centering \includegraphics[width=.95\linewidth]{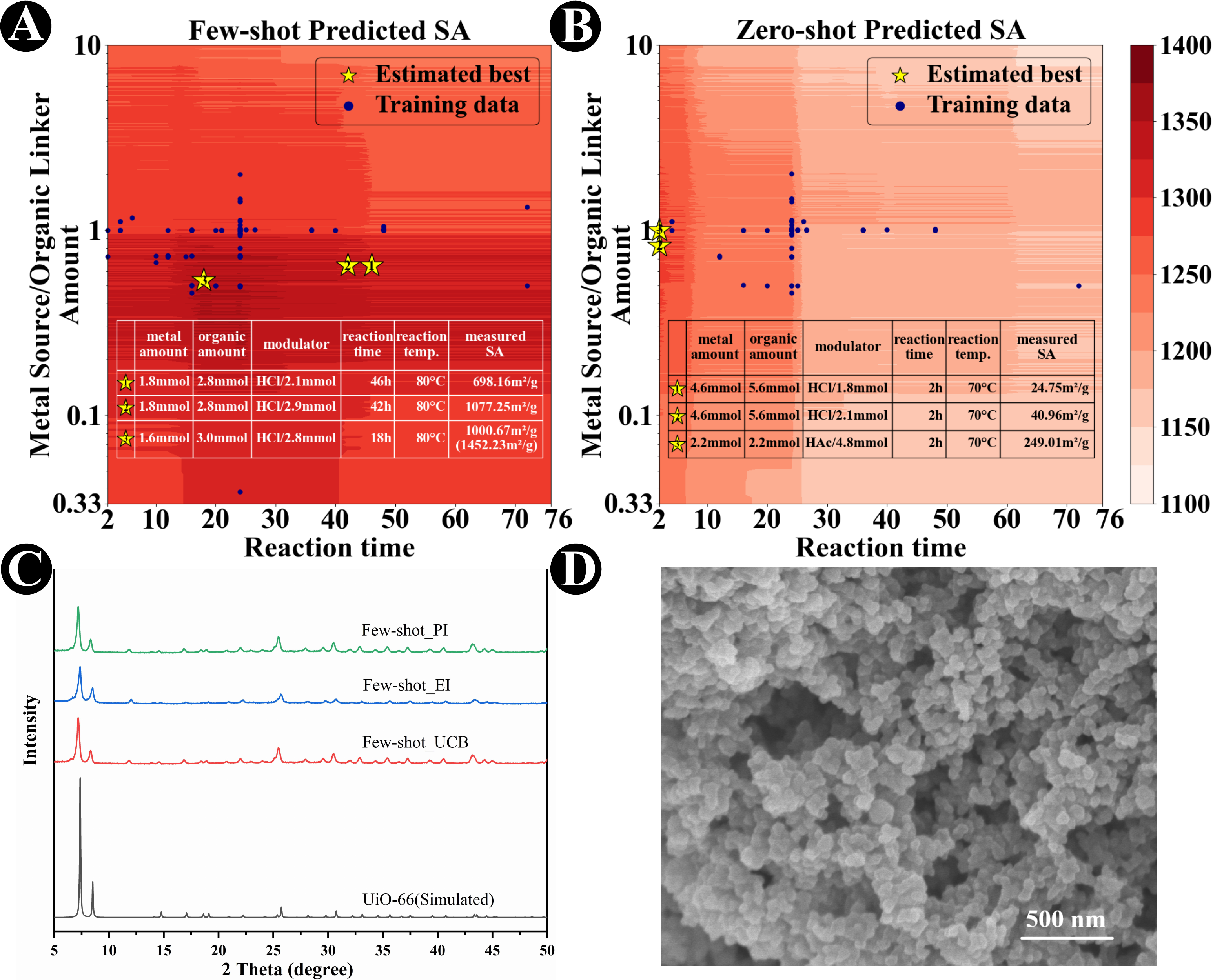}
    \vspace{-0.15 in}
	\caption{\textbf{Real-world MOFs design assisted by LLM.} (A) few-shot LLM extracted synthesis conditions (blue points), the SA inference result (heat map), and the suggested optimal conditions for the largest SA value (yellow stars); (B) results by zero-shot LLM for comparison; (C) XRD interpretation of lab-synthesized UiO-66 MOFs suggested by few-shot LLM; (D) SEM result on one of the MOFs suggested by few-shot LLM.}
	\label{fig:SurfaceAreaInference}
\end{figure}

The synthesis conditions extracted by our method and the baseline are compared in suggesting the best synthesis routes for designing pure UiO-66 (Zr) MOF with the largest specific surface area (SA). The class of UiO-66 MOFs is selected because it is popular in the research community and many literature is available for its synthesis. The mutable synthesis conditions of UiO-66 (Zr) are also limited, with metal and organic percursors fixed, so that most of the suggested synthesis routes are feasible for comparison. The experiment first builds a SA inference model by the random forest algorithm, using the extracted synthesis routes on 261 pure UiO-66 MOFs from the WoS-UiO-66 dataset \rappendix{MOFs-Dataset}.

\rfig{SurfaceAreaInference}(A)(B) present the SA inference model using few-shot and zero-shot LLM-extracted training data (blue points in the figure), respectively. The two most influential synthesis conditions, i.e., reaction time and the ratio of metal precursor amount to organic linker amount, are used as projection dimensions. The heat map in the figures illustrates the maximal SA value inferred from the corresponding synthesis conditions projected to the same point in the view. We apply the Bayesian optimization algorithm (Materials and Methods) to suggest three best synthesis routes for the maximally expected SA value of UiO-66, as indicated by the yellow stars in each figure. The six lab-synthesized UiO-66 samples suggested by few-shot and zero-shot LLMs are gone through BET test to measure their actual SA values. As shown in the embedded table of \rfig{SurfaceAreaInference}(A)(B), the measured SAs of MOFs by few-shot LLM are more than four times as large as the corresponding SA suggested by zero-shot LLM. The SA inference by zero-shot LLM shows a biased model preferring smaller reaction time in UiO-66 MOF synthesis, as in \rfig{SurfaceAreaInference}(B), which can explain its bad performance in MOFs design. UiO-66 (Zr) MOFs have slow reaction kinetics, rendering the suggested 2-hour reaction time insufficient to achieve a well-formed crystalline structure. In comparison, the synthesis output suggested by the few-shot LLM is proven to be pure UiO-66, as shown by the XRD pattern in \rfig{SurfaceAreaInference}(C) w.r.t. the simulated UiO-66. The SEM image in \rfig{SurfaceAreaInference}(D) also shows very high crystallinity in this production. Note that the MOFs measured in the literature mostly go through advanced rinsing and activation processes to increase SA values. We also process one of our best few-shot LLM produced MOF samples with standard activation. The SA value increases from 1000.67 to 1452.23, outperforming 91.1\% UiO-66 MOFs' SA in our literature collection. This showcases the great potential to apply the LLM synthesis extraction method to guide high-performance material design.

% Actual synthesis result and measurement and other proofs: output MOFs amount, surface area, XRD/SEM/BET experiment output

% Other optional
% Extracted UiO-66 conditions (Available paragraphs: 107, Human annotation: 73 + 15 MOFs)

% Lei
%%5
% High-Throughput MOFs Synthesis Database and Online Visualization System 
\bsec{Discussion}{Discussion}

This work studies the new paradigm of applying few-shot LLM in-context learning to the important problem of MOFs synthesis route extraction. It is shown through experiments that both the quality and the quantity of few-shot demonstrations are critical to the LLM performance on MOFs synthesis route extraction. We introduce a novel approach of human-AI interactive data curation to enhance few-shot demonstration quality and a calibrated BM25 RAG algorithm to size the optimal few-shot quantity. Experimental result reveals that a small overhead of 4$\sim$6 shots already achieve excellent performance for few-shot LLM on three typical MOFs datasets (average macro-F1 = 0.94), which are significantly better than the state-of-the-art approach by zero-shot LLM (average macro-F1 = 0.77). Practical issues regarding high-throughput MOFs synthesis extraction and downstream applications are resolved using elaborate methods including offline synthesis paragraph detection, data post-processing, and LLM prompt engineering. Our proposal is also thoroughly evaluated over real-life MOFs structure inference and high-performance design tasks. Compared with the baseline zero-shot LLM, the few-shot LLM method increases the MOFs framework density inference performance by 31.4\% and improves the surface area of lab-synthesized UiO-66 MOFs by more than three times. The lab-synthesized material guided by LLM surpasses 91.1\% high-quality MOFs of the same class reported in the literature, on the key physical property of specific surface area.

In addition, we summarize the following implications by this work.

\bsubsec{Superiority of Human-AI Interactive Data Curation}{Superiority}

We consulted MOF experts to evaluate all errors produced by the few-shot LLM method when human-only annotations are used as ground-truth demonstrations. Out of 261 reported errors, 103 LLM outputs (39.5\%) were identified as actually correct by the expert's reflection, 38 (14.6\%) had certain issues but contributed to refining the corresponding ground-truth, and only 120 (45.9\%) were true errors. In 54.1\% of cases when human and AI extraction have conflicts, AI can help to consolidate a better ground-truth example. This implies that human and AI can be complementary on the task of labeling ground-truth material synthesis conditions. The joint human-AI approach then enjoys three superiorities. First, though human labelers are excellent in the usage of material knowledge, they can fail to strictly follow pre-defined annotation rules. For example, to standardize the solvent condition, it is required to leave out all modifiers of a common solvent. Human annotators often extract ``hot water'' instead of ``water'', as s/he mostly focuses on knowledge extraction but neglects obligatory rules. Humans are poor multi-objective task executors compared with AI. AI introduces less errors when rules are provided in either background prompt or examples. Second, human labelers can suffer from fatigue when working with a large number of annotation tasks, leading to random errors, e.g., missing or adding a few characters/words. AI can be applied to eliminate this issue: a zero-shot LLM annotation in the first round reduces human efforts, their fatigue, and the resulting random errors. Third, general-purpose LLMs alone lack long-tailed material knowledge not appearing in both background prompts and demonstrations. For example, LLM can fail to correctly extract a minor MOF synthesis route appearing very few times in the literature. Human can fix such labeling errors using his/her strong generalization capability.

\bsubsec{Potential Extension to New Materials and Applications}{Potential}

% Overhead of application of this method to other material synthesis scenarios

% Raw data, which should be necessary for all applications

% Human input by our system, very few numbers of them over initial annotation by LLM, the number can be estimated in advance

% Optionally, prompts can be tuned for each application

% Also adapt to tables, pictures, or even movies, structured DBs

The cost to apply our method to extract a new material's synthesis conditions includes: first, the set of experimental material literature should be supplied. This can be off-the-shelf from the users or acquired from academic databases such as Web of Sciences \cite{WoS}. Second, human input is necessary which specifies the exact knowledge to be extracted and its example form in the literature as the seed for few-shot demonstrations. Our method provides an end-to-end solution to augment the human-specified knowledge into background prompts and few-shot demonstrations. More importantly, the number of required annotations, i.e., the example pool size, can be estimated by the empirical result of our work. Third, optionally the input prompt can be tuned per the characteristics of each material to improve the performance. Finally, by introducing newer versions of LLMs such as GPT-4v, extracting from not only textual forms but also pictures and videos can be envisioned. 
\bsec{Materials and Methods}{Method}

% YangYi
\bsubsec{Synthesis Paragraph Detection}{Paragraph}
\label{sec:offline-model}

% Lei after Zihan/YangYi
% Offline machine learning model for synthesis paragraph detection

By the latest GPT-4 turbo pricing (\$10 per 1M tokens), a single pass of LLM over all the 100k MOFs literature can sum up to a non-negligible cost of \$10k (assuming 10k words per literature), while performance tuning normally requires several passes. After investigating a large set of MOFs synthesis literature, it is found that in most cases the authors state the detailed synthesis route of a MOF only in a few paragraphs of the paper, e.g., those starting with ``Synthesis of [a MOF's chemical formula]''. By estimation, the length of these synthesis paragraphs (est. 600 words) only occupies 6\% of the content of a paper in average. By applying an offline synthesis paragraph detection model, the financial cost in using commercial LLMs is reduced by 94\% and the synthesis extraction performance and efficiency are also improved as LLM can now focus on a much smaller set of textual content for extraction.

% 其他方法：NC论文的大模型抽取方法、伯克利的embedding方法，以及更好的bert模型。

% Workflow

To this end, we build a binary classification model to determine whether a paragraph contains a suite of synthesis route. 440 papers from the CSD-MOFs dataset and 87 papers from the WoS-UiO-66 dataset are sampled as the training dataset, where ground-truth synthesis paragraphs are manually annotated as positive samples \rappendix{annotation}. The remaining paragraphs are used as negative samples. In total, 1349/11783 and 87/852 positive/negative synthesis paragraphs are obtained on the two datasets. A standard BERT model \cite{bert} is applied for the classification. A stratified 5-fold cross-validation is used to evaluate on the imbalanced data. Our model achieves F1 = 0.951, ACC=0.989 (CSD-MOFs dataset), and F1 = 0.956, ACC=0.979 (WoS-UiO-66 dataset), on detecting (positive) synthesis paragraphs. Finally, we detect 57081 synthesis paragraphs from 36177 papers on the CSD-MOFs dataset, and 4584 synthesis paragraphs from 4524 papers on the WoS-UiO-66 dataset.

\bsubsec{Few-shot RAG Algorithms}{RAG}

We apply few-shot RAG algorithms \cite{Zhu2023LargeLM}\cite{ram2023context} to improve the LLM performance, which retrieves $K$ demonstrations for each synthesis extraction from a pool constructed by human-AI interactive data curation. Formally, given a pool of demonstrations $D = \{d_1, d_2, \ldots, d_n\}$ and an input paragraph $p$, the RAG algorithm retrieves top $K$ similar demonstrations by:
\begin{equation}
K\text{-shots} = \text{sort}({(\text{score}(p, d_i), d_i)}^n_{i=1})[:K]
\end{equation}
Here, the score function is used to estimate the similarity between document $d_i$ and paragraph $p$. The function can be categorized into two classes by the document representation method: sparse vector encoders (e.g., TF-IDF, BM25~\cite{robertson2009probabilistic}) and semantic dense vector encoders \cite{gao2023retrieval} (e.g., SBERT~\cite{reimers2019sentence}, BERT~\cite{devlin2018bert}). We select the best-performing RAG algorithm, i.e., BM25, as the final choice, according to experimental results. BM25 is a probabilistic information retrieval model that ranks documents based on the frequency of query terms within the documents. It balances term frequency (how often a term appears in a document) with inverse document frequency (how rare a term is across the entire document set), thus giving more weight to terms that are significant. The scoring function of BM25 between a paragraph with $n$ terms and a document in a pool of length $N$ is defined as:
\begin{equation}
\text{Score}(p, d) = \sum_{i=1}^{n} \text{IDF}(p_i) \cdot \frac{f(p_i, d) \cdot (k_1 + 1)}{f(p_i, d) + k_1 \cdot (1 - b + b \cdot \frac{|d|}{\text{avg\_dl}})}
\end{equation}

\begin{equation}
\text{avg\_dl} = \frac{1}{N} \sum_{j=1}^{N} |d_j|
\end{equation}
where $ f(p_i, d) $ is the term frequency of $p_i$ in document $d$, $|d|$ is the length of document $d$, $\text{avg\_dl}$ is the average length of all documents in the demonstration pool, $\text{IDF}(p_i)$ is the inverse document frequency of term $p_i$. $k_1$ and $b$ are hyperparameters of the model and we use the default BM25 settings of $k_1 = 1.5$ and $b = 0.75$ in our experiment.

For the alternative class of semantic information retrieval, we utilize the document embedding by pre-trained language models to compute the similarity:
\begin{equation}
\text{Score}(p, d) = \frac{f(p) \cdot f(d)}{|f(p)||f(d)|}
\end{equation}
where $f(x) = \text{PLM}(x) $ denotes the document embedding and PLM refers to a pre-trained language model such as SBERT. The document embedding can be derived by either averaging the token embeddings (mean pooling) or using the $[CLS]$ token embedding \cite{devlin2018bert}.

% Hongbo
% BM-25 and SBert

% Ensemble learning + Reflection, if necessary

Note that when evaluating RAG algorithms, we mainly use the F1 and ACC of LLM extraction result following the standard definition. The basic metrics of TP (true positive), FP (false positive), TN (true negative), FN (false negative) are first computed. The LLM extraction of each synthesis condition will be classified into one of TP/FP/TN/FN by comparing with the predefined ground-truth annotation, as described in the confusion matrix of \rfig{BestVSZero}(E).

\bsubsec{LLM Prompt Engineering for Material Knowledge Augmentation}{PromptEngineering}

\begin{table}[!t]
    \centering
    \caption{\textbf{Material knowledge incorporated as part of LLM background prompts.} Both the definition of 10 MOFs synthesis conditions and their numerical/textual/structural constraints are included.}
    \vspace{0.1 in}
    \includegraphics[width=\textwidth]{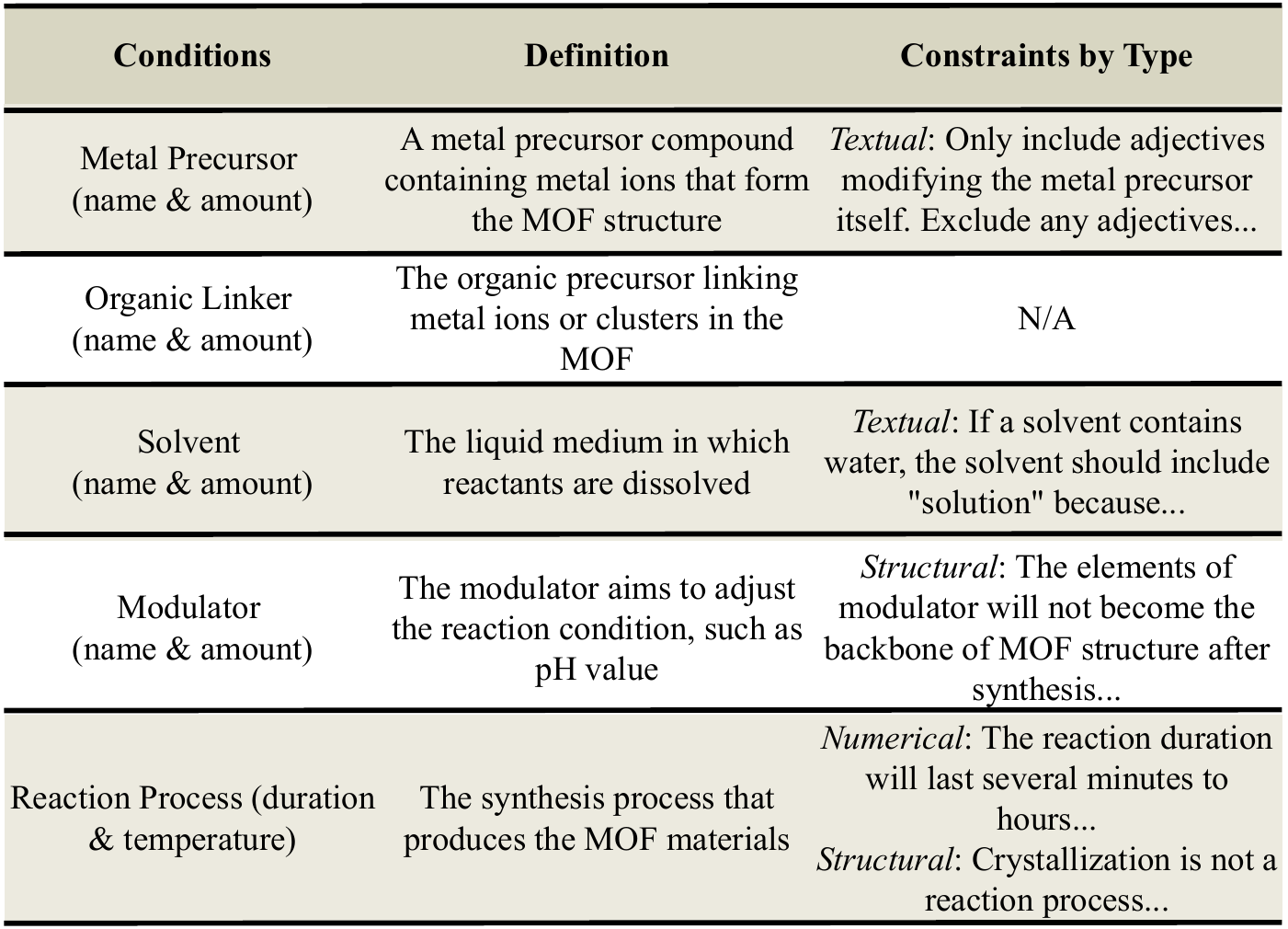}
    \label{tab:MaterialKnowledge}
\end{table}

% *** Read the current writing and record notes for discussion

% Background material knowledge and constraints

% impact of knowledge in experiment

% Lei & YangYi
%% 2
% background prompt
% static knowledge => Material constraints
% 有问题：based on latest prompt engineering.
% 参考:sammo。添加:gpt官方。

In addition to few-shot demonstrations, another way to augment the domain knowledge of general-purpose LLM is through the background prompt \cite{Dong2023ASF}. The previous LLM adaptation on MOFs synthesis extraction by Zheng et al. \cite{zheng2023chatgpt} introduces a preliminary prompt engineering approach, which includes the task description of MOFs synthesis extraction and the output format specification. In our work, based on the latest prompt engineering catalog \cite{white2023prompt}, we propose to further incorporate two types of material knowledge into the background prompt (\rappendfig{PromptEngineeringOurs}): definition of each MOF synthesis condition, and deterministic constraints on each condition's numerical/textual value or structure (if any). As shown in \rfig{BasicComparison}(D), by augmenting material knowledge, the macro-F1 metric increases from 0.91 to 0.93. However, when the few-shot examples are not incorporated, the background material knowledge will not lead to improvement by itself.

The list of newly introduced MOFs synthesis definitions and constraints as background prompts is provided in \rtab{MaterialKnowledge}. We summarize three types of constraints on synthesis conditions: \emph{numerical} that the value of a condition should fall into certain range according to prior knowledge, \emph{textual} that an extracted textual condition should adhere to certain format to speedup follow-up material application, and \emph{structural} that certain rules related to the condition are followed in all MOFs synthesis process. The full background prompt is provided in \rappendix{LLMPrompts}.

% Zhimeng
\bsubsec{Post-processing of Extracted Synthesis Routes}{PostProcessing}

The LLM-based method extracts the synthesis routes precisely as they appear in the literature. However, the same synthesis condition can have different writing styles. Post-processings can help to standardize these conditions so that the downstream material inference from synthesis route will have denser input data to achieve better performance.

\emph{Coreference Resolution}

% Weize
% proxy word in synthesis condition: "L", "H2L"

MOFs literature often use proxy words like ``L'' or ``H2L'' to represent specific organic linkers, a prototype called coreference in NLP. These proxy words are often defined before the synthesis paragraph in the same literature. Due to different writing styles, regular expression based extraction is not enough for resolving all the coreference of proxy words. We introduce a hybrid method combining LLM and regular expression: first, the synthesis paragraph is located in the literature and all the text before the paragraph is input to LLM. The LLM is asked to extract all anaphoric references and their original words. Second, a regular expression is designed to identify coreference proxy words from all the LLM-extracted organic linker conditions. Finally, the proxy words in the extracted synthesis route are matched with the detected anaphoric references. For each match, the proxy word is resolved into its original word. In all the 5269 synthesis paragraphs of the CSD-MOFs dataset, we detect 578 coreferences on the organic linker condition. 79\% of them can be resolved by our method. Only 2.3\% synthesis paragraphs still have unresolved proxy words in their organic linker condition.

\emph{Data Cleansing of Textual Conditions}

Textual synthesis conditions, such as metal, organic linker, solvent, and modulator names, often have different representations for the same substance, e.g., both ``$H_2O$" and ``Water" represent water. We propose a three-step post-processing approach to merge them together.

First, by similarity-based disambiguation, we generate a list of assimilated names to merge based on the similarity score between any two names $s_i$ and $s_j$:
\begin{equation}
		\text{Similarity} = 1 - \frac{L(s_i,s_j)}{max(|s_i|,|s_j|)}
\end{equation}
where $L(\cdot,\cdot)$ denotes the Levenshtein distance and $|\cdot|$ denotes the length of a name. By default, we use a similarity threshold of 0.9.

Second, we apply GPT-4 to conduct extended synonym merging. In this step, the synthesis condition is initially parsed by LLM to extract all chemical substance names. Next, a pre-defined LLM prompt is used to ask GPT-4 to recognize all the extracted substance names and group them into identical substances. Finally, a reflection prompt is used to re-evaluate all the synonym pairs. The result is applied to merge the condition names together.

%This method is suitable for synonym merging tasks in materials chemistry, such as Metal Source, Organic Linker, and Solvent. We merge data with frequency thresholds of 8/4/5 for Metal Source, Organic Linker, and Solvent, respectively, instead of merging all data. This reduces potential errors from merging low-frequency data and ensures fairness in subsequent comparisons.

Third, using regular expressions, we follow-up to clean and format data containing special characters (spaces, commas, etc.). By doing this, we ensure the integrity and usability of synthesis condition data.

\emph{Standardization of Numeric Conditions}

We also standardize numerical data in time and temperature conditions. These data can have quality issues such as inconsistent unit and presence of special characters. The following treatments are then performed. First, using regular expressions, we extract and format relevant data such as time and temperature conditions. Second, we define the standard unit and convert each synthesis condition into it. For example, time conditions are converted into hours, temperatures are converted into Celsius. Room temperature is converted into 25$^{\circ}$C uniformly.

\emph{Data Filtering by the Frequency of Synthesis Condition Values}

In the application of MOFs property inference, the number of unique values on some condition such as organic linkers can be quite high due to their long-tailed distribution, which can lead to poor inference performance. In our experiment, we focus on the MOFs synthesized by top conditions. Particularly, we filter the data to only leave those synthesized from top-100 metal sources, top-100 organic linkers, and top-20 solvents, all after synonym merging. This filter setting can be adjusted for different application scenarios.

\emph{Feature Embedding for Metal, Organic Linker, and Solvent}

The extracted synthesis conditions are embedded into vector representations for the follow-up machine learning, by augmenting their material/structural characteristics. The focused 10 synthesis conditions of a MOF are converted into a vector of length 206 by the following three steps.

First, using GPT-4, we obtain the chemical formulas and SMILES for the top-100 metal precursors and top-20 solvents of MOFs after synonym merging. For organic linkers, due to the complexity of their naming, GPT-4 can not obtain accurate SMILES. Therefore, we manually collect the SMILES for the top-100 organic linkers.

Second, based on the obtained SMILES, we use RDKit \cite{rdkit} to calculate the molecular features of metal precursors, organic linkers, and solvents, including their molecular weight, LogP values, the number of hydrogen bond donors and acceptors, Labute surface area, maximum molecular distance, molecular length, width, height, and topological polar surface area (TPSA).

Finally, using Pymatgen \cite{Ong2013} and Matminer \cite{WARD2018}, we calculate several chemical features of metal salts, including oxidation states, elemental properties, atomic orbitals, electron affinity, and electronegativity differences. Additionally, we include features of the metal element contained in the MOFs, such as its atomic mass, atomic radius, thermal conductivity, and detailed electronic configuration through vector representations.

\bsubsec{Bayesian Optimization for Best Synthesis Route Discovery}{Bayesian}

To locate the best synthesis route of a MOF for optimal target property, we apply Bayesian optimization algorithm. The algorithm depends on an acquisition function to select the best synthesis route which are likely to yield significant improvements in the target value while also exploring regions with high uncertainty. Three acquisition functions are applied, each of which selects a separate best synthesis route. Let $f(x)$ represent the target property value (such as SA of a MOF) synthesized by route $x$. We seek to maximize $f(x)$, given a set of known synthesis routes $x_i~(i=1, \cdots, N)$ and their target property values $f(x_i)$. As the Bayesian algorithm requires to quantify the uncertainty for each inference, we apply the random forest (RF) model to predict $f(x)$ given $N$ training data.

The first acquisition function, i.e., Expected Improvement (EI), quantifies the expected gain by evaluating the function at a new point in comparison to the current best observation.
\begin{equation}
\text{EI}(x) = \mathbb{E}\left[\max(0, f(x) - f(x^+))\right]
\end{equation}
where $f(x^+)$ represents the observed maximal value of the function and $f(x)$ is the predicted value at $x$. In addition, Probability of Improvement (PI) and Upper Confidence Bound (UCB) acquisition functions are defined by
\begin{equation}
\text{PI}(x) = \Phi \left( \frac{\mu(x) -  f(x^+)}{\sigma(x)} \right)
\end{equation}
\begin{equation}
\text{UCB}(x) = \mu(x) + \kappa \sigma(x)
\end{equation}
where $\mu$ and $\sigma$ denote the expectation and standard deviation of the prediction at $x$, $\Phi$ is the CDF of standard normal distribution, and $\kappa$ is a non-negative parameter balancing exploration and exploitation.

For all three acquisition functions, we create a grid of $T$ candidate synthesis route cells in the input space ($T =$4M in this work and we have also tried sparser grid settings). The candidate with the largest acquisition function is selected as the best synthesis route. In future, we plan to also experiment with iterative Bayesian optimization. After each iteration, the suggested best route will be sent for lab experiment. The actually synthesized material and its measured target value will be used to update the inference model and execute a next iteration of Bayesian optimization. 

% Research Articles and Reviews split the text into sections using headings
% Use a short (up 6 words) descriptive phrase, not generic 'Results' or 'Conclusions'
% Most other formats do not have headings, see the journal instructions to authors for details

%%%%%%%%%%%%%%%% REFERENCES %%%%%%%%%%%%%%%

\clearpage % Clear all remaining figures and tables then start a new page

% The list of references goes after the main text and before the acknowledgements
% When preparing an initial submission, we recommend you use BibTeX, like this:
%
\bibliography{LLM-SA} % for a file named science_template.bib
\bibliographystyle{sciencemag}

% After the paper has completed peer review and been revised ready for acceptance,
% you should comment out the lines above and copy-paste the contents of your .bbl
% file here instead. This will help ensure that our conversion software works correctly.
% Remember to re-run BibTeX first - check the timestamp!
%
% Example of the first three entries copy-pasted from science_template.bbl:
%
%\begin{thebibliography}{1}
%
%\bibitem{example}
%A.~N. {Author}, An example reference. \emph{Journal of Improbable Research}
%  \textbf{1}, 67 (2020).
%
%\bibitem{example2}
%F.~M. {Surname}, S.~{Author}, A second example. \emph{Interesting Research
%  Letters} \textbf{32}, 897 (2019).
%
%\bibitem{example_preprint}
%P.~{One}, P.~{Two}, P.~{Three}, {An unpublished preprint}. \emph{preprint}
%  (2021), arXiv:2101.12345.
%
%\end{thebibliography}

%%%%%%%%%%%%%%%% ACKNOWLEDGEMENTS %%%%%%%%%%%%%%%

\section*{Acknowledgments}
The authors would like to thank graduate students from School of Materials Science and Engineering, USTB, who helped to create the initial MOFs synthesis annotations on the CSD-MOFs dataset.

\paragraph*{Funding:}
National Key R\&D Program of China 2021YFB3500700 (LS, ZL, YY, WW, YZ, JL, SW, ZC, RL, NW, YL, ZL, HT, HG, GW)

NSFC Grant 62172026 (LS, YY, WW, ZC, NW, YL)

The Fundamental Research Funds for the Central Universities (LS, YY, WW, ZC, NW, YL)

State Key Laboratory of Complex \& Critical Software Environment (LS, YY, WW, ZC, NW, YL)

\paragraph*{Author contributions:}

Conceptualization: LS, GW

Machine learning and data analysis: ZL, YY, WW, HZ, ZC, NW, YL, YZ

System and visualization: YZ, SW, RL, ZL, HT

Material experiment: ZL, JL, HG, GW

Supervision: LS, HG, YZ, GW

Writing: LS and all co-authors

\paragraph*{Competing interests:} There are no competing interests to declare.
\paragraph*{Data and materials availability:}

Key data of this study are provided in the supplemental data file. The uploaded data include the extracted synthesis conditions and the annotation data for all the three MOFs datasets, as well as input and output data for MOFs density and SA value inference. Other supporting data of this study have been deposited at \url{https://github.com/BHT321/MOFs_Synthesis_Condition_Extraction/tree/main/Dataset}. Our method and the obtained large-scale MOFs synthesis data is also available as an online executable engine (\rappendfig{vis-panel-step1}) and database (\rappendfig{field_search}). \rappendix{VisualDatabase} and the supplemental video provide more details. All codes for LLM synthesis extraction and microstructure property inference are available at: \url{https://github.com/BHT321/MOFs_Synthesis_Condition_Extraction/tree/main/Code}.

%%%%%%%%%%%%%%%% SUPPLEMENT LIST %%%%%%%%%%%%%%%

% List the contents of your Supplementary Materials, including the numbers of any
% supplementary figures, tables, external data files etc. and any references that are
% cited only in the supplement. In this example, refs. 7-8 are cited only in the supplement.
% Fill out your numbers accordingly and delete any lines that aren't applicable.
\subsection*{Supplementary materials}
Supplementary Text\\
Figs. S1 to S19\\
Movie S1\\
Data S1

%%%%%%%%%%%%%%%% END OF MAIN TEXT %%%%%%%%%%%%%%%

\newpage

%%%%%%%%%%%%%%%% START OF SUPPLEMENT %%%%%%%%%%%%%%%

% Figures, tables, equations and pages in the supplement are numbered S1, S2 etc.
\renewcommand{\thefigure}{S\arabic{figure}}
\renewcommand{\thetable}{S\arabic{table}}
\renewcommand{\theequation}{S\arabic{equation}}
\renewcommand{\thepage}{S\arabic{page}}
\setcounter{figure}{0}
\setcounter{table}{0}
\setcounter{equation}{0}
\setcounter{page}{1} % not 0 as \newpage already started a supplementary page
% References continue the numbering from the main text.

%%%%%%%%%%%%%%%% SUPPLEMENT TITLE PAGE %%%%%%%%%%%%%%%

\begin{center}
\section*{Supplementary Materials for\\ \scititle}

% Author list for the supplement
% Indicate the corresponding authors, but do NOT include institutions here
% It would be nice if the template auto-generated this, but doing so is complicated...
% Author list for the supplement
% Indicate the corresponding authors, but do NOT include institutions here
Lei~Shi$^\dagger$,
Zhimeng~Liu$^\dagger$,
Yi~Yang,
Weize~Wu,
Yuyang~Zhang,
Hongbo~Zhang,
Jing~Lin,
Siyu~Wu,
Zihan~Chen,
Ruiming~Li,
Nan~Wang,
Yuankai~Luo,
Rui~Wang,
Zipeng~Liu,
Huobin~Tan,
Hongyi~Gao$^\ast$,
Yue~Zhang$^\ast$,
Ge~Wang$^\ast$\\ % Use \\ for a new line in this context

% Indicate corresponding authors and equal contribution
\small$^\ast$Corresponding author. Email: hygao@ustb.edu.cn, yue.zhang@wias.org.cn, gewang@ustb.edu.cn\\
\small$^\dagger$These authors contributed equally to this work.

\end{center}

% Fill out the numbers for each type of supplementary material,
% and delete any lines that aren't applicable.
% These are just example numbers that don't match the rest of this template.
\subsubsection*{This PDF file includes:}
Supplementary Text\\
Figs. S1 to S19\\
Caption for Movie S1\\
Caption for Data S1

\subsubsection*{Other Supplementary Materials for this manuscript:}
Movies S1\\
Data S1

\newpage

\subsection*{Supplementary Text}

\subsubsection*{MOFs Data}

% 1.Database retrieval and Format converting
\emph{CSD, WoS, and the retrieved datasets}

The primary data source of this work is the non-disordered MOF subset of Cambridge Structural Database (CSD) \cite{moghadam2017development} retrieved in June 2022 (v5.43), which lists 84,898 MOFs covering the bonding motifs of all common MOFs in CSD (CSD-MOFs). The entry of a MOF in the database related to this work includes the MOF structure in CIF format, its physical properties, a DOI linking to the relevant publication, and a unique MOF ID.

The dataset is pre-processed according to the goal of this work. First, the full-text describing the MOFs under study should be available. Out of all the 84,898 MOFs, 78,741 have non-empty DOIs. Since the same DOI could be linked to multiple MOFs (one paper reporting more than one MOF), this leaves 39,579 different DOI links after deduplication and 36,177 downloadable paper full-texts. For the convenience of follow-up processing, we focus on the DOIs where the associated publication reports the information of only one MOF in CSD. This leads to a subset of 22,461 MOFs, each with a unique publication file in PDF format.

Next, the PDF of each MOF is converted to plain text \cite{pdf2htmlEX} and segmented into paragraphs. The high-performance classification model in the main text is applied to detect synthesis paragraphs enclosing the desired synthesis condition information. Again, for the sake of convenience and accuracy, we only consider the 5,269 publications containing exactly one synthesis paragraph. Another 12,606 publications do not have any synthesis paragraph, probably because these papers are not related to MOFs experiments. The other 4,586 publications have more than one synthesis paragraph, as they are describing multiple MOFs or synthesis routes. Our pipeline could work with papers having more than one suite of synthesis conditions, but the potential MOF-synthesis mismatch may downgrade the application performance in evaluation. Therefore, throughout this work we stick to the core CSD-MOFs dataset with 5,269 publications/MOFs and their unique synthesis paragraphs.

To comprehensively evaluate our method and assess its application performance, we consider two additional data sources. The first focuses on the single type of UiO-66 MOF with the Zr metal. Synthesizing UiO-66 (Zr) MOFs using different conditions can result in varying structures and material properties. Utilizing the Web of Science platform~\cite{WoS}, we retrieved 4,524 unique research articles by searching the titles, abstracts, and keywords for ``Zr, BDC" or ``UiO, 66," which forms the WoS-UiO-66 dataset. These papers encompass studies on both the modification and synthesis of UiO-66. We then employed a LLM to extract the surface area value of each UiO-66 MOF from the paper full-text. From 918 papers, we successfully extracted the surface area value. As we focus on MOFs synthesis and surface area inference of UiO-66, only papers reporting the synthesis of pure UiO-66 (the MOF without modification) are considered. Finally, the papers having exactly one synthesis paragraph, describing pure UiO-66, and presenting its surface area value, count to 261 articles, which is the core WoS-UiO-66 dataset. The last dataset is the SIMM data provided by Zhang et al. \cite{zhang2024fine}. The SIMM dataset contains 600 MOFs collected in Zheng et al. \cite{zheng2023chatgpt}'s work, whose synthesis routes are manually re-annotated in Zhang et al. \cite{zhang2024fine} (as Zheng et al. \cite{zheng2023chatgpt} does not release the annotation data). Over the SIMM dataset, we keep 573 MOFs which has exactly one synthesis paragraph and one suite of synthesis route per MOF.

In addition, for material structure inference experiment, as the paper by Park et al. \cite{park2022mining} does not release the original annotated training data, we are not able to obtain their extraction result on our full dataset for comparison. We then consider the extraction output provided by their paper on a subset of CSD-MOFs data (the KAIST dataset, containing 46701 MOFs). A data joint is performed on the KAIST dataset and our dataset with 5269 MOFs, resulting 814 MOFs as the evaluation data. The same data post-processings are conducted on both extraction results.

\emph{Microstructure Property Computation}

For material evaluation purpose, we calculate structural and physical properties of the 5,269 MOFs in the core CSD-MOFs dataset. The CIF file of each MOF is retrieved from CSD and input to the Zeo++ tool \cite{willems2012algorithms}. In total, four structural and physical properties are calculated: global cavity diameter, pore limiting diameter, largest cavity diameter, and framework density. We set the probe radius to 1.29$\dot{A}$ to simulate helium gas molecules, and the number of Monte Carlo samples to 100,000 to ensure the accuracy of calculations. All Zeo++ parameters adhere to standard routines, guaranteeing that the computed properties accurately represent the behavior of gas molecules within the MOF structure.

% YangYi & Weize & Prof. Liu
\subsubsection*{Data Annotations for Synthesis Paragraph and Condition}

We invited eight experts on materials science and engineering to conduct the data annotations. Additionally, we developed a web-based interactive software to improve the efficiency of the annotation process (\rappendfig{AnnotationPlatform}). On the synthesis paragraph annotation task, as the task is relatively easy, we use human annotation only. On the synthesis condition annotation task, which is more difficult, we apply a new method of human-AI interactive data curation, as detailed in the main text.

\emph{Synthesis Paragraph Annotation}

440 papers were randomly selected from the CSD-MOFs dataset. Each paper was annotated by two different human experts, and in total 880 annotation tasks were assigned.  Experts used our software shown in \rappendfig{AnnotationPlatform} to annotate synthesis-related paragraphs. After annotation, only paragraphs agreed by both labelers were considered valid, and the other disagreeing paragraphs were discarded.

This process yielded 1,349 valid synthesis paragraph annotations. To train the binary classification model, non-synthesis paragraphs are needed as negative samples. After removing all paragraphs annotated as synthesis paragraphs from each paper, the remaining paragraphs serve as negative samples. This method resulted in 11,783 negative samples used for training the synthesis paragraph detection model on the CSD-MOFs dataset. The WoS-UiO-66 dataset is processed by the same annotation procedure.

\emph{Synthesis Condition Annotation}

On the CSD-MOFs dataset, we randomly selected 200 papers for synthesis condition annotation. The annotation process can be divided into five stages: task configuration, AI annotation, pilot annotation, batch annotation, and interactive data curation.

In the task configuration stage, domain experts define the key synthesis condition to be extracted and configure the annotation settings. The core objective is to maintain consistency for all ground-truth annotations. Due to the diversity of writing and representation styles in MOFs literature, consistent annotation helps to increase the frequency of major synthesis conditions, and therefore enhancing the effectiveness of follow-up machine learning based material inference. For example, we exclude MOF's activation conditions, which include activation temperature and activation time, after the pilot annotation process, as it is found that only few MOFs literature report activation conditions. Also, molecular formulas for all synthesis conditions are excluded for annotation because the condition name serves as better predictor in material inference.

In the next stage, the GPT-4 LLM is applied on each synthesis paragraph for an initial synthesis condition annotation, i.e., the AI annotation stage (\rfig{HumanAIReflection}(I)$\sim$(II)). The annotation task configuration and material domain knowledge is feed in as the background prompt for LLM. As there are currently no ground-truth annotations, LLM works in a zero-shot mode. Although the performance of initial AI annotation is limited, this stage helps to resolve the fatigue issue of human labeler when working with a large amount of annotation tasks.

On the initial AI annotation, human labelers are instructed to apply best-effort annotations (\rfig{HumanAIReflection}(II)$\sim$(III)). Here two annotation stages are designed to maximize the accuracy of human annotation. First, \emph{pilot annotation} on 20 papers of the CSD-MOFs dataset to validate and adjust the annotation task configuration. Each paper is annotated by two MOFs experts independently. The results on each paper are checked for their agreement between the two labelers. The expert labelers analyze and discuss any disagreement on the annotation, and resolve all the ambiguity and unclear issue during the annotation process. The pilot annotation stage also helps to optimize the design of our home-grown annotation software (\rappendfig{AnnotationPlatform}). Second, \emph{batch annotations} on the remaining 180 papers, handled by six human labelers in total. Again, each paper/paragraph is assigned two labelers for cross-check. After this annotation stage, the two labeling results on each paper are examined for intra-labeler agreement. For papers with higher degree of agreement than a pre-set threshold but not identical, another round of discussion is conducted to reach consensus on the final annotation. For papers with low degree of agreement, the paper/paragraph is simply discarded. The human annotation stages result in 147 papers with valid and consistent labels, and the other 53 papers are excluded for subsequent process.

Finally, we apply the human-AI interactive data curation (\rfig{HumanAIReflection}(III)$\sim$(IV)$\sim$(V)). First, the human annotation result is used as the demonstrations for a few-shot LLM extraction. Second, human labelers combine the few-shot LLM output and the last-stage human annotation result, known as the second-round human-AI data curation. In this way, both random errors induced by humans and common errors induced by AI due to the lack of specialized knowledge are largely resolved.

\subsubsection*{Synthesis Extraction and Structure Inference Performance on WoS-UiO-66 and SIMM Dataset}
%% intro of few-shot on uio66

Few-shot LLM, as the core of our technical pipeline, exhibits excellent performance across different MOFs datasets. Apart from CSD-MOFs dataset reported in the main text, we also present additional results on another dataset, i.e., WoS-UiO-66 and SIMM dataset.

%% prompt design
\emph{Prompt Engineering for Synthesis Extraction of UiO-66 MOFs}

Similar to the synthesis extraction of CSD-MOFs dataset, we augment the domain knowledge of LLM through the background prompt. Due to differences between the datasets, prompts are modified to better meet the requirements. In the prompt, we explicitly require the LLM to focus on the synthesis of pure UiO-66 (also known as Zr-BDC) and ignore the other synthesis procedures like drying or crystallization process. Furthermore, we check whether our constraints are necessary for synthesis extraction of UiO-66. A preliminary extraction of UiO-66 synthesis conditions indicates that the LLM rarely makes errors in word sequence or modulator identification. Therefore, we remove some unnecessary constraints from our previous prompt, except for the clarification of synthesis and crystallization processes.

%% uio66 performance
%% result, detailed number of result
%% defference between zero-shot and few-shot
\emph{Synthesis Extraction Performance on WoS-UiO-66 Dataset}

As mentioned in the main text, the few-shot examples used for our task are shown to be a critical factor to the synthesis extraction performance. Therefore, 87 UiO-66 MOFs with annotated paragraphs and synthesis conditions are used as examples. For synthesis extraction of UiO-66 MOFs, we employ the most efficient BM25 algorithm and a setting of $K=6$ for the few-shot in-context learning.

In the comparison result of \rappendfig{UiO66BestVSZero}, it is shown that the few-shot LLM (\rappendfig{UiO66BestVSZero}(A)) achieves much higher macro-F1 (0.93) and ACC (0.90) than the zero-shot LLM (\rappendfig{UiO66BestVSZero}(B)), which achieves a macro-F1 score of 0.76 and ACC of 0.71. This result is similar to the extraction performance reported in the main text on the CSD-MOFs dataset. There are two detailed deficiency of the zero-shot LLM method. The first is that the zero-shot method can not distinguish between solvent and modulator, and it tends to classify some modulators (e.g., HAc or HCl) as solvents. The second is that the zero-shot method can not extract all the units of conditions. For example, if an amount is written as ``30mg, 0.1mmol'', the LLM may extract either ``30mg'' or ``0.1mmol'', but not both. These two deficiencies of zero-shot method are addressed by the few-shot LLM by introducing domain-specific material knowledge. This demonstrates the universal advantage of our few-shot LLM method over the zero-shot LLM.

On the WoS-UiO-66 dataset, we also compare different RAG algorithms and the choice of number of shots ($K$). As shown in \rfig{UiO66ShotsComparison}, the blue lines with circle symbols represent the performance variation with different $K$s in our task, using the best BM25 RAG algorithm. Both F1 and ACC improve most significantly from zero-shot to one-shot, with an increase of more than 0.1. This improvement is greater than the results on the CSD-MOFs dataset, possibly because of the lack of specialized knowledge on UiO-66 in the LLM. Provided with high-quality UiO-66 material knowledge, LLM is able to achieve the same extraction performance obtained on the CSD-MOFs dataset. The performance continues to improve until $K=6$. After this peak (F1=0.93, ACC=0.90), the metrics fluctuate without surpassing the best performance. Meanwhile, the few-shot method with random example selection (orange lines), BERT algorithm (green lines) and SBERT (red lines) exhibit the same trend as $K$ increases. The performance of both BERT and random example selection consistently fall behind that of the BM25 algorithm, while SBERT and BM25 perform almost identically. The variation among these four algorithms is smaller than the results on the CSD-MOFs dataset, possibly due to the higher similarity among different UiO-66 synthesis routes. Because a lot of synthesis conditions used for UiO-66 synthesis are usually the same, especially the precursor names, the overall synthesis context and paragraphs are highly similar to each other. Therefore, retrieving different synthesis paragraphs as examples does not significantly impact the synthesis extraction performance on the WoS-UiO-66 MOFs dataset.

\emph{Structure Inference Performance on UiO-66}

We also design a material inference task to compare the performance of few-shot vs. zero-shot synthesis extraction methods. A key structure property of MOFs, i.e., the surface area, is inferred in this task, using the extracted synthesis conditions including metals, organic linkers, solvents, modulators, and reaction duration/temperature. The input data is the extraction result on the WoS-UiO-66 dataset. They are post-processed using the same approach applied to the CSD-MOFs dataset. After filtering out sparsely appearing synthesis conditions, we have a smaller dataset in both few-shot and zero-shot extraction result, with 151 and 85 MOFs respectively. A 19-dimensional embedding vector is obtained for each MOF's synthesis conditions.

The random forest model is selected for the inference as it supports quantifying the uncertainty of prediction. We compare few-shot LLM and zero-shot LLM in a 10-fold cross-validation. The inference performance is reported by the coefficient of determination ($R^2$). It is shown that the synthesis conditions obtained by the few-shot method exhibit significantly higher predictive power than those obtained by the zero-shot method. The $R^2$ value of few-shot is 0.1404 while the zero-shot is 0.0751. This inference performance is lower than the prediction of MOF's microstructural property on the CSD-MOFs dataset, mainly because we have less data on UiO-66. The inference model is further utilized in suggesting the best synthesis conditions for optimal surface area value.

\emph{Synthesis Extraction Performance on SIMM Dataset}

In addition to the WoS-UiO-66 dataset, we also compare the performance of Few-shot RAG algorithm with zero-shot LLM extraction on SIMM dataset. We apply the Human-AI annotation process on the origin SIMM training and testing datasets, resulting in 573 MOFs for further extraction. For the synthesis extraction of SIMM MOFs, we employ the BM25 algorithm with a setting of $K=4$ for few-shot in-context learning.

As shown in the comparison results of \rappendfig{SIMMBestVSZero}, the few-shot LLM achieves significantly higher macro-F1 (0.96) and ACC (0.94) compared to the zero-shot LLM, which achieves a macro-F1 score of 0.74 and ACC of 0.70.

% Yuyang & Siyu & Prof. Liu
\subsubsection*{MOFs Synthesis Extraction Engine, Database, and Visualization}

We developed an online system using the method proposed in this paper, which is capable of automatically executing the LLM-based synthesis extraction and supporting the visual analysis and database retrieval of the extraction result on the CSD-MOFs dataset.

\emph{Online MOFs Synthesis Extraction Engine}

To streamline the entire synthesis extraction workflow, allow easy access to our method, and support online usage of the engine over any related literature, we developed the MOFs Synthesis Condition Extraction Engine.

The engine is developed using a frontend-backend separation architecture. The frontend follows the MVC design pattern and is built using the well-established open-source Vue 3 framework, TypeScript scripting language, and the Vuetify component library. Data is retrieved by sending Axios requests to the backend. The backend adopts a layered architecture and leverages the Spring Boot 3 framework to handle requests. High-performance paragraph synthesis extraction models and synthesis condition extraction models are exposed via FastAPI endpoints, and the backend communicates with these services using FeignClient to send requests and process the returned results.

The system utilizes MySQL as the relational database and MinIO for file storage services. All data related to the synthesis extraction workflow is stored in the databases. The data manipulation and connection operations are managed via Spring.

\emph{MOFs Synthesis Database}

Using the synthesis paragraph detection algorithm, we processed 36,177 papers from the CSD-MOFs dataset and extracted 57,081 synthesis paragraphs, on which we then executed the proposed few-shot LLM extraction method. The large amount of extraction results are stored in the MySQL database. Figure \ref{fig:basic_statistics} gives the basic statistics of the database. The database is mainly composed of tables on the 10 extracted synthesis conditions and the original synthesis paragraph, mostly with over 10k records. We implemented the faceted search capability on the database. As shown in \rappendfig{field_search}, the visual user interface allows to search on any retrieved synthesis conditions as well as the original literature metadata and the synthesis paragraph content. \rfig{basic_search} gives the page of search result upon the basic search. By the Elasticsearch technology, the system also supports advanced search that combines multiple queries via boolean logic operators (\rfig{advanced_search}). Typical advanced search result visualization is given in \rfig{boolean_search1}.

% Siyu
\emph{User Interface for Online MOFs Synthesis Extraction and LLM Output Visualization}

% step fig*4
We integrate the entire workflow of MOFs synthesis extraction from scientific literature into the same interactive user interface. The user operation on the interface is composed of the following steps:

%-------------- Step 1 --------------
%\emph{Literature Upload and File Conversion}

First, users upload one or multiple papers related to the MOFs synthesis to our system (\rappendfig{vis-panel-step1}). The paper files will be automatically converted into textual file format suitable for the follow-up synthesis condition extraction.

%-------------- Step 2 --------------
%\emph{Synthesis Paragraph Detection}

Second, after uploading and converting paper files, the system will automatically detect synthesis paragraphs from all the uploaded papers. A literature status list is displayed to show the state of synthesis paragraph detection on all uploaded papers (Figure \ref{fig:vis-panel-step2}).

%-------------- Step 3 --------------
% \emph{Synthesis Condition Extraction}

Third, users can view the detected synthesis paragraphs and configure the algorithm parameters for the follow-up synthesis condition extraction (Figure \ref{fig:vis-panel-step3}). At the top of the interface, a drop-down list allows to select a specific paper from the current batch. The left panel will displays a preview of the selected literature. The central panel shows the detected synthesis paragraphs, allowing users to choose a few paragraphs to process. In the right part, a configuration panel allows users to set the current algorithm parameters for the synthesis extraction, which include the choice of LLM model and RAG algorithm, as well as the number of few-shots ($K$).

%-------------- Step 4 --------------
\emph{LLM Output Visualization}

% 3 tabs of vis interface

%% -------------- Step 4 :: Tab 1 --------------

The system supports visualization and interactive analysis of the synthesis extraction result given by LLM. In the visualization interface shown in \rappendfig{vis-panel-tab1}, (A) gives the four overall performance metrics for the synthesis extraction. (B) expands the performance metrics on each of the 10 synthesis conditions, using a heat map display. (C) provides the detailed view of synthesis condition extraction, showing both the extracted entities and the ground-truth annotations for comparison.

The interface can be switched to the second tab view (Figure \ref{fig:vis-panel-tab2}) to drill down to more detailed performance on each uploaded literature. In a third tab view (Figure \ref{fig:vis-panel-tab3}), the 2D projection visualization shows the distribution of extracted 10 synthesis conditions. Here, red dots indicate newly extracted suites of synthesis conditions. Blue dots represent the corresponding ground-truth synthesis conditions by annotation. Black dots indicates the other synthesis conditions already in the database, showing a typical synthesis condition distribution.
%% -------------- Step 4 :: Tab 2 --------------

%% -------------- Step 4 :: Tab 3 --------------

%-------------- Database retrieval --------------
\emph{Database Retrieval}

The system also supports the visualization of database retrieval. As shown in Figure \ref{fig:vis-panel-DB}, users can view the distribution of retrieved synthesis conditions through a 2D scatterplots. The detailed information of each synthesis condition is displayed in the list below.

\FloatBarrier
% LLM Prompt used in this work
\subsubsection*{LLM Prompts For Synthesis Extraction}

% Original background prompt used by Yahgi's team

% Our background prompt

% The two version of prompts are shown by pictures. Little text is required to describe the pictures. The description will focus on the difference

\emph{Introduction of LLM Prompt Structure}

This section presents the structure and specific content of the prompts used for synthesis condition extraction. In our implementation, we use the OpenAI API as the GPT calling tool, which differs slightly in its display and usage compared to the web version. Without otherwise noted, we use the GPT-4 turbo model.

An example of a one-shot prompt for extracting MOFs is shown below:
\begin{tcolorbox}[title=Prompt Demonstration, colback=blue!5!white, colframe=blue!75!black, breakable]
\textbf{Prompt:}\\
I require your assistance in efficiently and precisely extracting synthesis parameters from chemical literature...
\newline

\textbf{Sample Input 1:}\\
2.2.1. Synthesis of \lbrack Cu I (4,4' -bpy)] 2 \lbrack H 2 SiW 12 O 40 ] $\cdot$ 3H 2 O (1). H 4 SiW 12 O 40 $\cdot$ 18H 2 O \lbrack 28] (1.15 g, 0.356 mmol), Cu(Ac) 2 $\cdot$ 2H 2 O (0.16 g, 0.7 mmol), 4,4' -bipy (0.05 g, 0.32 mmol), and en (0.25 g, 4.16 mmol) were dissolved in 10 mL water...\\

\textbf{Sample Output 1:}
\begin{verbatim}
[
  {
    "Compound_Name": [
      "[Cu I (4,4' -bpy)] 2 [H 2 SiW 12 O 40 ] · 3H 2 O"
    ],
    "Metal_Source": [
      {
        "precursor_name": "H 4 SiW 12 O 40 · 18H 2 O",
        "amount": "1.15 g, 0.356 mmol"
      },
      {
        "precursor_name": "Cu(Ac) 2 · 2H 2 O",
        "amount": "0.16 g, 0.7 mmol"
      }
    ],
...
\end{verbatim}
\vspace{1em}
\hrule
\vspace{1em}
\textbf{Input:}\\
Synthesis of [Cd(L)(BIB) 0.5 ] n  (1) A mixture of H 2 L (0.2703 g, 1 mmol), Cd(OAc) 2  $\cdot$ 2H 2 O (0.2673 g, 1 mmol), NaOH (0.0801 g, 2 mmol) and BIB (0.189 g, 1 mmol) in 10 mL of distilled water was sealed in a 23 mL vial, and then heated to 150 $^\circ$C for 5 days.
\vspace{1em}
\hrule
\vspace{1em}
\textbf{Output:}
\begin{verbatim}
[
  {
    "Compound_Name": [
      "[Cd(L)(BIB) 0.5 ] n"
    ],
    "Metal_Source": [
      {
        "precursor_name": "Cd(OAc) 2 · 2H 2 O",
        "amount": "0.2673 g, 1 mmol"
      }
    ],
...
\end{verbatim}
\end{tcolorbox}

% The design of the prompt section follows the latest relevant studies and official guidelines; any unique content is described separately.

The above box illustrates the workflow of our GPT-based extraction process, divided into five sections: ``Prompt,'' ``Sample Input,'' ``Sample Output,'' ``Input,'' and ``Output.'' The ``Prompt'' describe the task for the model and provide necessary information. The ``Sample Input'' and ``Sample Output'' provide examples of input texts and the desired output format, which help guide the model's responses. The ``Input'' section contains the actual synthesis paragraph to be processed, and the ``Output'' section presents the GPT-generated extraction results.

The prompt structure when programmatically invoking GPT differs from that of the web version, and an example implementation is shown in the box below, which displays the actual code used for invocation and the method for obtaining responses. The complete code can be found on GitHub: \url{https://github.com/passingby000/MOFs_Synthesis_Condition_Extraction}

\begin{tcolorbox}[title=GPT Calling Code in Python, colback=gray!10!white, colframe=gray!75!black, breakable]
\begin{verbatim}
response = client.chat.completions.create(
  model="gpt-4-turbo",
  messages=[
    {"role": "system", "content": Prompt},
    {"role": "user", "content": Sample_Input_1},
    {"role": "assistant", "content": Sample_Output_1},
    {"role": "user", "content": Input}
  ],
  temperature=0
)

Output = response.choices[0].message.content.strip()
return Output
\end{verbatim}
\end{tcolorbox}

For clearer and more intuitive demonstration, the prompt figures use ``Prompt'' to represent the ``system'' in code, ``Sample Input 1'' to indicate the first ``user'', ``Sample Output 1'' to indicate the first ``assistant''. Each ``user-assistant'' pair acts as a sample input/output pair. The prompt figure use ``Input'' to represent the final ``user'' containing the synthesis paragraph for extraction, from which the model generates an output. Models are set according to the task, and temperature is set to 0 to keep the stability of output. The GPT-API ``Output'' is retrieved and parsed from the API model response.

Our prompt structure, depicted in \rappendfig{PromptEngineeringOurs}, consists of five functional components, including role establishment, task definition, background, mission details, and structure sample. For brevity, some specific content has been omitted. The rationale behind our prompt design is detailed in the subsequent sections of the main text. The design references several related works \cite{zheng2023chatgpt}, including the official OpenAI guidelines \cite{openai-prompt-engineering}, and incorporates the latest paradigms in prompt engineering \cite{white2023prompt, bsharat2023principled, sahoo2024systematic}. As a comparative study, the prompt structure used by Zheng et al. is shown in \rfig{PromptEngineeringBerkeley}. We have similarly adopted their method for eliminating hallucinations and added background knowledge and constraints specific to our task to enhance the model's performance on the CSD-MOFs and WoS-UiO-66 datasets.

In summary, the carefully designed prompt structure and API interaction are crucial for accurately and efficiently extracting synthesis conditions of MOFs from chemical literature using GPT models.

\emph{Prompts for Zero-Shot LLM in CSD-MOF Experiment}

Zero-shot LLM extraction was conducted over the 5,269 filtered synthesis paragraphs from the CSD-MOFs dataset, using the GPT-4 API. The extracted synthesis conditions were then cleaned and utilized in subsequent experiments.

The following table presents the prompts, processes, and sample results used in this extraction task. The DOI of following condition extraction example paper is \url{10.1134/S1063774518040296} .

\begin{tcolorbox}[title=Zero-shot Prompts for CSD-MOF, colback=blue!5!white, colframe=blue!75!black, breakable]
\textbf{Prompt:}\\
You are a chemical expert with 20 years of experience in reviewing literature and extracting key information. Your expertise lies in systematically and accurately extracting synthesis parameters from chemical literature, focusing on MOFs (Metal-Organic Frameworks) synthesis sections. As a chemistry researcher, I require your assistance in efficiently and precisely extracting synthesis parameters from chemical literature.
\newline \\
Your task is to summarize the following details for a JSON format table from some input: `Compound\_Name', `Metal\_Source', `Organic\_Linker', `Solvent', `Modulator', `Reaction\_Time', `Reaction\_Temperature'. Among them, `Metal\_Source', `Organic\_Linker', `Solvent', and `Modulator' should also contain their amounts.\\
Only focus on the MOF synthesis process. Ignore other processes, including the organic precursor synthesis pre-process and the active and crystallization post-process, if mentioned in the paragraph.\\
\newline \\
The detailed format descriptions for each class are below: \\
The output should be a JSON table list. Each JSON format table represents a MOF.\\
If there is only one MOF, the JSON list should only have one JSON format table. If there is more than one MOF, they should be put in different JSON tables.\\
Each JSON format table should contain: ``Compound\_Name'', ``Metal\_Source'', ``Organic\_Linker'', ``Solvent'', ``Modulator'', ``Reaction\_Time'', ``Reaction\_Temperature''.\\
The ``Compound\_Name'', ``Reaction\_Time'', ``Reaction\_Temperature'' should be lists of strings.\\
The ``Metal\_Source'', ``Organic\_Linker'', ``Solvent'', and ``Modulator'' should all be lists of dicts. The dict should have the keys ``precursor\_name'' and ``amount''. ``amount'' includes weight, volume, or molar weight, presented according to the paragraph text, but not including the concentration, proportion, or rate.\\
The ``Compound\_Name'' should contain the full MOF chemical name that has been synthesized according to the paragraph text. Extract the suffix like ``\}n'' or ``]n'', because it represents the structure of the compound, acting as part of the chemical name. Remove the prefix which is not part of the chemical MOF name like ``compound'' or ``complex'' and the suffix like ``(1)'' or ``(2)''.\\
Extract the name and amount of ``Metal\_Source'', ``Organic\_Linker'', ``Solvent'', and ``Modulator'' during the synthesis process of MOF according to the paragraph, and put them in the dict with the keys ``precursor\_name'' and ``amount''.\\
The ``Reaction\_Time'', ``Reaction\_Temperature'' should be extracted from the paragraph. If there is no time or temperature in the text, set the value to be an empty list.\\
If some values are missing, not presented, or not mentioned, keep them empty in the output JSON text. The output should only contain the list.
\newline \\
The output structure should be like this:
\begin{verbatim}
[
  {
    "Compound_Name": [],
    "Metal_Source": [
      {
        "precursor_name": "",
        "amount": ""
      }
    ],
    "Organic_Linker": [
      {
        "precursor_name": "",
        "amount": ""
      },
      {
        "precursor_name": "",
        "amount": ""
      }
    ],
    "Solvent": [
      {
        "precursor_name": "",
        "amount": ""
      },
      {
        "precursor_name": "",
        "amount": ""
      }
    ],
    "Modulator": [
      {
        "precursor_name": "",
        "amount": ""
      }
    ],
    "Reaction_Time": [],
    "Reaction_Temperature": []
  }
]
\end{verbatim}
\vspace{1em}
\hrule
\vspace{1em}
\textbf{Input:}\\
Synthesis of \lbrack Cd(L)(BIB)0.5\rbrack n (1) A mixture of H\textsubscript{2}L (0.2703 g, 1 mmol), Cd(OAc)\textsubscript{2}\textperiodcentered 2H\textsubscript{2}O (0.2673 g, 1 mmol), NaOH (0.0801 g, 2 mmol) and BIB (0.189 g, 1 mmol) in 10 mL of distilled water was sealed in a 23 mL vial, and then heated to 150$^\circ$C for 5 days. Colorless block crystals of 1 were collected in 56\% yield based on Cd. Anal. calcd. (\%) for C\textsubscript{20}H\textsubscript{15}CdN\textsubscript{2}O\textsubscript{5}: C, 50.49; H, 3.18; N, 5.89. Found (\%): C, 50.45; H, 3.15; N, 5.82.
\vspace{1em}
\hrule
\vspace{1em}
\textbf{Output:}
\begin{verbatim}
[
  {
    "Compound_Name": [
      "[Cd(L)(BIB)0.5]n"
    ],
    "Metal_Source": [
      {
        "precursor_name": "Cd(OAc)2·2H2O",
        "amount": "0.2673 g"
      }
    ],
    "Organic_Linker": [
      {
        "precursor_name": "H2L",
        "amount": "0.2703 g"
      },
      {
        "precursor_name": "BIB",
        "amount": "0.189 g"
      }
    ],
    "Solvent": [
      {
        "precursor_name": "distilled water",
        "amount": "10 mL"
      }
    ],
    "Modulator": [
      {
        "precursor_name": "NaOH",
        "amount": "0.0801 g"
      }
    ],
    "Reaction_Time": [
      "5 days"
    ],
    "Reaction_Temperature": [
      "150°C"
    ]
  }
]
	\end{verbatim}
\end{tcolorbox}

\emph{Prompts for Few-Shot LLM in CSD-MOF Experiment}

Few-shot LLM extraction was conducted over the 5,269 filtered synthesis paragraphs from the CSD-MOFs dataset, using the GPT-4 API. To select 3 relevant examples from 137 annotations, we employed the BM25 algorithm. These examples were then used to extract synthesis conditions. To specifically improve extraction quality, background information and constraints based on domain knowledge were added to the prompts. The results were subsequently cleaned and utilized in subsequent experiments.

The following table presents the prompts, examples, processes, and sample results used in this extraction task. The DOI of following condition extraction example paper is \url{10.1134/S1063774518040296} .

\begin{tcolorbox}[title=Few-shot Prompts for CSD-MOF, colback=blue!5!white, colframe=blue!75!black, breakable]
\textbf{Prompt:}\\
You are a chemical expert with 20 years of experience in reviewing literature and extracting key information. Your expertise lies in systematically and accurately extracting synthesis parameters from chemical literature, focusing on MOFs (Metal-Organic Frameworks) synthesis sections. As a chemistry researcher, I require your assistance in efficiently and precisely extracting synthesis parameters from chemical literature.
\newline \\
Your task is to summarize the following details for a JSON format table from some input: `Compound\_Name', `Metal\_Source', `Organic\_Linker', `Solvent', `Modulator', `Reaction\_Time', `Reaction\_Temperature'. Among them, `Metal\_Source', `Organic\_Linker', `Solvent', and `Modulator' should also contain their amounts.\\
Only focus on the MOF synthesis process. Ignore other processes, including the organic precursor synthesis pre-process and the active and crystallization post-process, if mentioned in the paragraph.
\newline \\
Background Information and Detailed Instructions: \\
Compound\_Name of MOFs (Metal-Organic Frameworks): MOFs are porous materials formed by the coordination of metal ions or clusters with organic ligands. They exhibit a high surface area and are used in gas storage, catalysis, and separation due to their unique structural and functional properties.\\
Metal\_Source: In MOF synthesis, a Metal Source is a precursor compound containing metal ions that form part of the MOF structure. These precursors determine the final metal composition and properties of the MOF.\\
Organic\_Linker: Refers to the organic precursor molecule linking metal ions or clusters in the MOF, influencing the framework's topology, porosity, and functionality.\\
Solvent: The liquid medium in which reactants are dissolved.\\
Modulator: The modulator aims to adjust the reaction condition, such as PH value.\\
Reaction process: The synthesis process that produces the MOF materials.
\newline \\
The detailed format descriptions for each class are below: \\
The output should be a JSON table list. Each JSON format table represents a MOF.\\
If there is only one MOF, the JSON list should only have one JSON format table. If there is more than one MOF, they should be put in different JSON tables.\\
Each JSON format table should contain: ``Compound\_Name'', ``Metal\_Source'', ``Organic\_Linker'', ``Solvent'', ``Modulator'', ``Reaction\_Time'', ``Reaction\_Temperature''.\\
The ``Compound\_Name'', ``Reaction\_Time'', ``Reaction\_Temperature'' should be lists of strings.\\
The ``Metal\_Source'', ``Organic\_Linker'', ``Solvent'', and ``Modulator'' should all be lists of dicts. The dict should have the keys ``precursor\_name'' and ``amount''. ``amount'' includes weight, volume, or molar weight, presented according to the paragraph text, but not including the concentration, proportion, or rate.\\
The ``Compound\_Name'' should contain the full MOF chemical name that has been synthesized according to the paragraph text. Extract the suffix like ``\}n'' or ``]n'', because it represents the structure of the compound, acting as part of the chemical name. Remove the prefix which is not part of the chemical MOF name like ``compound'' or ``complex'' and the suffix like ``(1)'' or ``(2)''.\\
Extract the name and amount of ``Metal\_Source'', ``Organic\_Linker'', ``Solvent'', and ``Modulator'' during the synthesis process of MOF according to the paragraph, and put them in the dict with the keys ``precursor\_name'' and ``amount''.\\
The ``Reaction\_Time'', ``Reaction\_Temperature'' should be extracted from the paragraph. If there is no time or temperature in the text, set the value to be an empty list.\\
If some values are missing, not presented, or not mentioned, keep them empty in the output JSON text. The output should only contain the list.
\newline \\
The output structure should be like this:
\begin{verbatim}
[
  {
    "Compound_Name": [],
    "Metal_Source": [
      {
        "precursor_name": "",
        "amount": ""
      }
    ],
    "Organic_Linker": [
      {
        "precursor_name": "",
        "amount": ""
      },
      {
        "precursor_name": "",
        "amount": ""
      }
    ],
    "Solvent": [
      {
        "precursor_name": "",
        "amount": ""
      },
      {
        "precursor_name": "",
        "amount": ""
      }
    ],
    "Modulator": [
      {
        "precursor_name": "",
        "amount": ""
      }
    ],
    "Reaction_Time": [],
    "Reaction_Temperature": []
  }
]
\end{verbatim}
\textbf{Sample Input 1:}\\
2.2.1. Synthesis of \lbrack Cu I (4,4' -bpy)] 2 \lbrack H 2 SiW 12 O 40 ] $\cdot$ 3H 2 O (1). H 4 SiW 12 O 40 $\cdot$ 18H 2 O \lbrack 28] (1.15 g, 0.356 mmol), Cu(Ac) 2 $\cdot$ 2H 2 O (0.16 g, 0.7 mmol), 4,4' -bipy (0.05 g, 0.32 mmol), and en (0.25 g, 4.16 mmol) were dissolved in 10 mL water. The pH of the resulting mixture was adjusted to 2.7 by adding HCl solution, and then the mixture was stirred for 30 min in air. The final solution was transferred into a 25 mL Teflon-lined autoclave at 170 $^\circ$C for 3 days. Then the autoclave was cooled at 10 $^\circ$Ch 1 to room temperature. The resulting dark red block crystals were filtered off, washed with distilled water, and dried in a desiccator at room temperature (38\% yield based on W).
\newline \\
\textbf{Sample Output 1:}
\begin{verbatim}
[
  {
    "Compound_Name": [
      "[Cu I (4,4' -bpy)] 2 [H 2 SiW 12 O 40 ] · 3H 2 O"
    ],
    "Metal_Source": [
      {
        "precursor_name": "H 4 SiW 12 O 40 · 18H 2 O",
        "amount": "1.15 g, 0.356 mmol"
      },
      {
        "precursor_name": "Cu(Ac) 2 · 2H 2 O",
        "amount": "0.16 g, 0.7 mmol"
      }
    ],
    "Organic_Linker": [
      {
        "precursor_name": "4,4' -bipy",
        "amount": "0.05 g, 0.32 mmol"
      },
      {
        "precursor_name": "en",
        "amount": "0.25 g, 4.16 mmol"
      }
    ],
    "Solvent": [
      {
        "precursor_name": "water",
        "amount": "10 mL"
      }
    ],
    "Modulator": [
      {
        "precursor_name": "HCl solution",
        "amount": ""
      }
    ],
    "Reaction_Time": [
      "3 days"
    ],
    "Reaction_Temperature": [
      "170 °C"
    ]
  }
]
\end{verbatim}
\textbf{Sample Input 2:}\\
2.2. Preparation of {\textmu}-o-[Ni 2 (4,4' -bpy) 2 (oba) 2 ] n (1) NiSO 4 6H 2 O (0.1314 g, 0.5 mmol), H 2 oba (0.1291 g, 0.5 mmol) and 4,4 ' -bipyridine (0.0781 g, 0.5 mmol) were dissolved in 10 mL distilled water, 150 mL triethylamine was added slowly to the mixture upon stirring, after further 20 min continuous stirring the final mixture was sealed in a 23 mL Teflon-lined stainless vessel, and heated at 160 $^\circ$C for 72 h.
\newline \\
\textbf{Sample Output 2:}
\begin{verbatim}
[
  {
    "Compound_Name": [
      "{\textmu}-o-[Ni 2 (4,4' -bpy) 2 (oba) 2 ] n"
    ],
    "Metal_Source": [
      {
        "precursor_name": "NiSO 4 · 6H 2 O",
        "amount": "0.1314 g, 0.5 mmol"
      }
    ],
    "Organic_Linker": [
      {
        "precursor_name": "H 2 oba",
        "amount": "0.1291 g, 0.5 mmol"
      },
      {
        "precursor_name": "4,4 ' -bipyridine",
        "amount": "0.0781 g, 0.5 mmol"
      }
    ],
    "Solvent": [
      {
        "precursor_name": "water",
        "amount": "10 mL"
      }
    ],
    "Modulator": [
      {
        "precursor_name": "triethylamine",
        "amount": "150 mL"
      }
    ],
    "Reaction_Time": [
      "72 h"
    ],
    "Reaction_Temperature": [
      "160 °C"
    ]
  }
]
\end{verbatim}
\textbf{Sample Input 3:}\\
Synthesis of {[Ag(H 3 bptc)(bpe)]$\cdot$2H 2 O} n (1) Compound 1 was obtained by reaction of AgNO 3 (0.2 mmol), bpe (0.05 mmol) and H 4 bptc (0.05 mmol) in 4:1:1 molar ratio in 15 mL of water and 1 mL of triethylamine under hydrothermal conditions (at 150 $^\circ$C for 6 days and cooled to room temperature with a 5 $^\circ$Ch 1 rate).
\newline \\
\textbf{Sample Output 3:}
\begin{verbatim}
[
  {
    "Compound_Name": [
      "{[Ag(H 3 bptc)(bpe)] · 2H 2 O} n"
    ],
    "Metal_Source": [
      {
        "precursor_name": "AgNO 3",
        "amount": "0.2 mmol"
      }
    ],
    "Organic_Linker": [
      {
        "precursor_name": "bpe",
        "amount": "0.05 mmol"
      },
      {
        "precursor_name": "H 4 bptc",
        "amount": "0.05 mmol"
      }
    ],
    "Solvent": [
      {
        "precursor_name": "water",
        "amount": "15 mL"
      },
      {
        "precursor_name": "triethylamine",
        "amount": "1 mL"
      }
    ],
    "Reaction_Time": [
      "6 days"
    ],
    "Reaction_Temperature": [
      "150 °C"
    ]
  }
]
\end{verbatim}
\vspace{1em}
\hrule
\vspace{1em}
\textbf{Input:}\\
Synthesis of [Cd(L)(BIB) 0.5 ] n  (1) A mixture of H 2 L (0.2703 g, 1 mmol), Cd(OAc) 2 $\cdot$ 2H 2 O (0.2673 g, 1 mmol), NaOH (0.0801 g, 2 mmol) and BIB (0.189 g, 1 mmol) in 10 mL of distilled water was sealed in a 23 mL vial, and then heated to 150 $^\circ$C for 5 days. Colorless block crystals of 1 were collected in 56\% yield based on Cd.
	
\vspace{1em}
\hrule
\vspace{1em}
\textbf{Output:}
\begin{verbatim}
[
  {
    "Compound_Name": [
      "[Cd(L)(BIB) 0.5 ] n"
    ],
    "Metal_Source": [
      {
        "precursor_name": "Cd(OAc) 2 · 2H 2 O",
        "amount": "0.2673 g, 1 mmol"
      }
    ],
    "Organic_Linker": [
      {
        "precursor_name": "H 2 L",
        "amount": "0.2703 g, 1 mmol"
      },
      {
        "precursor_name": "BIB",
        "amount": "0.189 g, 1 mmol"
      }
    ],
    "Solvent": [
      {
        "precursor_name": "water",
        "amount": "10 mL"
      }
    ],
    "Modulator": [
      {
        "precursor_name": "NaOH",
        "amount": "0.0801 g, 2 mmol"
      }
    ],
    "Reaction_Time": [
      "5 days"
    ],
    "Reaction_Temperature": [
      "150 °C"
    ]
  }
]
\end{verbatim}
\end{tcolorbox}

\emph{Prompts for Zero-Shot LLM in WoS-UiO-66 Experiment}

Zero-shot LLM extraction was conducted over the 261 filtered synthesis paragraphs from the WoS-UiO-66 dataset, using the GPT-4 API. These paragraphs were then cleaned and utilized in subsequent experiments.

Since this experiment was conducted on a pure UiO-66 dataset, relevant background and requirements were added to the prompts to determine whether the material is indeed pure UiO-66. The model was also instructed to exclude the synthesis of modified processes, or unrelated materials.

The following table presents the prompts, processes, and sample results used in this extraction task. The DOI of following condition extraction example paper is \url{10.1016/j.ces.2019.04.006} .

\begin{tcolorbox}[title=Zero-shot Prompts for WoS-UiO-66, colback=blue!5!white, colframe=blue!75!black, breakable]
\textbf{Prompt:}\\
You are a chemical expert with 20 years of experience in reviewing literature and extracting key information. Your expertise lies in systematically and accurately extracting synthesis parameters from chemical literature, specifically focusing on the synthesis sections of Metal-Organic Frameworks (MOFs). As a chemistry researcher, I require your assistance in efficiently and precisely extracting the synthesis parameters of UIO-66, a type of MOF, from chemical literature. I will tip 10\$ for more precise extraction result.
\newline \\
Your task is to summarize the following details into a JSON format table from the input paragraph: `Metal\_Source', `Organic\_Linker', `Solvent', `Modulator', `Reaction\_Time', and `Reaction\_Temperature'. `Metal\_Source', `Organic\_Linker', `Solvent', and `Modulator' should also include their respective amounts.
Only focus on the synthesis process. Ignore other processes, including the organic precursor synthesis pre-process, the drying process, the crystallization post-process, and any modification processes if mentioned. Do not extract them.
\newline \\
Only extract details for pure UIO-66, also known as pure Zr-BDC. Do not extract information on other chemical compounds such as UIO-67. Pure UIO-66 refers to the unmodified compound; therefore, do not extract information on UIO-66-NH2 or Cu@UIO-66, as they are not considered pure.
\newline \\
Detailed Format Descriptions and Requirements for Each Class:\\
The output should be a JSON table list. Each JSON format table represents a UIO-66 synthesis process. If there is only one synthesis process, the JSON list should contain only one JSON format table. If there is more than one synthesis process, they should be placed in separate JSON tables. \\
Each JSON format table should contain: `Metal\_Source', `Organic\_Linker', `Solvent', `Modulator', `Reaction\_Time', and `Reaction\_Temperature'. \\
- `Reaction\_Time' and `Reaction\_Temperature' should be lists of strings. \\
- `Metal\_Source', `Organic\_Linker', `Solvent', and `Modulator' should all be lists of dictionaries. Each dictionary should have the keys `precursor\_name' and `amount'. `amount' includes weight, volume, or molar weight, presented according to the paragraph text, but not including the concentration, proportion, or rate. Do not include any proportions, ratios, concentrations, or percentages in the name or amount of any parameter.
\newline \\
Extract the name and amount of `Metal\_Source', `Organic\_Linker', `Solvent', and `Modulator' during the synthesis process of MOF according to the paragraph, and place them in the dictionary with the keys `precursor\_name' and `amount'. \\
`Reaction\_Time' and `Reaction\_Temperature' should be extracted from the paragraph. If there is no time or temperature mentioned in the text, set the value to an empty list. \\
If some values are missing, not presented, or not mentioned, leave them empty in the output JSON text. The output should only contain the list.\newline \\
Example Output Structure:
\begin{verbatim}
[
  {
    "Metal_Source": [
      {
        "precursor_name": "",
        "amount": ""
      }
    ],
    "Organic_Linker": [
      {
        "precursor_name": "",
        "amount": ""
      },
      {
        "precursor_name": "",
        "amount": ""
      }
    ],
    "Solvent": [
      {
        "precursor_name": "",
        "amount": ""
      },
      {
        "precursor_name": "",
        "amount": ""
      }
    ],
    "Modulator": [
      {
        "precursor_name": "",
        "amount": ""
      }
    ],
    "Reaction_Time": [],
    "Reaction_Temperature": []
  }
]

\end{verbatim}

\vspace{1em}
\hrule
\vspace{1em}
\textbf{Input:}\\
2.1.2. Synthesis UiO-66 was synthesized according to a slightly modified protocol based on previous report by Schaate (2011). In detail, ZrCl\textsubscript{4} (1.00 g, 4.29 mmol) and acetic acid (7.36 mL, 128.7 mmol) were dissolved in DMF (250 mL, 3.23 mol) by ultrasound for about 10 min, which was then supplemented with H\textsubscript{2}BDC (0.713 g, 4.29 mmol). The solution was transferred to a 500 mL Teflon-lined autoclave and maintained at 393 K under static conditions for 24 h. The precipitates were isolated and then subject to three-time solvent exchange using fresh 30 mL DMF, in which the precipitates were soaked in DMF for 12h each time. Afterwards, the precipitates were soaked in free ethanol (30 mL) for three-time solvent exchange similarly. Finally, the solids were first dried in an oven at 80$^\circ$C for 12h, and then dried under vacuum at 453K for 12h. UiO-67 was synthesized using a previously reported method in literature (Oien-Odegaard et al., 2016) with a slight modification.

\vspace{1em}
\hrule
\vspace{1em}
\textbf{Output:}
\begin{verbatim}
[
  {
    "Metal_Source": [
      {
        "precursor_name": "ZrCl4",
        "amount": "1.00 g, 4.29 mmol"
      }
    ],
    "Organic_Linker": [
      {
        "precursor_name": "H2BDC",
        "amount": "0.713 g, 4.29 mmol"
      }
    ],
    "Solvent": [
      {
        "precursor_name": "DMF",
        "amount": "250 mL, 3.23 mol"
      },
      {
        "precursor_name": "acetic acid",
        "amount": "7.36 mL, 128.7 mmol"
      }
    ],
    "Modulator": [
      {
        "precursor_name": "Acetic acid",
        "amount": "7.36 mL"
      }
    ],
    "Reaction_Time": [
      "24 hours"
    ],
    "Reaction_Temperature": [
      "393 K"
    ]
  }
]
\end{verbatim}
\end{tcolorbox}

\emph{Prompts for Few-Shot LLM in WoS-UiO-66 Experiment}

Few-shot LLM extraction was conducted over the 261 filtered synthesis paragraphs from the WoS-UiO-66 dataset, using the GPT-4 API. To select 3 relevant examples from 87 annotations, we employed the BM25 algorithm. These examples were then used to extract synthesis conditions. To specifically improve extraction quality, background information and constraints based on domain knowledge were added to the prompts (\rtab{MaterialKnowledge}). The results were subsequently cleaned and utilized in subsequent experiments.

Since this experiment was conducted on a pure UiO-66 dataset, relevant background and requirements were added to the prompts to determine whether the material is indeed pure UiO-66. The model was also instructed to exclude the synthesis of modified processes, or unrelated materials.

The following table presents the prompts, examples,  processes, and sample results used in this extraction task. The DOI of following condition extraction example paper is \url{10.1016/j.ces.2019.04.006} .

\begin{tcolorbox}[title=Few-shot Prompts for WoS-UiO-66, colback=blue!5!white, colframe=blue!75!black, breakable]
\textbf{Prompt:}\\
You are a chemical expert with 20 years of experience in reviewing literature and extracting key information. Your expertise lies in systematically and accurately extracting synthesis parameters from chemical literature, specifically focusing on the synthesis sections of Metal-Organic Frameworks (MOFs). As a chemistry researcher, I require your assistance in efficiently and precisely extracting the synthesis parameters of UIO-66, a type of MOF, from chemical literature. I will tip 10\$ for more precise extraction result.\\

Your task is to summarize the following details into a JSON format table from the input paragraph: \lq{}Metal\_Source\rq{}, \lq{}Organic\_Linker\rq{}, \lq{}Solvent\rq{}, \lq{}Modulator\rq{}, \lq{}Reaction\_Time\rq{}, and \lq{}Reaction\_Temperature\rq{}. \lq{}Metal\_Source\rq{}, \lq{}Organic\_Linker\rq{}, \lq{}Solvent\rq{}, and \lq{}Modulator\rq{} should also include their respective amounts.\\
Only focus on the synthesis process. Ignore other processes, including the organic precursor synthesis pre-process, the drying process, the crystallization post-process, and any modification processes if mentioned. Do not extract them.\\
Only extract details for pure UIO-66, also known as pure Zr-BDC. Do not extract information on other chemical compounds such as UIO-67. Pure UIO-66 refers to the unmodified compound; therefore, do not extract information on UIO-66-NH\textsubscript{2} or Cu@UIO-66, as they are not considered pure.\\

Background Information and Detailed Instructions:\\
\lq{}Compound\_Name\rq{} of MOFs (Metal-Organic Frameworks): MOFs are porous materials formed by the coordination of metal ions or clusters with organic ligands.\\
\lq{}Metal\_Source\rq{}: In MOF synthesis, a Metal Source is a precursor compound containing metal ions that form part of the MOF structure. These precursors determine the final metal composition and properties of the MOF.\\
\lq{}Organic\_Linker\rq{}: Refers to the organic precursor molecule linking metal ions or clusters in the MOF, influencing the framework's topology, porosity, and functionality.\\
\lq{}Solvent\rq{}: The liquid medium in which reactants are dissolved. The reaction occurs in the solvent environment.\\
\lq{}Modulator\rq{}: A substance used to adjust reaction conditions, such as pH value.\\
\lq{}Reaction process\rq{}: The synthesis process that produces the MOF materials.\\
\lq{}Crystallization process\rq{}: The process of forming crystals, which should NOT be included.\\

Detailed Format Descriptions and Requirements for Each Class:\\
The output should be a JSON table list. Each JSON format table represents a UIO-66 synthesis process. If there is only one synthesis process, the JSON list should contain only one JSON format table. If there is more than one synthesis process, they should be placed in separate JSON tables.\\
Each JSON format table should contain: \lq{}Metal\_Source\rq{}, \lq{}Organic\_Linker\rq{}, \lq{}Solvent\rq{}, \lq{}Modulator\rq{}, \lq{}Reaction\_Time\rq{}, and \lq{}Reaction\_Temperature\rq{}.\\
- \lq{}Reaction\_Time\rq{} and \lq{}Reaction\_Temperature\rq{} should be lists of strings.\\
- \lq{}Metal\_Source\rq{}, \lq{}Organic\_Linker\rq{}, \lq{}Solvent\rq{}, and \lq{}Modulator\rq{} should all be lists of dictionaries. Each dictionary should have the keys \lq{}precursor\_name\rq{} and \lq{}amount\rq{}. \lq{}amount\rq{} includes weight, volume, or molar weight, presented according to the paragraph text, but not including the concentration, proportion, or rate. Do not include any proportions, ratios, concentrations, or percentages in the name or amount of any parameter.\\
Extract the name and amount of \lq{}Metal\_Source\rq{}, \lq{}Organic\_Linker\rq{}, \lq{}Solvent\rq{}, and \lq{}Modulator\rq{} during the synthesis process of MOF according to the paragraph, and place them in the dictionary with the keys \lq{}precursor\_name\rq{} and \lq{}amount\rq{}.\\
\lq{}Reaction\_Time\rq{} and \lq{}Reaction\_Temperature\rq{} should be extracted from the paragraph. If there is no time or temperature mentioned in the text, set the value to an empty list.\\
If some values are missing, not presented, or not mentioned, leave them empty in the output JSON text. The output should only contain the list.\\
\newline
Important Notes:\\
The crystallization process is not the synthesis process. A synthesis process is often related to heating, refluxing, stirring, or layering, which provide the most extreme reaction conditions. The step that involves these intense conditions is typically the synthesis step. Ignore all related parameters during the crystallization process, including crystallization temperature, time, solvent, or other parameters.\\
The synthesis process often lasts several hours. If the process takes more than several weeks, it is usually crystallization unless it specifically mentions synthesis and often involves heating.\\
\newline
Example Output Structure:\\
\begin{verbatim}
[
  {
    "Metal_Source": [
      {
        "precursor_name": "",
        "amount": ""
      }
    ],
    "Organic_Linker": [
      {
        "precursor_name": "",
        "amount": ""
      },
      {
        "precursor_name": "",
        "amount": ""
      }
    ],
    "Solvent": [
      {
        "precursor_name": "",
        "amount": ""
      },
      {
        "precursor_name": "",
        "amount": ""
      }
    ],
    "Modulator": [
      {
        "precursor_name": "",
        "amount": ""
      }
    ],
    "Reaction_Time": [],
    "Reaction_Temperature": []
  }
]
\end{verbatim}

\textbf{Sample Input 1:}\\
2.2.1. Synthesis of Zn(H\textsubscript{3}L)\textperiodcentered 2H\textsubscript{2}O(1) A mixture of 1.0 mmol of zinc(II) acetate, 0.5 mmol of H\textsubscript{5}L and 10.0 ml of deionized water was sealed into an autoclave equipped with a Teflon liner (25 ml), and then heated at 180 $^\circ$C for 5 days. The initial and final PH values are 2.5 and 2.0, respectively. Colorless crystals of 1 were recovered in a ca. 48\% yield (based on zinc). Elemental analysis for 1 C\textsubscript{20}H\textsubscript{34}N\textsubscript{2}O\textsubscript{20}P\textsubscript{4}Zn\textsubscript{2}: C, 27.33; H, 4.21; N, 3.12\%. Calcd: C, 27.39; H, 3.91; N, 3.19. IR data (KBr, cm\textsuperscript{-1}): 3412 s, 3010 w, 2957 w, 1718 m, 1650 m, 1417 w, 1331 m, 1253 m, 1143 vs, 1029 m, 1015 m, 924 w, 761 w, 598 w, 504 w, 466 w.\\

\textbf{Sample Output 1:}
\begin{verbatim}
[
  {
    "Metal_Source": [
      {
        "precursor_name": "ZrCl 4",
        "amount": "0.40 g, 1.71 mmol"
      }
    ],
    "Organic_Linker": [
      {
        "precursor_name": "H 2 BDC",
        "amount": "0.28 g, 1.71 mmol"
      }
    ],
    "Solvent": [
      {
        "precursor_name": "DMF",
        "amount": "45 mL"
      },
      {
        "precursor_name": "DMF",
        "amount": "15 mL"
      }
    ],
    "Modulator": [],
    "Reaction_Time": [
      "24 h"
    ],
    "Reaction_Temperature": [
      "120 °C"
    ]
  }
]
\end{verbatim}
\textbf{Sample Input 2:}\\
Synthesis of UiO-66 and H-UiO-66. UiO-66 was synthesized according to the reported literature; 30 ZrCl 4 (0.96 g, 4 mmol) and H 2 BDC (1.328 g, 8 mmol) were dissolved in 160 mL DMF at room temperature in a Teﬂon reaction still (200 mL) and sonicated for 20 min. Then, the mixture was sealed and placed in a preheated oven at 393 K for 24 h, after which it was allowed to cool down to room temperature in air. The generated solid was centrifuged, washed with DMF (twice) and then ethanol (twice), and dried at 333 K for 12 h.
\newline \\
The dried sample was further activated at 573 K in a furnace under air atmosphere for 3 h.
\newline \\
\textbf{Sample Output 2:}
\begin{verbatim}
[
  {
    "Metal_Source": [
      {
        "precursor_name": "ZrCl 4",
        "amount": "0.96 g, 4 mmol"
      }
    ],
    "Organic_Linker": [
      {
        "precursor_name": "H 2 BDC",
        "amount": "1.328 g, 8 mmol"
      }
    ],
    "Solvent": [
      {
        "precursor_name": "DMF",
        "amount": "160 mL"
      }
    ],
    "Modulator": [],
    "Reaction_Time": [
      "24 h"
    ],
    "Reaction_Temperature": [
      "393 K"
    ]
  }
]
\end{verbatim}
\textbf{Sample Input 3:}\\
UIO-66 crystals were synthesized according to Schaate et al. \lbrack46].\\
Typically, ZrCl\textsubscript{4} (0.080 g, 0.343 mmol) and 0.3 mL acetic acid were dissolved in 20 mL DMF in a Teflon-lined bomb under ultrasonication for 1 min.\\
Terephthalic acid (0.057 g, 0.343 mmol) was then added to the clear solution.\\
The Teflon-lined bomb was sealed and placed in an oven at 120 \textdegree C for 24 h under static conditions, and then cooled to room temperature.\\
The precipitates were isolated by centrifugation.\\
The solid was suspended in DMF (5 mL), and washed with DMF.\\
After washing, the suspension was centrifuged and the solvent was decanted off.\\
This procedure was repeated three times.\\
The obtained particles were washed with ethanol (5 mL) in the same way as described for washing with DMF.\\
Finally, the solid was dried at 60 \textdegree C under reduced pressure.\\
\textbf{Sample Output 3:}
\begin{verbatim}
[
  {
    "Metal_Source": [
      {
        "precursor_name": "ZrCl 4",
        "amount": "0.080 g, 0.343 mmol"
      }
    ],
    "Organic_Linker": [
      {
        "precursor_name": "Terephthalic acid",
        "amount": "0.057 g, 0.343 mmol"
      }
    ],
    "Solvent": [
      {
        "precursor_name": "DMF",
        "amount": "20 mL"
      }
    ],
    "Modulator": [
      {
        "precursor_name": "acetic acid",
        "amount": "0.3 mL"
      }
    ],
    "Reaction_Time": [
      "24 h"
    ],
    "Reaction_Temperature": [
      "120 °C"
    ]
  }
]
\end{verbatim}
\textbf{Sample Input 4:}\\
In order to obtain UiO-66, ZrCl 4 (0.080 g, 0.343 mmol) and the linker precursor, benzene-1,4-dicarboxylic acid (0.057 g, 0.343 mmol), in an equivalent molar ratio were dissolved in 20 mL of DMF in a 120 mL Teflon-capped glass jar by using ultrasound for about 1 min. 0.7 mL of acetic acid (AcOH, acting as a modulator) was then added into the solution and dispersed by ultrasound for about 1 min. The tightly capped jars were kept in an oven at 120 \textdegree C under static conditions for 24 h. White precipitates were produced and were isolated by centrifugation after cooling down to room temperature.\\
\textbf{Sample Output 4:}
\begin{verbatim}
[
  {
    "Metal_Source": [
      {
        "precursor_name": "ZrCl 4",
        "amount": "0.080 g, 0.343 mmol"
      }
    ],
    "Organic_Linker": [
      {
        "precursor_name": "benzene-1,4-dicarboxylic acid",
        "amount": "0.057 g, 0.343 mmol"
      }
    ],
    "Solvent": [
      {
        "precursor_name": "DMF",
        "amount": "20 mL"
      }
    ],
    "Modulator": [
      {
        "precursor_name": "acetic acid (AcOH",
        "amount": "0.7 mL"
      }
    ],
    "Reaction_Time": [
      "24 h"
    ],
    "Reaction_Temperature": [
      "120 °C"
    ]
  }
]
\end{verbatim}
\textbf{Sample Input 5:}\\
3.1.1. Synthesis of UiO-66. UiO-66 was synthesized by a reported solvothermal method 43 with slight modiﬁcation. In a typical synthesis, both ZrCl 4 (1.631 g, 7 mmol) and H 2 BDC (1.163 g, 7 mmol) were dissolved separately in 40 mL of a DMF solution and stirred for 30 min, and then the ZrCl 4 -DMF solution was transferred into the H 2 BDC-DMF solution and stirred for 30 min. The so-formed mixed solution is then transferred to a 100 mL Teﬂon-lined stainless-steel autoclave. The autoclave was sealed and subjected to hydrothermal treatment in a hotair oven at 120 \textdegree C for 24 h. After cooling naturally to room temperature, the sample was isolated by centrifugation, and the residuals DMF and H 2 BDC were removed from the crystalline pores by subsequent rinsing with methanol several times. Then it was allowed to soak in methanol for 3 days and isolated by centrifugation, followed by drying under vacuum (100 \textdegree C, 12 h), and the ﬁnal obtained white sample was named UiO-66.\\
\textbf{Sample Output 5:}
\begin{verbatim}
[
  {
    "Metal_Source": [
      {
        "precursor_name": "ZrCl 4",
        "amount": "1.631 g, 7 mmol"
      }
    ],
    "Organic_Linker": [
      {
        "precursor_name": "H 2 BDC",
        "amount": "1.163 g, 7 mmol"
      }
    ],
    "Solvent": [
      {
        "precursor_name": "DMF",
        "amount": "40 mL"
      },
      {
        "precursor_name": "DMF",
        "amount": "40 mL"
      }
    ],
    "Modulator": [],
    "Reaction_Time": [
      "24 h"
    ],
    "Reaction_Temperature": [
      "120 °C"
    ]
  }
]
\end{verbatim}
\textbf{Input:}\\
Synthesis of UiO-66. UiO-66 was synthesized according to the previously reported method with some modiﬁcation.\\
37 Zirconium chloride (0.18 g) was dissolved in 2 mL of DMF by stirring for 30 min. In another beaker, 0.127 g of terephthalic acid was dissolved in 2 mL of DMF by stirring for 15 min; 0.065 mL of ammonia aqueous (2 mol L -1 ) was then added. The terephthalic acid solution was slowly added to the dissolved zirconium chloride, and then, 6 mL of DMF was added and stirred for 20 min. After this, the mixture was transferred to a 15 mL Teﬂon liner stainless steel autoclave tube and placed in an oven at 120 \textdegree C for 24 h. The ﬁnal products were separated by centrifugation (9000g for 3 min), washed with DMF, activated by methanol exchange (immersing in methanol for 12 h at room temperature and repeating the procedure three times), and then dried under a vacuum at 100 \textdegree C overnight.\\
\textbf{Output:}
\begin{verbatim}
[
  {
    "Metal_Source": [
      {
        "precursor_name": "Zirconium chloride",
        "amount": "0.18 g"
      }
    ],
    "Organic_Linker": [
      {
        "precursor_name": "terephthalic acid",
        "amount": "0.127 g"
      }
    ],
    "Solvent": [
      {
        "precursor_name": "DMF ",
        "amount": "2 mL"
      },
      {
        "precursor_name": "DMF",
        "amount": "2 mL"
      },
      {
        "precursor_name": "DMF",
        "amount": "6 mL "
      }
    ],
    "Modulator": [
      {
        "precursor_name": "ammonia aqueous",
        "amount": "0.065 mL"
      }
    ],
    "Reaction_Time": [
      "24 h"
    ],
    "Reaction_Temperature": [
      "120 °C"
    ]
  }
]
\end{verbatim}
\end{tcolorbox}

\emph{Prompts for Surface Area Extraction in WoS-UiO-66 Experiment}

We extracted the specific surface area data from the WoS-UiO-66 paper dataset. In each paper, after the first occurrence of the full term "specific surface area," it is typically abbreviated as "surface area." Therefore, we employed a pattern matching method for each paper to identify all paragraphs containing both "surface area" and "UiO-66." These paragraphs were then merged to extract the specific surface area information. The GPT-4 API was utilized as the extraction LLM model. We obtained a total of 918 specific surface area extraction results, which were subsequently used, along with the synthesis conditions obtained from the synthesis paragraph extraction, as model inputs for subsequent experiments.

The following table presents the prompts, processes, and sample results used in this extraction task. The DOI of following parameter extraction example paper is \url{10.1016/j.carbon.2019.03.049} .

\begin{tcolorbox}[title=Surface Area Extraction Prompts for WoS-UiO-66, colback=blue!5!white, colframe=blue!75!black, breakable]
\textbf{Prompt:}\\
You are a chemical expert, experienced on extracting specific information from scientific papers. Your task is to extract the specific surface area values of pure UIO-66 MOF material from given paragraphs. UIO-66 is a MOF material, also called Zr-BDC with Zr as its metal part and BDC as its organic framework. Only extract values for pure UIO-66, and ignore any mention of non-pure UIO-66 variants, such as those with additional compounds, modification or dopants (e.g., ``Pd@UiO-66'' or ``UiO-66-NH2''). Extract only the numerical value of the specific surface area, without the unit. Provide the results in JSON format.\\
\newline
Please reply with a list in JSON format as follows:
\begin{verbatim}
[
  {"name":"", "specific surface area":""},
  {"name":"", "specific surface area":""}
]
\end{verbatim}
If no specific surface area value is found, reply with an empty list:
\begin{verbatim}
[]
\end{verbatim}
\vspace{1em}
\hrule
\vspace{1em}
\textbf{Input:}\\
The crystallinity of synthesized MOF particles was characterized by XRD, their XRD process (Fig. 3(a)) consistent with those reported in previous studies \lbrack 42,43\rbrack. The morphology of MOF particles was examined by scanning electron microscope (SEM). Fig. 3b shows that MIL-140A exhibits a plate-like structure with a plate thickness of 60-70 nm and a lateral dimension of \~0.8-2{\textmu}m. In contrast, UiO-66 is spherical with a particle size of 60-70 nm (Fig. 3c). The BrunauereEmmetteTeller (BET) specific surface areas of MIL-140A and UiO-66 are 449.2 and 1210.4 m\textsuperscript{2}/g, respectively, as measured by N\textsuperscript{2} physisorption (Fig. S2 in the Supplementary data, SD). The pore size distribution analysis indicates that the pores of UiO-66 are in the range of 0.8e1.6 nm, and MIL-140A exhibits a pore size of 0.9 nm. As shown in Fig. S3 in the SD, GO nanosheets used in this study have a mean lateral dimension of \~2 {\textmu}m, which is much larger than that for individual UiO-66 particles and many MIL-140A particles. The Raman spectrum (Fig. S4 in the SD) shows an intense D band, indicating defective graphitic structures and thus the existence of abundant surface functional groups on GO nanosheets. In order to obtain reliable results, all the GO solutions were prepared with the same treatment to ensure the consistency of the physicochemical properties of GO solutions used in each membrane. On leveraging this unique GO/MOF architecture for potential selective adsorption for water treatment. The larger surface area and pore size of UiO-66 could be a potential good filter for such an application.

\vspace{1em}
\hrule
\vspace{1em}
\textbf{Output:}
\begin{verbatim}
[
    {
        "name": "UiO-66",
        "specific surface area": "1210.4"
    }
]
\end{verbatim}
\end{tcolorbox}

%%%%%%%%%%%%%%%% SUPPLEMENTARY FIGURES %%%%%%%%%%%%%%%

\clearpage

\begin{figure}
    \centering
    \includegraphics[width=\textwidth]{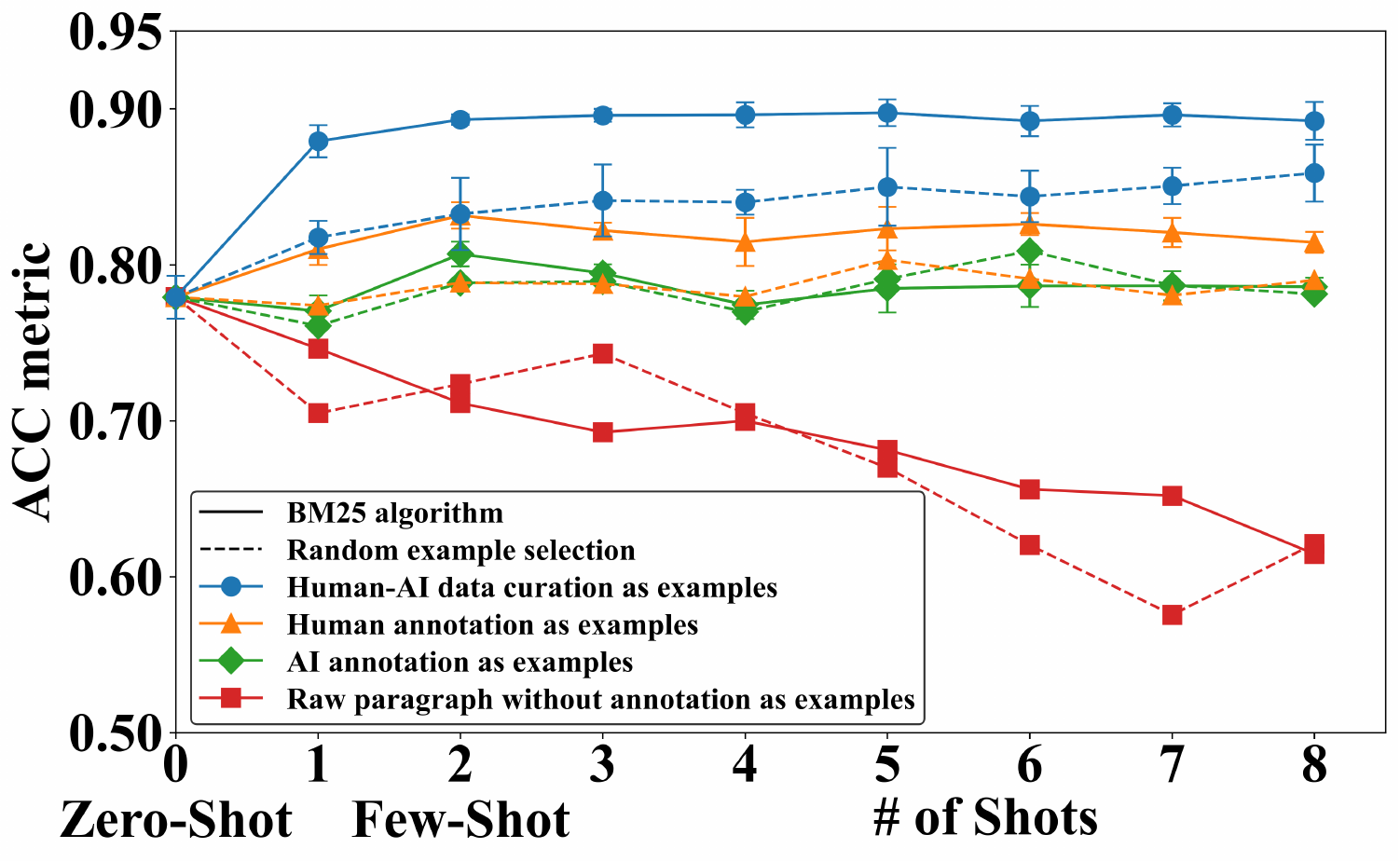}
    \caption{\textbf{The ACC of few-shot LLMs with different example pools and varying number of shots.}}
    \label{fig:ShotsComparisonACC}
\end{figure}

\clearpage

\begin{figure}
    \centering
    \includegraphics[width=\textwidth]{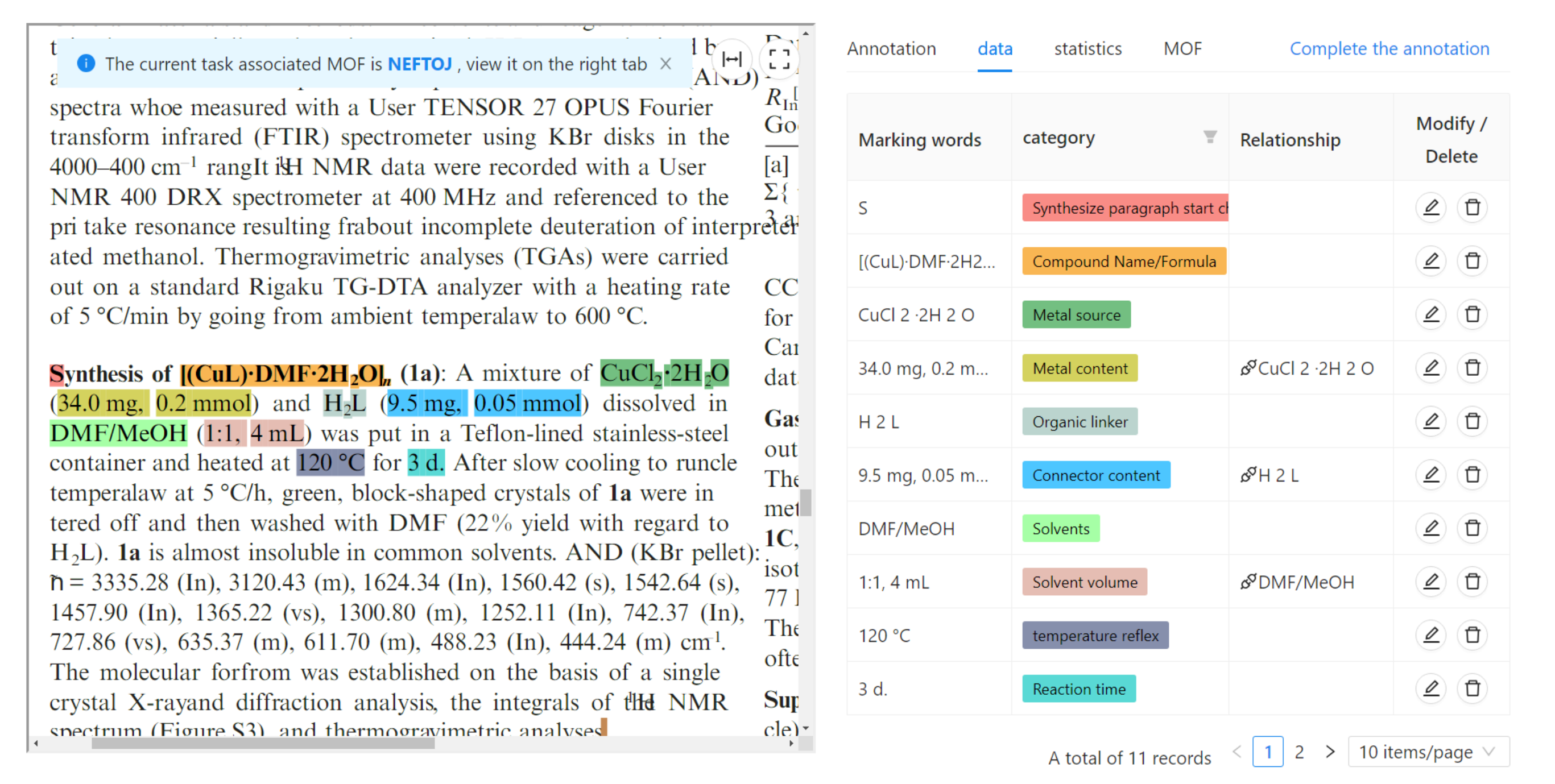}
    \caption{\textbf{User interface of our home-grown annotation software.}}
    \label{fig:AnnotationPlatform}
\end{figure}

\clearpage

\begin{figure}
    \centering
    \includegraphics[width=\textwidth]{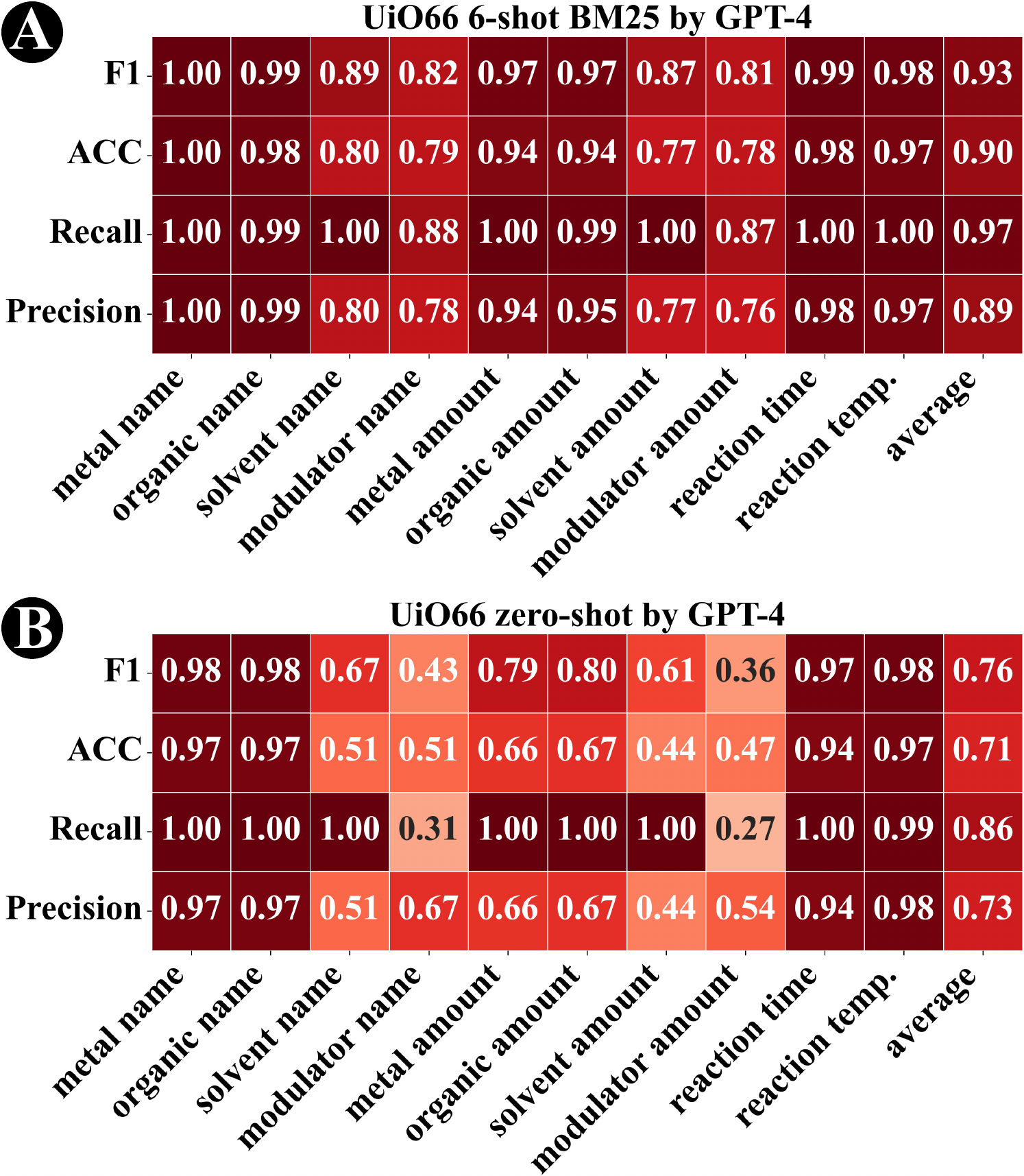}
    \caption{\textbf{Extraction performance on the WoS-UiO-66 dataset. Key indicators (F1, ACC, Precision, Recall) of the synthesis condition extraction performance on 87 WoS-UiO-66 MOFs with ground-truth data.} (A) Our 6-shot RAG algorithm; (B) Zero-shot LLM as the baseline.}
    \label{fig:UiO66BestVSZero}
\end{figure}

\clearpage

\begin{figure}
	\centering
    \includegraphics[width=\textwidth]{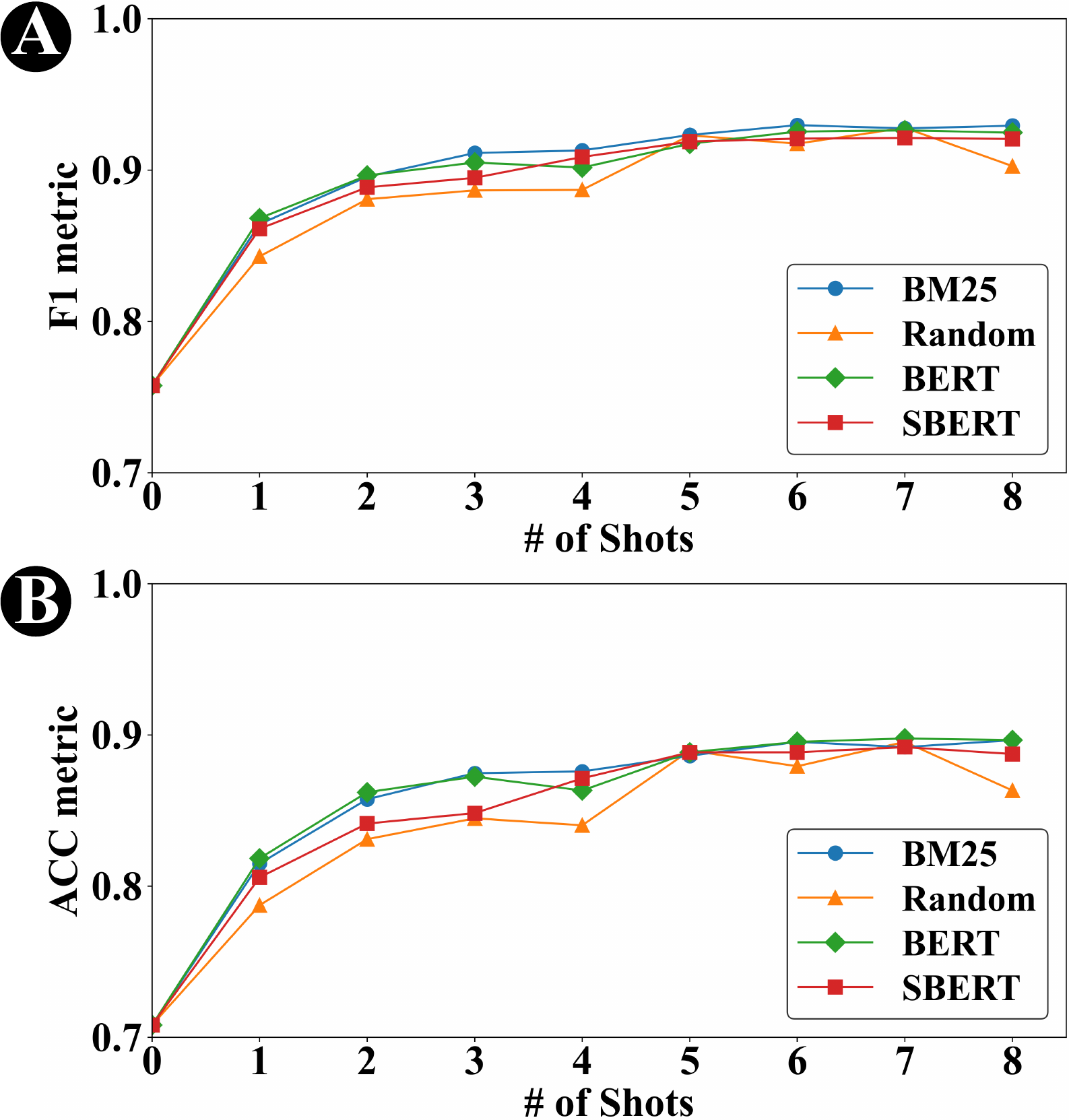}
    \vspace {-1.2em}
    \caption{\textbf{The impact of example data quality on extraction performance, with varying number of shots on WoS-UiO-66 dataset.}}
    \vspace{-0.15 in}
    % \vspace{-1.0em}
    \label{fig:UiO66ShotsComparison}
\end{figure}

\clearpage

\begin{figure}
    \centering
    \includegraphics[width=\textwidth]{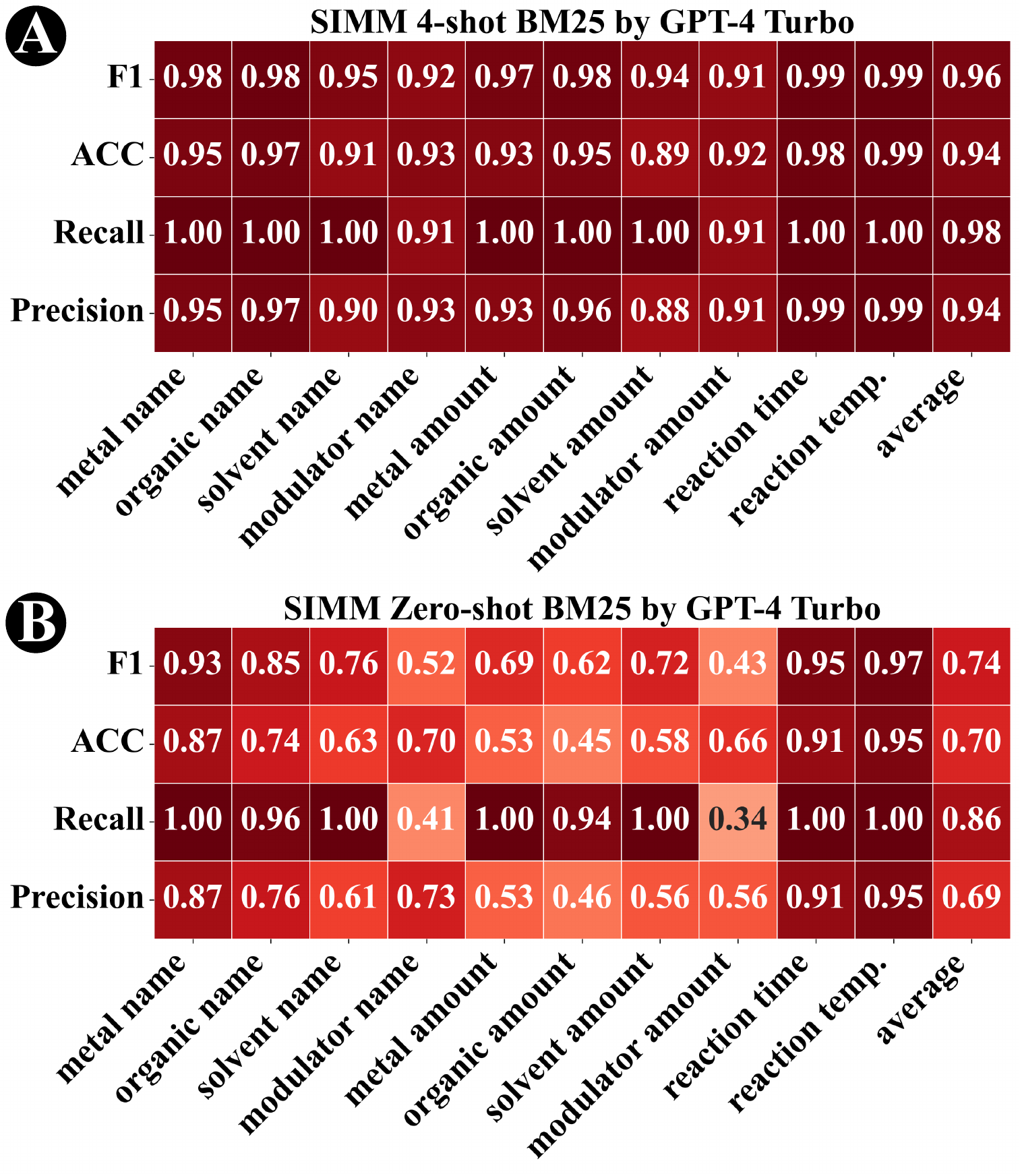}
    \caption{\textbf{Extraction performance on the SIMM dataset. Key indicators (F1, ACC, Precision, Recall) of the synthesis condition extraction performance on 573 SIMM MOFs with ground-truth data.} (A) Our 4-shot RAG algorithm; (B) Zero-shot LLM as the baseline.}
    \label{fig:SIMMBestVSZero}
\end{figure}

\clearpage

\begin{figure}
    \centering
    \includegraphics[width=\textwidth]{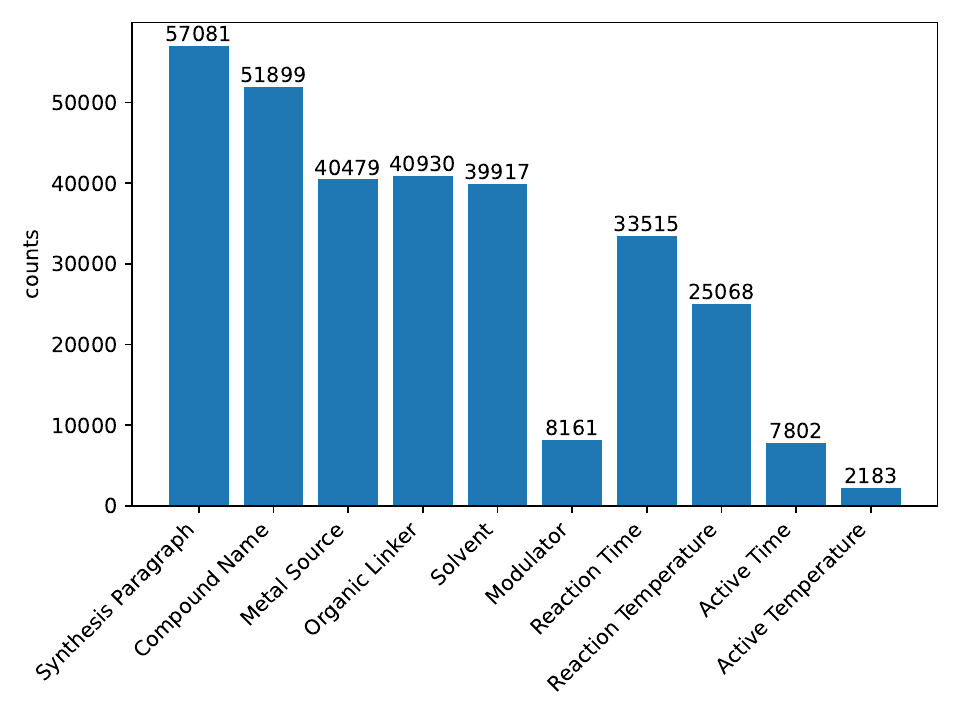}
    \caption{\textbf{Database statistics on the extraction result over the CSD-MOFs dataset.}}
    \label{fig:basic_statistics}
\end{figure}

\clearpage

\begin{figure}
    \centering
    \includegraphics[width=\textwidth]{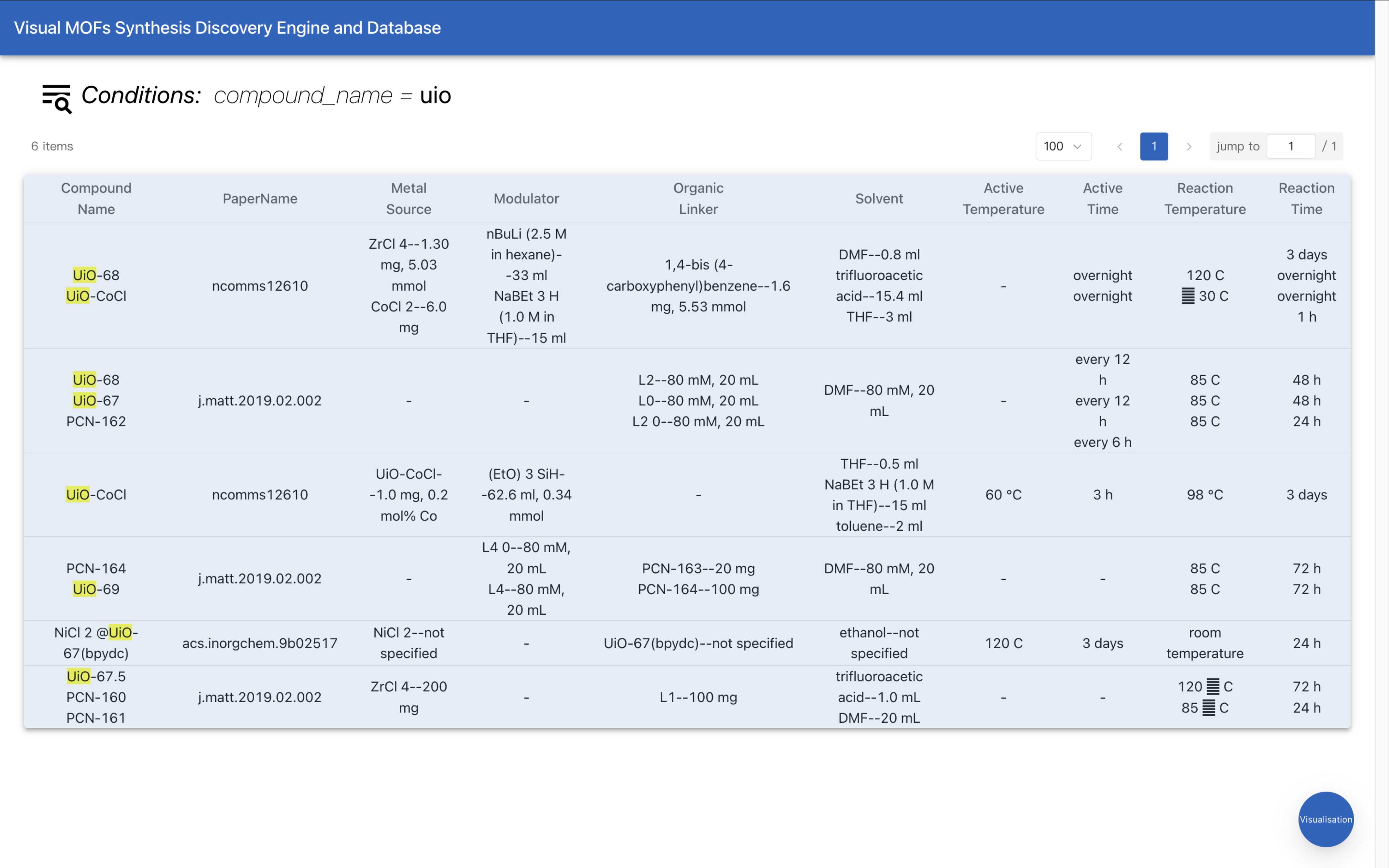}
    \caption{\textbf{An example of result page upon basic search operation.}}
    \label{fig:basic_search}
\end{figure}

\clearpage

\begin{figure}
    \centering
    \includegraphics[width=\textwidth]{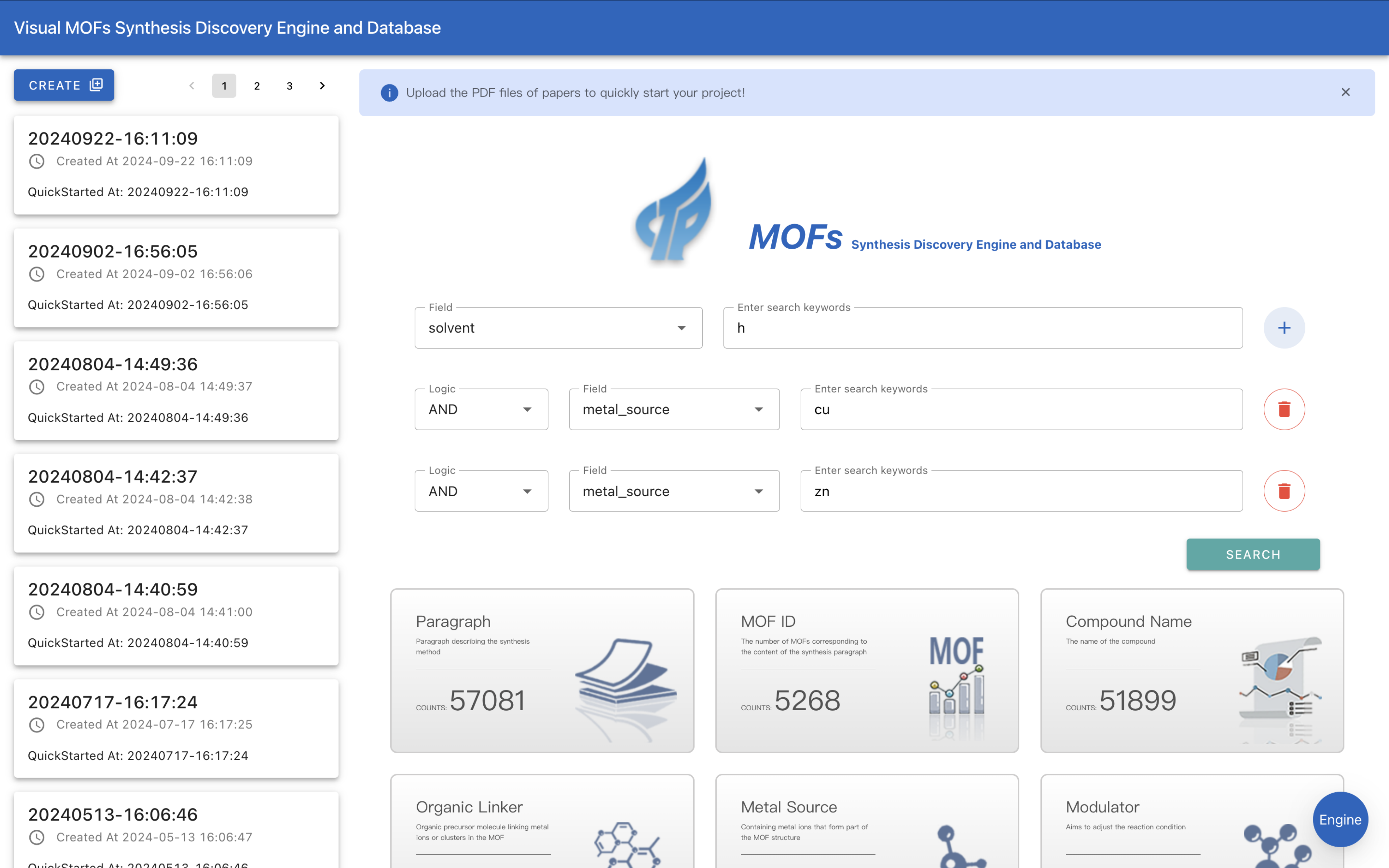}
    \caption{\textbf{An example of the advanced search interface combining boolean logic operators.}}
    \label{fig:advanced_search}
\end{figure}

\clearpage

\begin{figure}
    \centering
    \includegraphics[width=\textwidth]{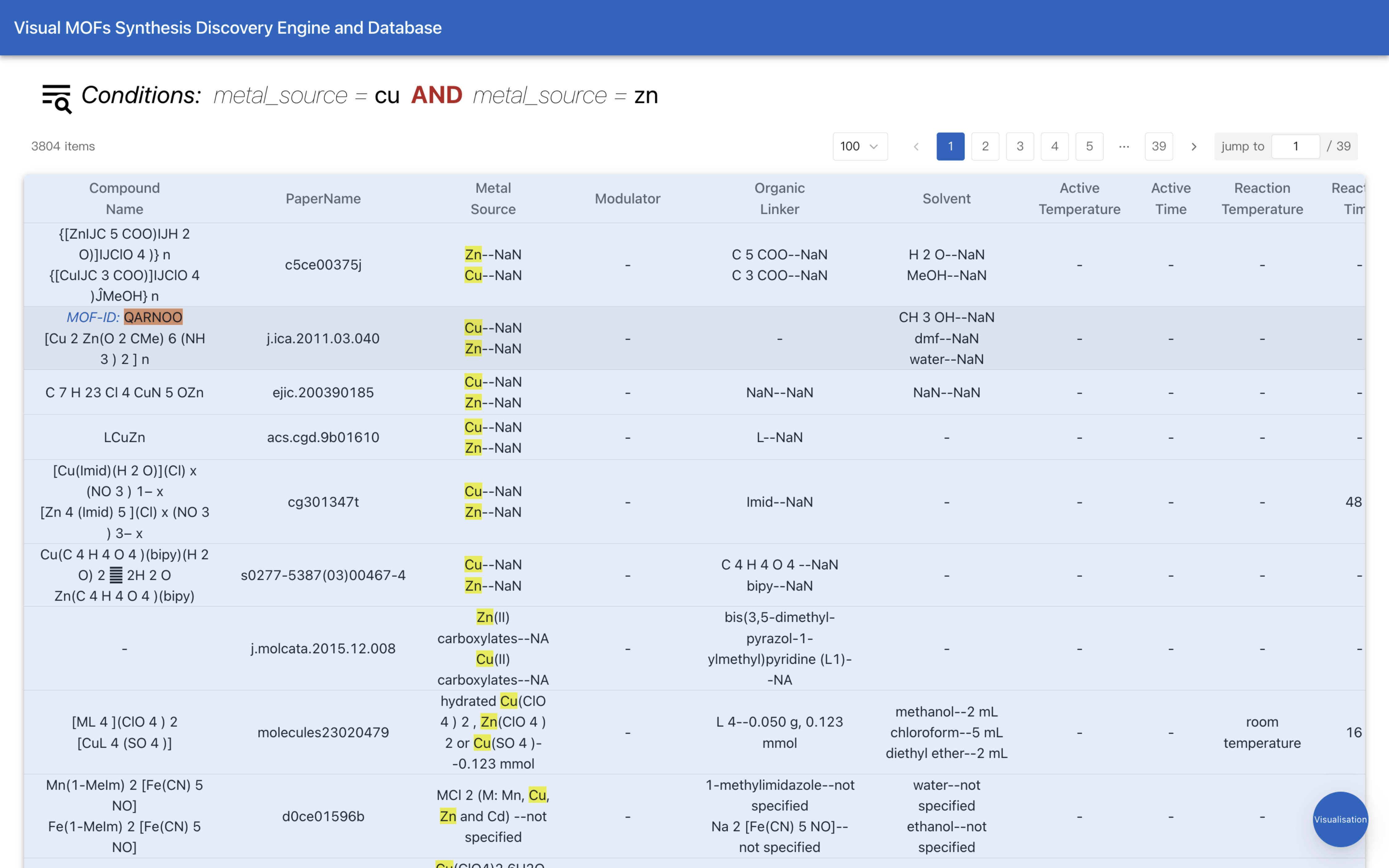}
    \caption{\textbf{An example of advanced search result.}}
    \label{fig:boolean_search1}
\end{figure}

\clearpage

\begin{figure}
    \centering
    \includegraphics[width=\textwidth]{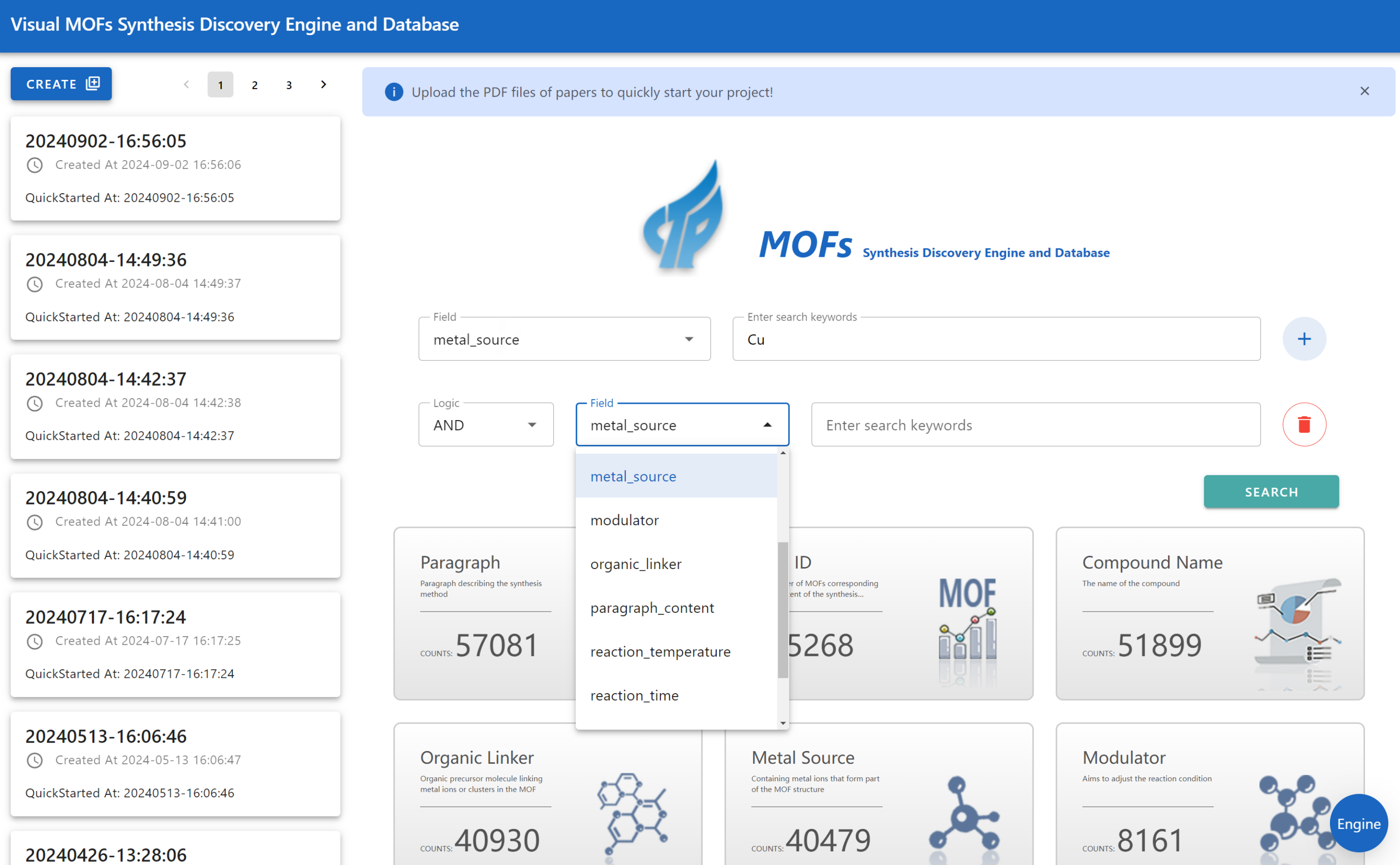}
    \caption{\textbf{User interface of the released MOFs synthesis database.} Users can perform faceted search on any relevant fields.}
    \label{fig:field_search}
\end{figure}

\clearpage

\begin{figure}
    \centering
    \includegraphics[width=\textwidth]{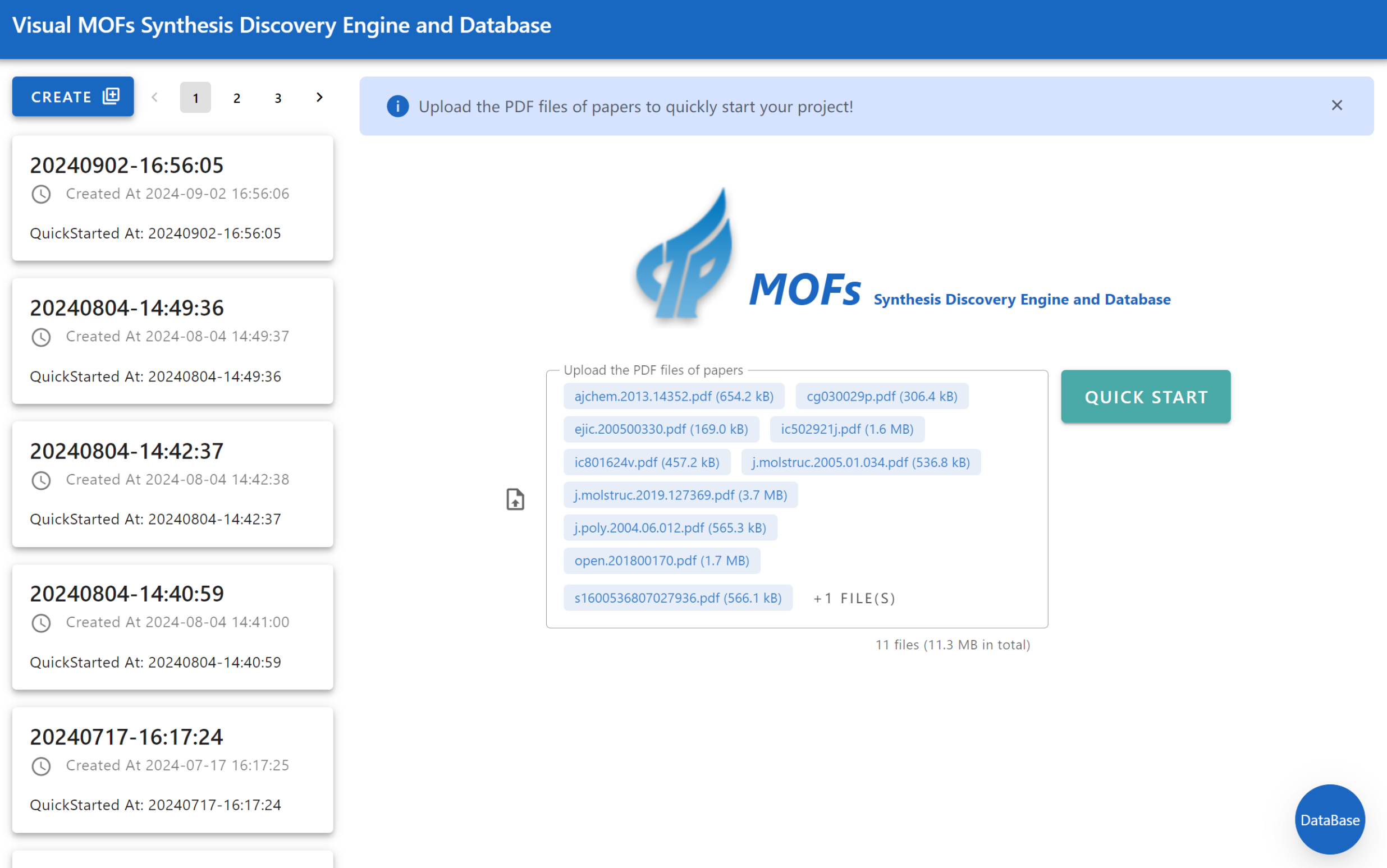}
    \caption{\textbf{The user interface of the online MOFs synthesis extraction engine.} Users can upload one or multiple papers.}
    \label{fig:vis-panel-step1}
\end{figure}

\clearpage

\begin{figure}
    \centering
    \includegraphics[width=\textwidth]{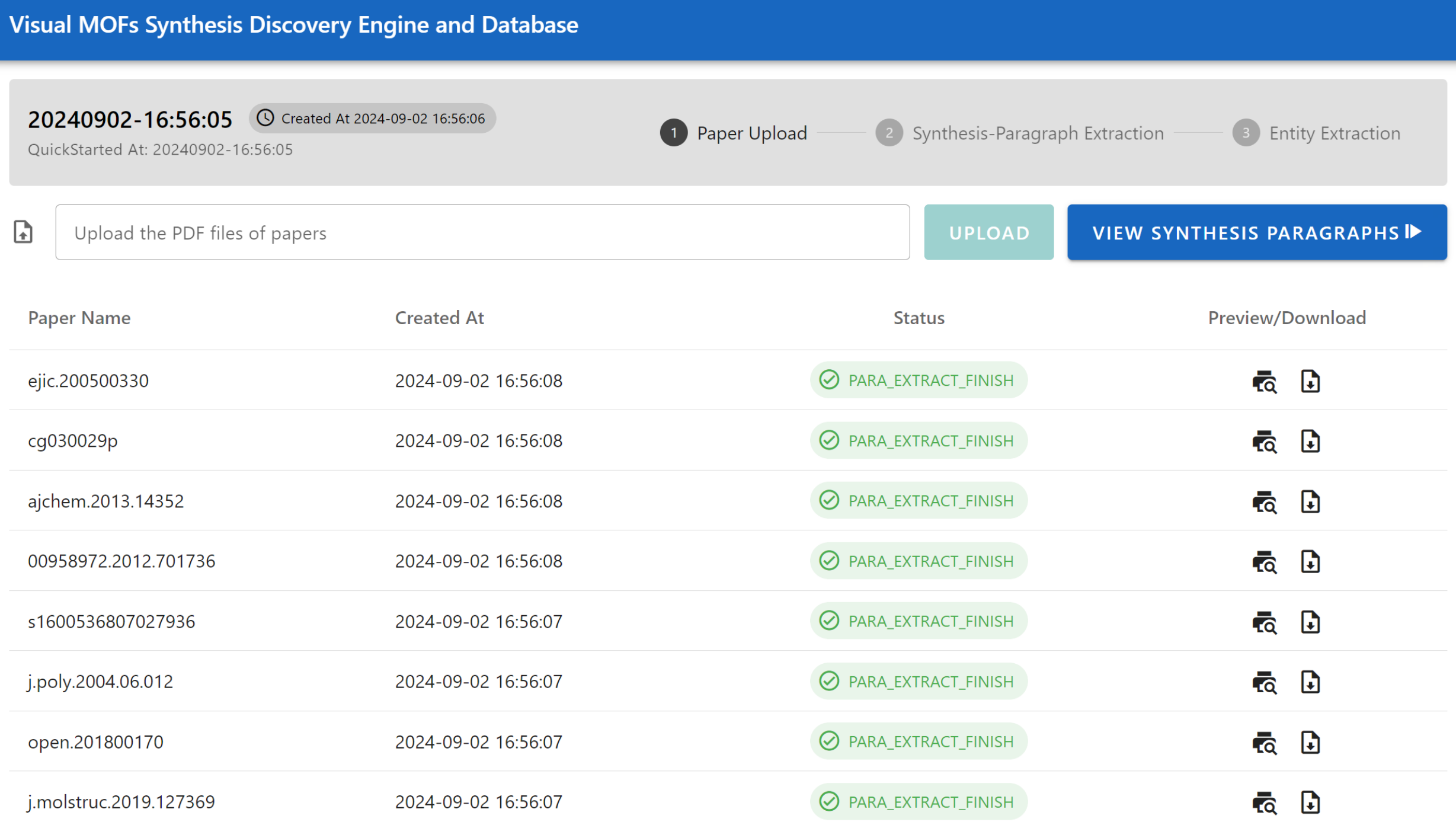}
    \caption{\textbf{The literature status interface in the engine to show the state of synthesis paragraph detection on all uploaded paper files.}}
    \label{fig:vis-panel-step2}
\end{figure}

\clearpage

\begin{figure}
    \centering
    \includegraphics[width=\textwidth]{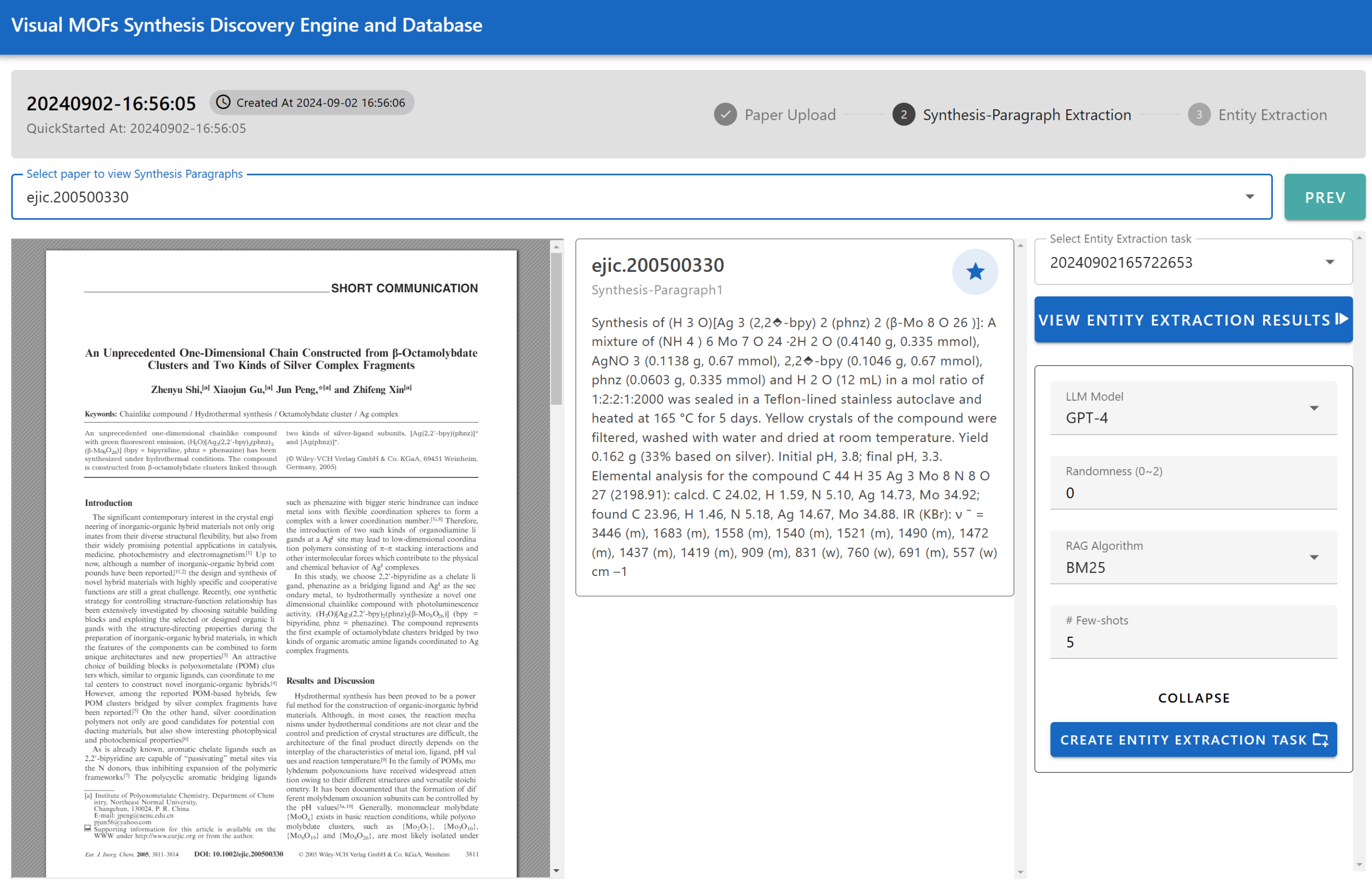}
    \caption{\textbf{The paragraphs info \& configuration panel in the engine.}}
    \label{fig:vis-panel-step3}
\end{figure}

\clearpage

\begin{figure}
    \centering
    \includegraphics[width=\textwidth]{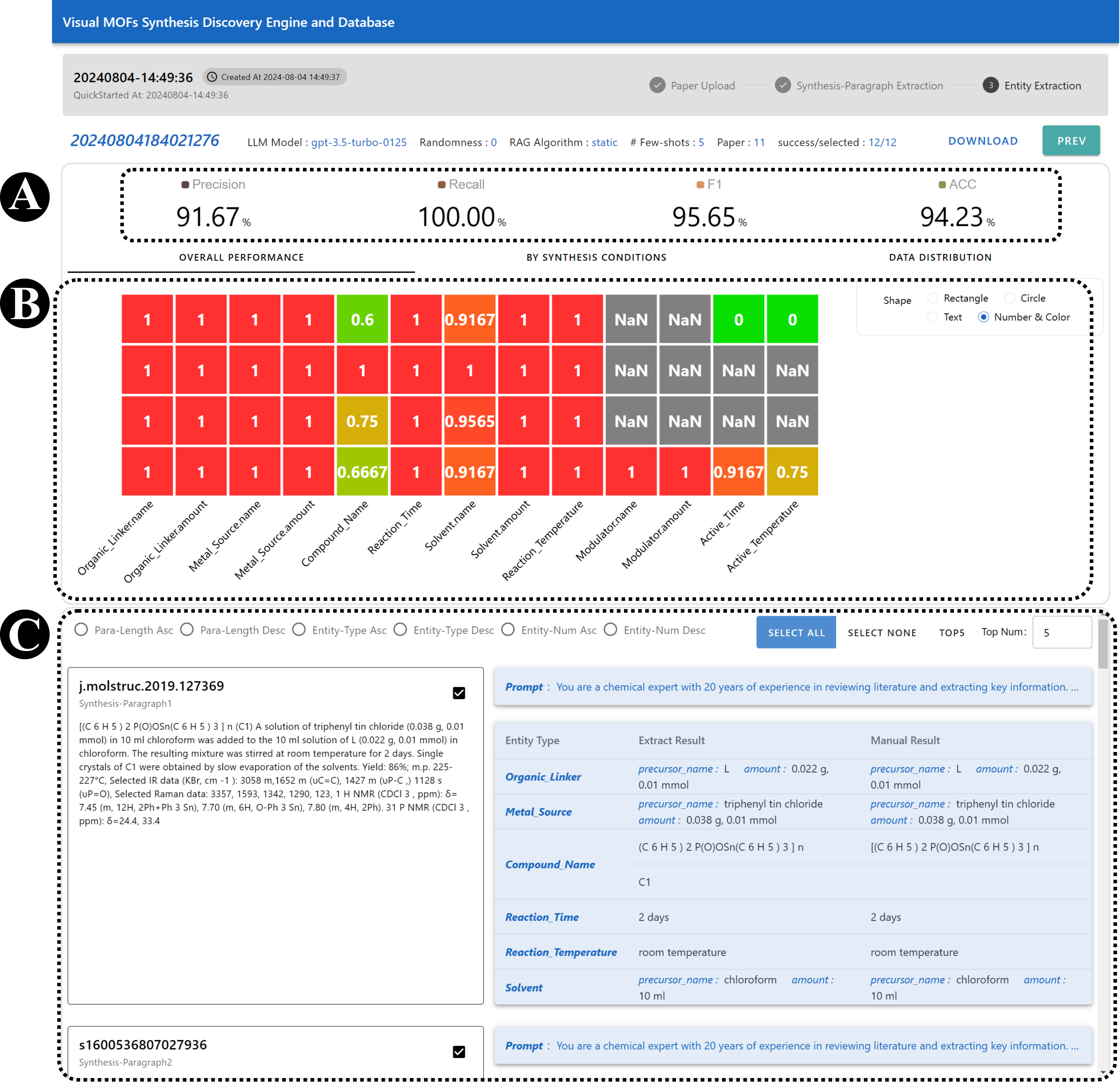}
    \caption{\textbf{Visualization interface for illustrating the synthesis extraction result.} (A) Overall metric panel; (B) Multi-tab detail panel (with the ``OVERALL PERFORMANCE'' tab selected); (C) Extraction result panel.}
    \label{fig:vis-panel-tab1}
\end{figure}

\clearpage

\begin{figure}
    \centering
    \includegraphics[width=\textwidth]{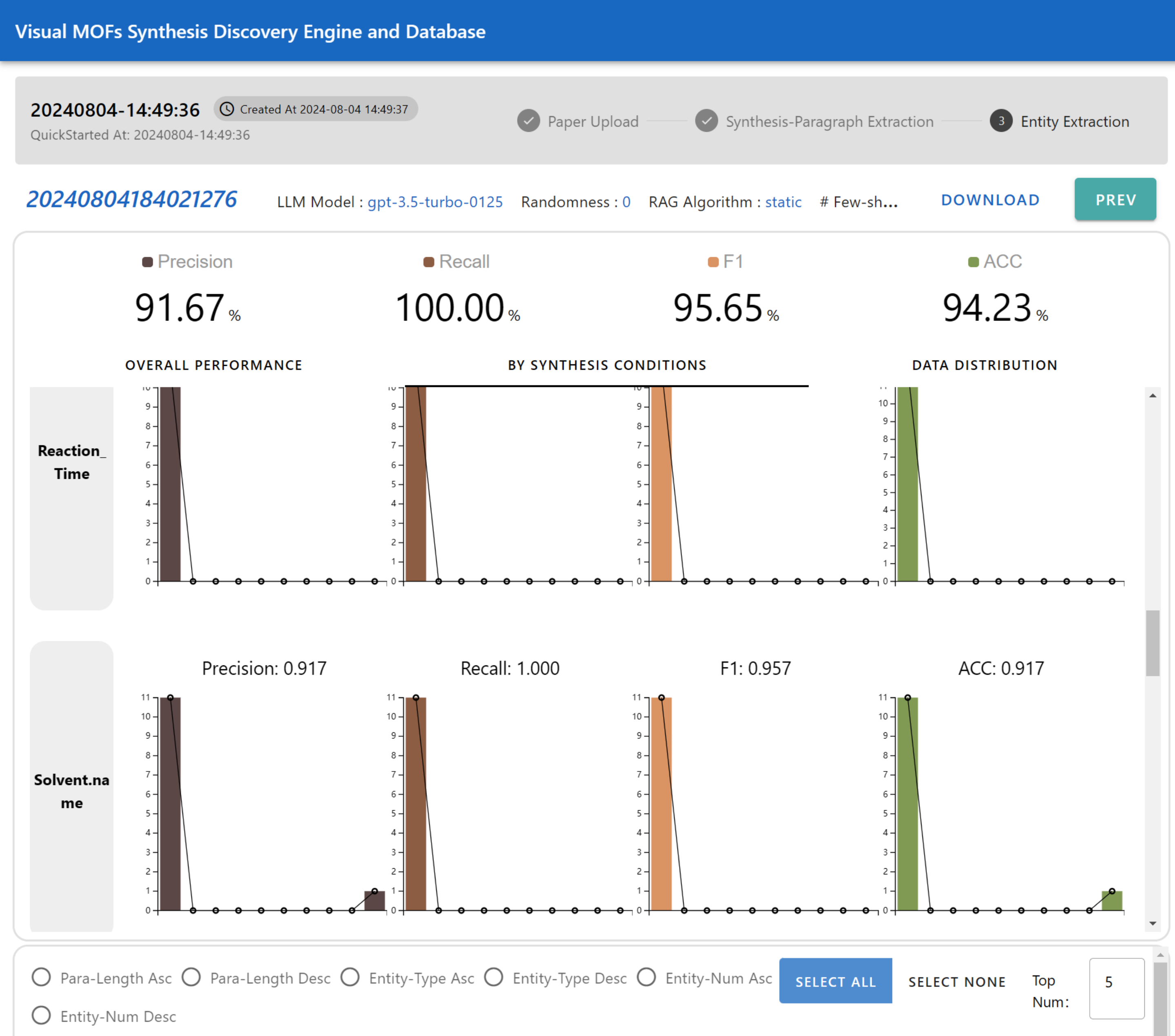}
    \caption{\textbf{Synthesis extraction performance panel on each uploaded MOFs literature.}}
    \label{fig:vis-panel-tab2}
\end{figure}

\clearpage

\begin{figure}
    \centering
    \includegraphics[width=\textwidth]{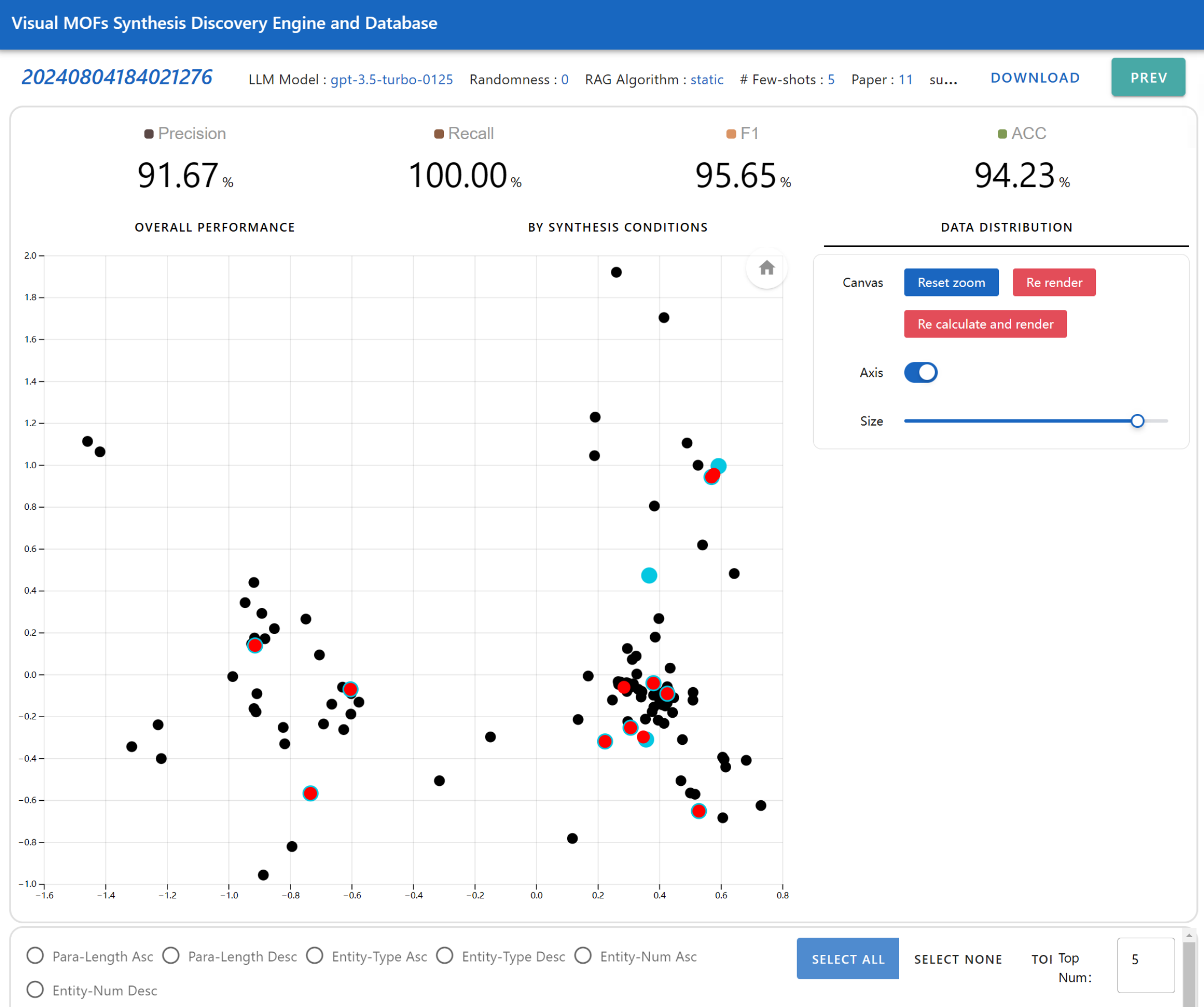}
    \caption{\textbf{Visualization of the synthesis condition distribution in the 2D projection view.}}
    \label{fig:vis-panel-tab3}
\end{figure}

\clearpage

\begin{figure}
    \centering
    \includegraphics[width=\textwidth]{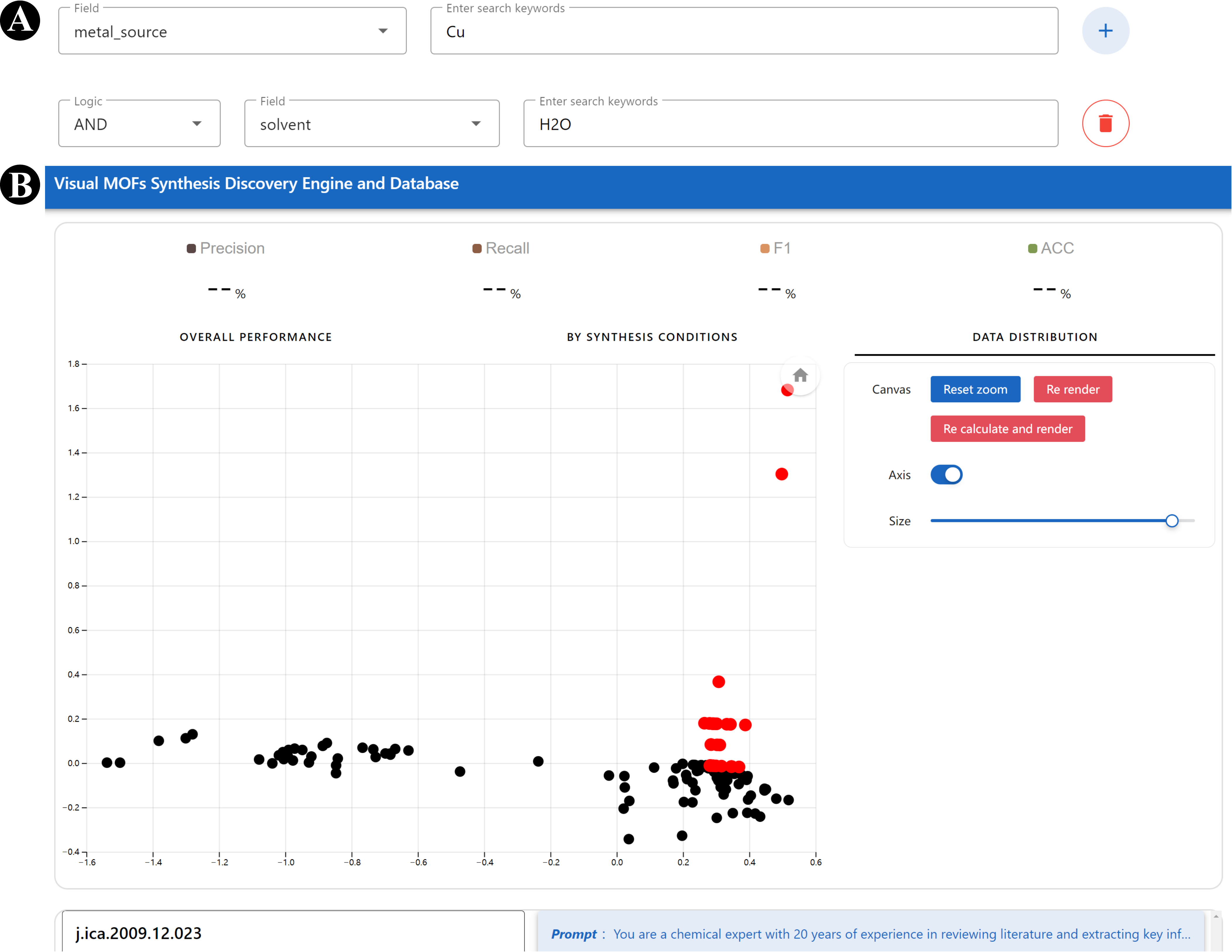}
    \caption{\textbf{Visualization interface for showing the result of database retrieval.} (A) Database queries; (B) Visualization of the retrieval result.}
    \label{fig:vis-panel-DB}
\end{figure}

\clearpage

\begin{figure}
    \centering
    \includegraphics[width=\textwidth]{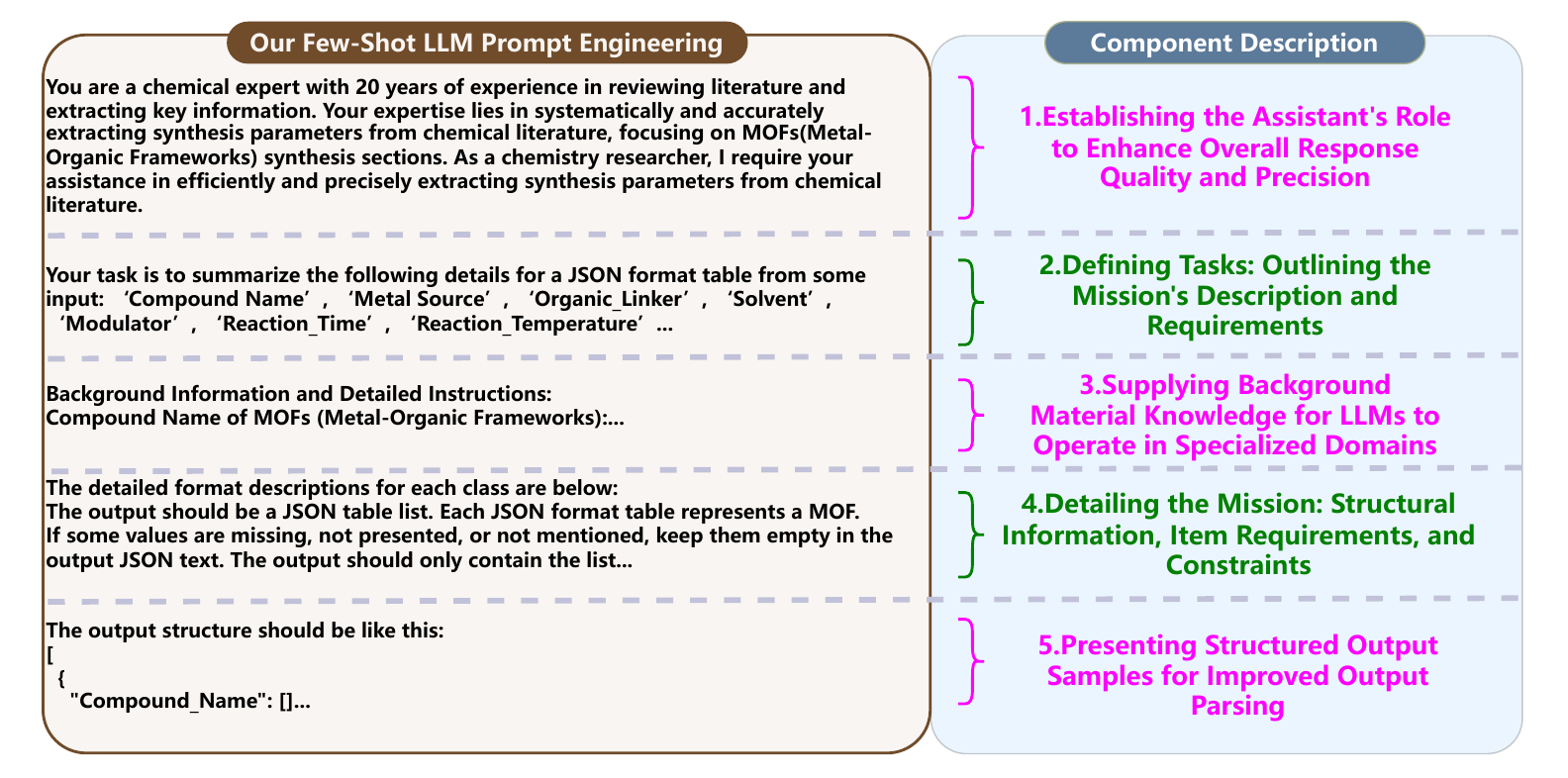}
    \caption{\textbf{Structure of the LLM prompt used throughout this work.}}
    \label{fig:PromptEngineeringOurs}
\end{figure}

\clearpage

\begin{figure}
    \centering
    \includegraphics[width=\textwidth]{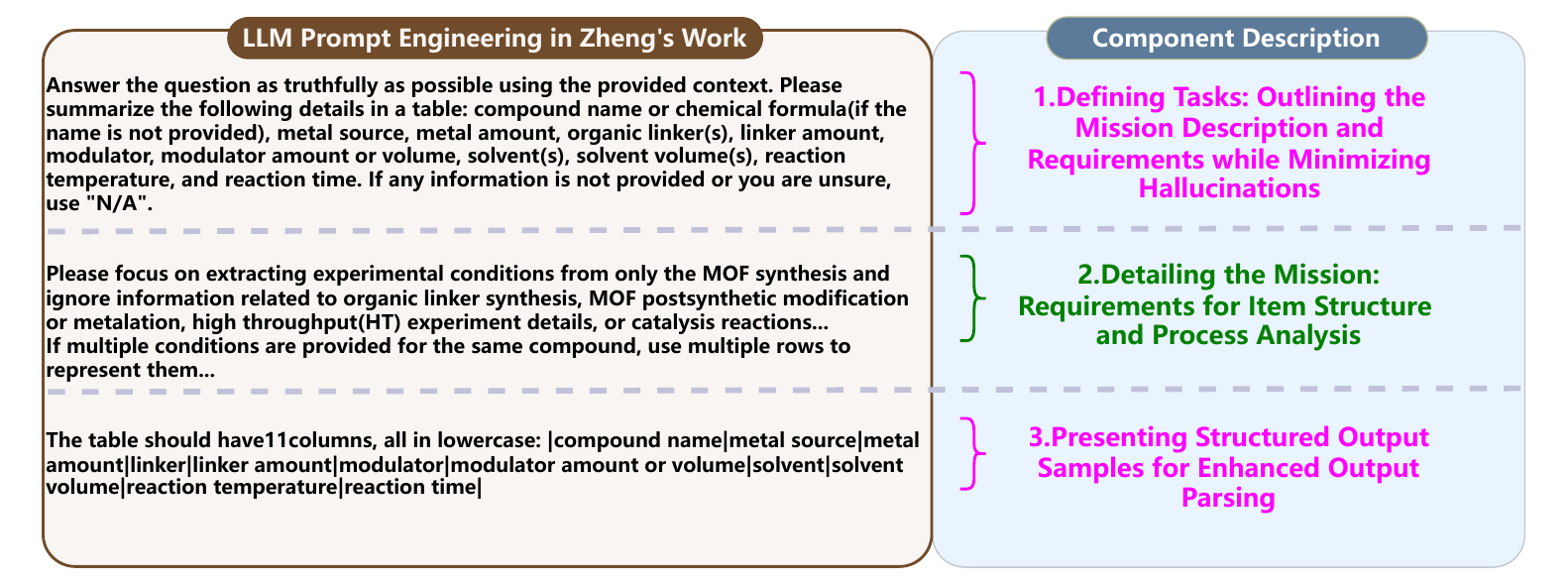}
    \vspace {-1.2em}
    \caption{\textbf{Structure of the LLM prompt used in Zheng's work \cite{zheng2023chatgpt}.}}
    \vspace{-0.15 in}
    \label{fig:PromptEngineeringBerkeley}
\end{figure}

%%%%%%%%%%% CAPTIONS FOR OTHER SUPPLEMENTARY FILES %%%%%%%%%%

\clearpage % Clear all remaining figures and tables then start a new page

\paragraph{Caption for Movie S1.}
\textbf{Demonstration of the online LLM-based MOFs synthesis extraction engine, database retrieval, and visualization.} The video shows a typical usage example composed of literature upload, synthesis paragraph detection, LLM-based synthesis route extraction, visualization-assisted extraction result analysis, as well as the information retrieval on all the extracted MOFs synthesis routes.

\paragraph{Caption for Data S1.}
\textbf{Extracted MOFs synthesis route data using LLM, as well as the original human-AI annotation data and the raw literature file. Additional tabs also contain training/test data for MOFs structure inference and design.} Both the data file and a full data description file are provided in the package.

\end{document}